\documentclass[10pt]{article} 
\usepackage[preprint]{tmlr}

\usepackage{amsmath,amsfonts,bm}

\def\eqref#1{equation~\ref{#1}}

\def\1{\bm{1}}

\DeclareMathAlphabet{\mathsfit}{\encodingdefault}{\sfdefault}{m}{sl}
\SetMathAlphabet{\mathsfit}{bold}{\encodingdefault}{\sfdefault}{bx}{n}

\usepackage{hyperref}
\usepackage{url}
\usepackage{graphicx}
\usepackage{subcaption}
\usepackage{booktabs}
\usepackage{algorithm}
\usepackage{algpseudocode}
\usepackage{accents}

\title{Meltdown: Circuits and Bifurcations in Point-Cloud-Conditioned 3D Diffusion Transformers}

\author{\name Maximilian Plattner \email plattner@ml.jku.at \\
      \addr Institute for Machine Learning, JKU Linz
      \AND
      \name Fabian Paischer  \\
      \addr Institute for Machine Learning, JKU Linz
      \AND
      \name Johannes Brandstetter \\ \addr Institute for Machine Learning, JKU Linz\\
      Emmi AI
      \AND
      \name Arturs Berzins \\ \addr Institute for Machine Learning, JKU Linz }

\def\month{MM}  
\def\year{YYYY} 
\def\openreview{\url{https://openreview.net/forum?id=XXXX}} 

\begin{document}

\maketitle

\begin{abstract}
Sparse point clouds are a common input modality for 3D surface reconstruction, including in safety-critical settings such as surgical navigation and autonomous perception. Recent point-cloud-conditioned 3D diffusion transformers achieve state-of-the-art results in this regime by leveraging learned priors. We show that these models can fail catastrophically under realistic input variation, and present a mechanistic case study of why.  We identify a failure mode we call \textit{Meltdown}: tiny on-surface perturbations to a sparse input point cloud can fracture the reconstructed output into hundreds of disconnected pieces. Adversarial search recovers Meltdown in 89.9--100\% of shapes across the two open-weight state-of-the-art architectures we study (\textsc{WaLa}, \textsc{Make-a-Shape}) on real-world datasets (GSO, SimJEB) and under both DDPM and DDIM sampling. We trace Meltdown along the forward pass: it is governed by how uniformly the points are distributed on the surface, faithfully transduced through the point-cloud encoder, and committed by a single early-denoising cross-attention write in the diffusion backbone. Diffusion-trajectory ensembles exhibit symmetry-breaking near this commit step, consistent with a bifurcation of the reverse process. Through a suite of matched-magnitude controls, we show that the variable on which the model commits is directional, concentrated in a low-rank subspace of the write's perturbation drift. Motivated by this finding, we introduce \texttt{PowerRemap}, a test-time control that reshapes the singular spectrum of the localized write to suppress this drift, with rescue rates of 98.3\% on \textsc{WaLa} and 84.6\% on \textsc{Make-a-Shape}. Together, these results link a circuit-level cross-attention mechanism to a trajectory-level account of the failure, demonstrating how mechanistic analysis can explain and guide behavior in conditional diffusion transformers.
\end{abstract}

\begin{figure}[h!]
    \centering
    \includegraphics[width=0.65\linewidth]{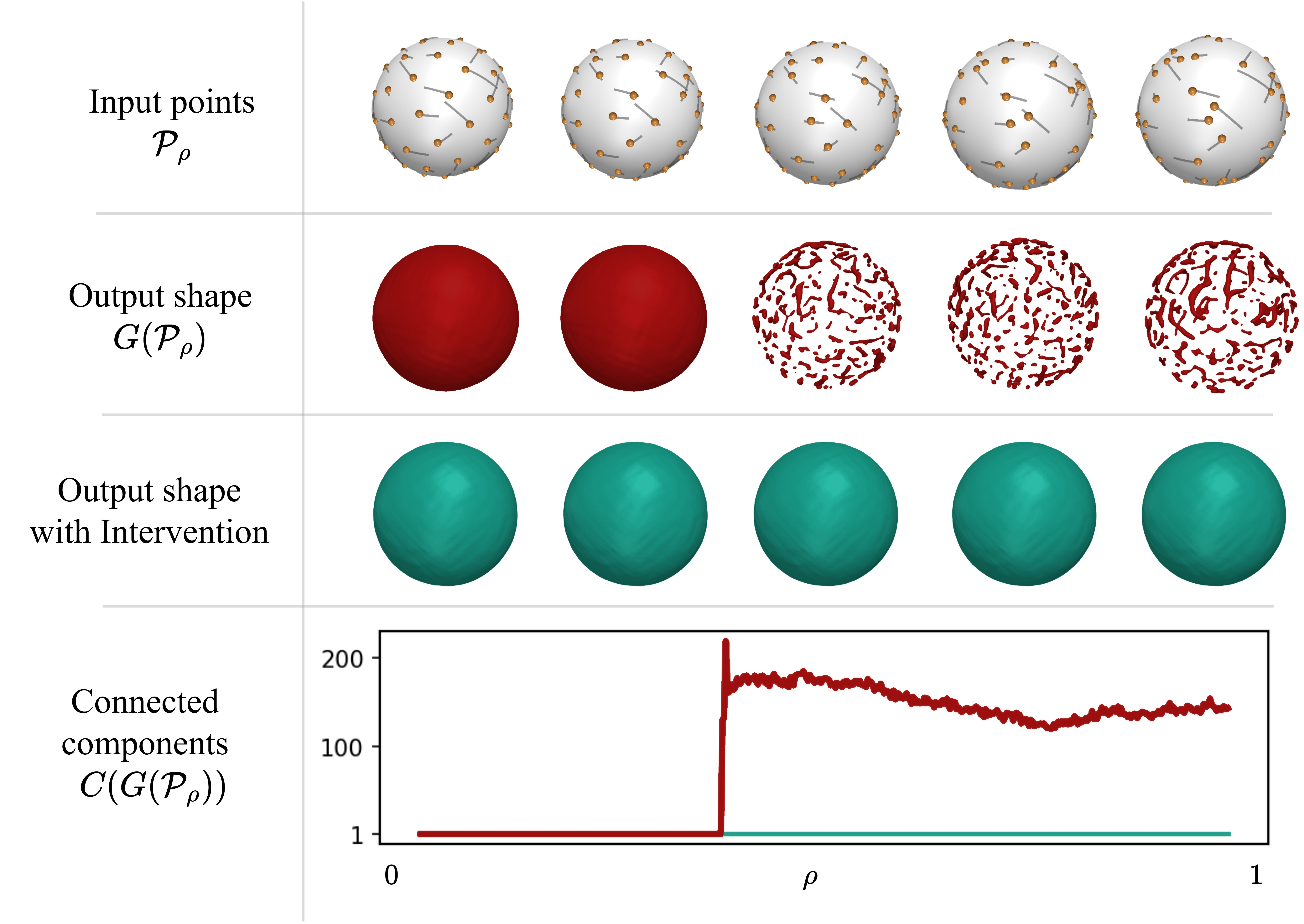}
\caption{\textbf{Meltdown.} We investigate point-cloud-conditioned diffusion transformers on the task of 3D surface reconstruction from sparse point clouds. Tiny on-surface perturbations to a point cloud can fracture the output into many disconnected pieces --- a failure we call \emph{Meltdown}. We trace Meltdown along the forward pass and localize a single cross-attention activation early in the denoising process that commits the trajectory to the fragmented attractor. Based on this analysis, we propose \texttt{PowerRemap}, a test-time intervention that stabilizes diffusion-based surface reconstruction under sparse conditions.}
    \label{fig:discover_Meltdown}
\end{figure}

\section{Introduction}
\label{sec:introduction}

Mechanistic interpretability has produced detailed circuit-level
accounts of how transformer language models compute, with rigorous
causal interventions identifying components responsible for specific
behaviors \citep{interpret_in_wild, automated_circuit, mi_open_problems}.
Extending these techniques to diffusion transformers is an emerging
direction \citep{tinaz2025emergenceevolutioninterpretableconcepts,
surkov2025onestepenoughsparseautoencoders, tang2022daaminterpretingstablediffusion,
shabalin2025interpretinglargetexttoimagediffusion}. We present a mechanistic case study on how large-scale diffusion transformers can fail unexpectedly, given the controlled task of 3D surface reconstruction from sparse point clouds.

3D surface reconstruction is a well-studied problem in computer vision
and graphics \citep{huang2022surfacereconstructionpointclouds}. Many
practical pipelines operate on point-cloud inputs, particularly where
passive imaging cannot reliably resolve geometry
\citep{stathopoulou2023survey}, including safety-critical settings such as surgical navigation \citep{Liu2024_neurosurgical} and perception for autonomous driving \citep{zhang2023perception}. In such settings, these point clouds are typically \emph{sparse}
\citep{quan2024deep, huang2024surface, sulzer2024survey}, which makes
reconstruction ill-posed and motivates the use of generative priors.
\emph{Diffusion transformers}, which attain state-of-the-art results
across many generative modalities \citep{edt_sketching, diff_language,
ditar, video_diffusion}, have recently been adapted to this task: priors
learned from large-scale datasets compensate for the missing geometric
information \citep{sanghi2024waveletlatentdiffusionwala,
hui2024makeashapetenmillionscale3dshape, direct3d, Motion2VecSets}. Understanding when and how such priors fail under realistic input variation is therefore a prerequisite for trustworthy deployment, and a natural target for mechanistic analysis. To
the best of our knowledge, two large-scale point-cloud-conditioned diffusion transformers are
currently open-weight\footnote{Concurrent work extends 3D generation in
different directions: point clouds as auxiliary control over
image-conditioned bases \citep{hunyuan3domni2025} and rectified-flow
rather than diffusion backbones \citep{xia2026points}.}: \textsc{WaLa}
\citep{sanghi2024waveletlatentdiffusionwala} and \textsc{Make-A-Shape}
\citep{hui2024makeashapetenmillionscale3dshape}.

We study \textsc{WaLa} and \textsc{Make-A-Shape} on the task of surface reconstruction from sparse point clouds. We observe a striking \emph{failure mode}: a tiny on-surface perturbation to the input point cloud can fracture the output into many disconnected pieces. We call this failure mode \emph{Meltdown} and identify it as a robustness concern for sparse-input deployment. We analyze the phenomenon through two complementary lenses: causal interventions on the network's internal circuits, and the bifurcation structure of the diffusion process itself.

First, we trace Meltdown along the forward pass: it is governed by how uniformly the input points are distributed on the surface, faithfully passed through the encoder, and committed by a single cross-attention activation early in the denoising process. Second, targeted controls show that the variable on which the model commits is directional. Motivated by this finding, we introduce \texttt{PowerRemap}, a test-time control on the localized activation that stabilizes sparse point-cloud conditioning. Third, interpreted through diffusion dynamics, the localized commit at the first denoising step sets the trajectory's position just before a symmetry-breaking bifurcation of the reverse process: small differences written into the residual stream by the cross-attention lever are amplified across a basin separatrix into the fragmented attractor over the subsequent steps.

Our contributions are summarized as follows:
\begin{enumerate}
    \item \textbf{Interpretability case study.} We provide a worked
    example of how circuit-level mechanistic analysis and
    diffusion-dynamics theory can be combined to explain and steer
    behavior in conditional diffusion transformers. We link a
    single-cell cross-attention mechanism, isolated by activation
    patching with matched-magnitude directional controls, to a
    trajectory-level account of spontaneous symmetry breaking in the
    reverse process.
    \item \textbf{Failure phenomenon: Meltdown.} We show that the
    state-of-the-art point-cloud-conditioned 3D diffusion transformers
    \textsc{WaLa}~\citep{sanghi2024waveletlatentdiffusionwala} and
    \textsc{Make-A-Shape}~\citep{hui2024makeashapetenmillionscale3dshape}
    perform surface reconstruction from sparse point clouds in a
    brittle manner: small on-surface perturbations to the input point
    cloud can fracture the output into multiple disconnected pieces.
    We call this failure phenomenon Meltdown.
    \item \textbf{Test-time intervention: \texttt{PowerRemap}.}
    Motivated by the mechanistic analysis, we propose a test-time
    spectral control at the identified cross-attention site.
    \texttt{PowerRemap} averts Meltdown in $98.3\%$ ($84.6\%$) of
    cases on the \textit{Google Scanned Objects (GSO)}
    dataset~\citep{downs2022googlescannedobjectshighquality} and in
    $97.7\%$ ($83.3\%$) of cases on the
    \textit{SimJEB}~\citep{Whalen_2021} dataset for \textsc{WaLa}
    (\textsc{Make-A-Shape}).
\end{enumerate}

We introduce the failure phenomenon, Meltdown, in
Section~\ref{sec:Meltdown}. In Section~\ref{sec:MI}, we analyze
Meltdown from the perspective of mechanistic interpretability.
Section~\ref{sec:power_remap_main} introduces our method,
\texttt{PowerRemap}, and presents results on the GSO and SimJEB
datasets. Finally, we link Meltdown to diffusion dynamics in
Section~\ref{sec:diffusion-dynamics} and discuss current limitations
in Section~\ref{sec:discussion}.

\section{Failure phenomenon: Meltdown}
\label{sec:Meltdown}

In this work, we investigate two leading open-weight point-cloud-conditioned 3D diffusion transformers, namely \textsc{WaLa} \citep{sanghi2024waveletlatentdiffusionwala} and \textsc{Make-a-Shape} \citep{hui2024makeashapetenmillionscale3dshape}. Such models can generate surfaces from point clouds, thus solving the \emph{surface reconstruction} task: given a set $\mathcal{P}=\{p_i\}_{i=1}^N\subset \mathcal{S} \subset \mathbb{R}^3$ of $N$ points sampled from an underlying surface $\mathcal{S}$, the model $G$ should reconstruct a surface consistent with the input and approximating the underlying surface $G(\mathcal{P}) \approx \mathcal{S}$. In many real-world scenarios (e.g., fast scene capture), $N$ can be small, i.e. the point cloud is \emph{sparse}.

As illustrated in Figure \ref{fig:discover_Meltdown}, we observe that there exist two sparse point clouds $\mathcal{P},\mathcal{Q}$ that are close in the input space, but the corresponding outputs differ severely: $G(\mathcal{P})$ is a connected surface while $G(\mathcal{Q})$ is a fragmented ``speckle'' of disconnected pieces. We will refer to this sudden catastrophic fracture as \emph{Meltdown}.

To study this failure phenomenon systematically, let us first introduce the topological quantity $C$ that counts the connected components of the output surface and serves as a quantifiable identifier of the healthy ($C=1$) versus unhealthy output ($C>1$). Furthermore, let us consider a running example where the points are sampled from a simple sphere: $\mathcal{S} = \{x : \|x\|_2=1\}$. This allows us to perform experiments that precisely control for the distribution of the points.
Specifically, we fix the random seed and first identify two point clouds of the same size $N=400$: $\mathcal{P}_0$ which produces a sphere output $C(G(\mathcal{P}_0))=1$ and $\mathcal{P}_1$ which produces a speckle output $C(G(\mathcal{P}_1))\gg1$ (typically around 100). Using spherical interpolation (geodesics on general surfaces), we can construct a continuous family of point clouds $\mathcal{P}_\rho \subset \mathcal{S}$. We sweep $\rho \in [0,1]$ and record $C(\rho) := C\left(G\left(\mathcal{P}_\rho\right)\right)$.

Figure \ref{fig:discover_Meltdown} illustrates the outcome of this experiment. 
As we sweep $\rho$ from $0$ to $1$, we first observe a long plateau of $C(\rho)=1$, followed by a sudden jump to $C(\rho)\gg1$ over a very narrow range of $\rho$. Refining the steps around this transition, we observe an effectively discontinuous jump in the macroscopic descriptor $C(\rho)$.

In Appendix \ref{app:experiments}, we report observing Meltdown across state-of-the-art point-cloud-conditioned diffusion transformers, i.e., \textsc{WaLa} \citep{sanghi2024waveletlatentdiffusionwala} and \textsc{Make-A-Shape} \citep{hui2024makeashapetenmillionscale3dshape}, real-world datasets, i.e., Google Scanned Objects \citep{downs2022googlescannedobjectshighquality} and SimJEB \citep{Whalen_2021}, and denoising strategies, i.e., DDIM \citep{song2021ddim} and DDPM \citep{ho2020denoisingdiffusionprobabilisticmodels}. Furthermore, we examine the prevalence of Meltdown depending on the sparsity of the input point-cloud in Appendix \ref{app:density}.


\section{Mechanistic analysis and intervention}
\label{sec:MI}

After observing and quantifying Meltdown, we ask what causes it. We center our analysis on \textsc{WaLa} for clarity, while demonstrating that our insights transfer robustly to \textsc{Make-a-Shape} (Appendix~\ref{app:experiments}).

\subsection{\textsc{WaLa}: Diffusion transformer}

Before we investigate the behavior, we briefly summarize the relevant parts of the \textsc{WaLa} diffusion transformer. A more detailed description is available in Appendix \ref{app:models}, the original work \citep{sanghi2024waveletlatentdiffusionwala}, and the references therein.

\paragraph{Transformer.}\label{sec:transformer}
\textsc{WaLa} is a latent diffusion model with a point-net encoder $E$, U‑ViT‑style \citep{hoogeboom2023} denoising backbone $B$, and VQ-VAE decoder \citep{oord2017neural} $D$.
The U-ViT $B = B^{K-1}\circ\cdots\circ B^0$ has $K=32$ transformer \emph{blocks} $B^k$. The condition $\mathbf{C} \in \mathbb{R}^{1024 \times 1024}$ enters via both AdaLN modulation \cite{esser2024} and cross‑attention. 
Denoting by \(\mathbf{Z}^{k}\in\mathbb{R}^{1728 \times 1152}\) the tokens entering the $k$-th block, it computes $B^k : \mathbf{Z}^{k} \mapsto \mathbf{Z}^{k+1}$ as a combination of multi-head self-attention $\mathrm{SA}$ and cross-attention $\mathrm{CA}$ layers (col. 2) with residual connections (col. 3):
\begin{subequations}
\begin{align}
\mathring{\mathbf{Z}} &= \mathring{\mathrm{AdaLN}}(\mathbf{Z}^{k}, \mathbf{C}), 
& \mathring{\mathbf{Y}} &= \mathrm{SA}(\mathring{\mathbf{Z}} ), 
& \mathring{\mathbf{R}} &= \mathring{\mathbf{Y}} + \mathbf{Z}^{k},
\label{eq:block_a} \\ 
{\mathbf{Z}} &= {\mathrm{AdaLN}}(\mathring{\mathbf{R}}, \mathbf{C}), 
& {\mathbf{Y}} & = \mathrm{CA}({\mathbf{Z}}, \mathbf{C}),
& {\mathbf{R}} &= {\mathbf{Y}} + \mathring{\mathbf{R}}, 
\label{eq:block_b} \\ 
\bar{\mathbf{Z}} &= \bar{\mathrm{AdaLN}} ( {\mathbf{R}}, \mathbf{C} ),
& \bar{\mathbf{Y}} &= \mathrm{MLP}( \bar{\mathbf{Z}} ),
& \mathbf{Z}^{k+1} &= \bar{\mathbf{Y}} + {\mathbf{R}}.
\label{eq:block_c} 
\end{align}
\end{subequations}
\paragraph{Diffusion.}
\textsc{WaLa} is trained in the standard DDPM \citep{ho2020denoisingdiffusionprobabilisticmodels} framework. At inference, the reverse diffusion maps an initial Gaussian latent \(\mathbf{Z}_T \sim \mathcal{N}(0, I)\) to \(\mathbf{Z}_0\) by iterating over a fixed schedule of denoising steps \(t \in \mathcal{T} = \{T, \ldots, 0\}\), where at each step the denoiser conditioned on \(\mathbf{C}\) updates \(\mathbf{Z}_t \to \mathbf{Z}_{t-1}\). At inference-time, we can sample using DDIM \citep{song2021ddim} or DDPM \citep{ho2020denoisingdiffusionprobabilisticmodels}.

We ask whether the failure is determined by the input cloud $\mathcal{P}$ itself , by the encoder $E$ that maps it to $\mathbf{C}$, or by the diffusion backbone $B$ that reads $\mathbf{C}$. In Appendix~\ref{app:input_sweep}, we establish a link to the
statistical properties of the input data. We find that a
classical sphere-uniformity functional, the Riesz $s{=}2$ energy,
is predictive of Meltdown. The $\rho$-path of is one trajectory through this scalar and crosses its threshold exactly where $C(\rho)$ jumps. Appendix~\ref{app:encoder_propagation} traces this scalar layer-by-layer through the encoder $E$ and finds it transduced into $\mathbf{C}$ without amplification or distortion. Hence, we conjecture that the failure must commit somewhere in the diffusion backbone $B$, on a signal that is already present in $\mathbf{C}$. We ask where: which submodule of which block $B^k$ at which denoising step $t$ reads the melt-relevant content of $\mathbf{C}$ in a way that decides the trajectory's fate.

To localize this commit site, we turn to \emph{activation patching}~\citep{heimersheim2024useinterpretactivationpatching, zhang2024bestpracticesactivationpatching}, a standard tool in mechanistic interpretability for testing the causal role of individual activations: by swapping a single activation between a healthy and an unhealthy forward pass and measuring the change in outcome, one isolates which sites carry the signal responsible for a behavior. Our setup is well-suited to this technique. The continuous $\rho$-path provides a controlled transition from a healthy run on $\mathcal{P}_0$ ($C{=}1$) to an unhealthy run on $\mathcal{P}_1$ ($C\gg 1$) along the Meltdown trajectory, and $C$ provides an objective scalar outcome to score against.

\subsection{Localizing Meltdown via activation patching}\label{sec:activation_patching}

\begin{figure}[h!]
    \centering
    \includegraphics[width=0.6\linewidth]{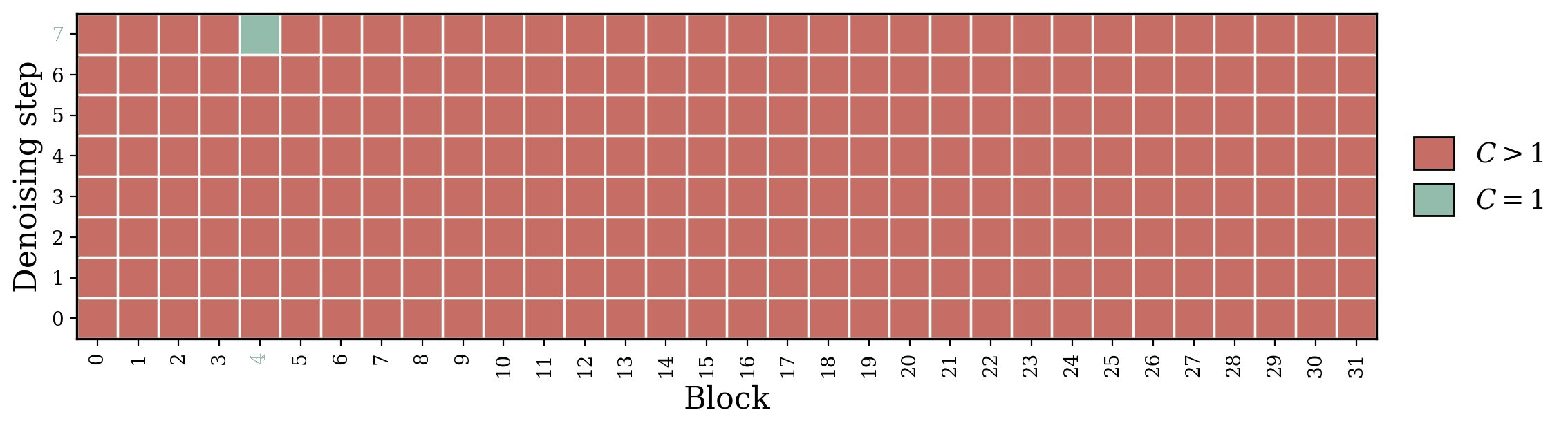}
    \caption{Our search in activation space finds that a single \textcolor[HTML]{8CBCAA}{cross-attention write $\mathbf{Y}_{4,7}$} controls Meltdown.}
    \label{fig:activation_patching_heatmap}
\end{figure}

Activations live on a depth-time grid indexed by block $k\in\mathcal{K}=\{0,\dots,31\}$ and denoising step $t\in\mathcal{T}=\{7,\dots,0\}$. Since the perturbation enters through $\mathbf{C}$, we begin at the sites where $\mathbf{C}$ is read --- AdaLN modulation and cross-attention --- giving $32\times 8\times 2 = 512$ conditioning cells. A single cell restores connectivity when its activation is replaced by the corresponding healthy value: the cross-attention write $\mathbf{Y}_{4,7}\in\mathbb{R}^{1728\times 1152}$ at block~$4$, first denoising step (Fig.~\ref{fig:activation_patching_heatmap}). We henceforth write $\mathbf{Y}\equiv\mathbf{Y}_{4,7}$. Appendix~\ref{app:act_patching_extended} extends the scan to all $19$ within-block activations across the full $32{\times}8$ grid ($4{,}864$ patches per seed). $\mathbf{Y}$ remains the unique site that rescues with full shape quality, and other rescuing components carry its signal downstream. \textsc{Make-A-Shape} exhibits the analogous early-step localization (Appendix~\ref{app:experiments}), and patching is robust across diffusion seeds and shape classes (Appendix~\ref{app:more_datapoints}). The early-step location is consistent with the latent being close to noise at $t{=}7$, so the model leans most heavily on conditioning then~\citep{liu2024faster}.
 
\subsection{Investigating the effect of patching}\label{sec:patching_effect}

Having localized $\mathbf{Y}$ as the single-cell handle on Meltdown, we ask what variable inside $\mathbf{Y}$ the model commits on. We will find that the answer is directional rather than scalar: the commit is concentrated in a low-rank subspace of $\mathbf{Y}$'s perturbation drift between healthy and unhealthy runs. We then introduce a scalar probe of $\mathbf{Y}$'s spectrum that co-moves reliably with this directional commit and serves as a useful diagnostic of the transition, while not itself being the causal variable.

\subsubsection{The committed variable is directional}\label{sec:spectral_causality}

To identify the variable on which the model commits, we ablate $\mathbf{Y}$ along the directions in which it moves with $\rho$: from the unhealthy $\mathbf{Y}$ we subtract its projection onto the top twenty singular directions of the drift between healthy and unhealthy runs. This rescues every diffusion seed we tested. Four magnitude-matched controls --- a random subspace orthogonal to the drift, scalar attenuation, isotropic Gaussian noise, and removing $\mathbf{Y}$'s own top components --- fail across thousands of runs (Appendix~\ref{app:Y_spectral_analysis}). The variable on which the model commits is therefore the directional content of $\mathbf{Y}$'s perturbation drift, concentrated in a low-rank subspace, and is not explained by Frobenius magnitude, isotropic perturbation, or own-basis truncation. \textsc{Make-a-Shape}, which exposes $\mathbf{Y}$ as one of only eight cross-attention writes rather than thirty-two, admits an analogous conclusion: removing the tail of $\mathbf{Y}$'s spectrum directly rescues, while removing its top is destructive (Appendix~\ref{app:mas_spectral_ablation}). The converse intervention is informative on the other side: transplanting the unhealthy $\mathbf{Y}$ into an otherwise-healthy run does not by itself induce Meltdown (Appendix~\ref{app:act_patching_extended}). Therefore, we conjecture that $\mathbf{Y}$ is a single-cell handle on the failure, and that handle is the directional content of its spectrum.

\subsubsection{A scalar probe of the transition}\label{sec:spectral_probe}

\begin{figure}[h]
    \centering
    \begin{subfigure}[b]{0.4\linewidth}
        \centering
        \includegraphics[width=\linewidth]{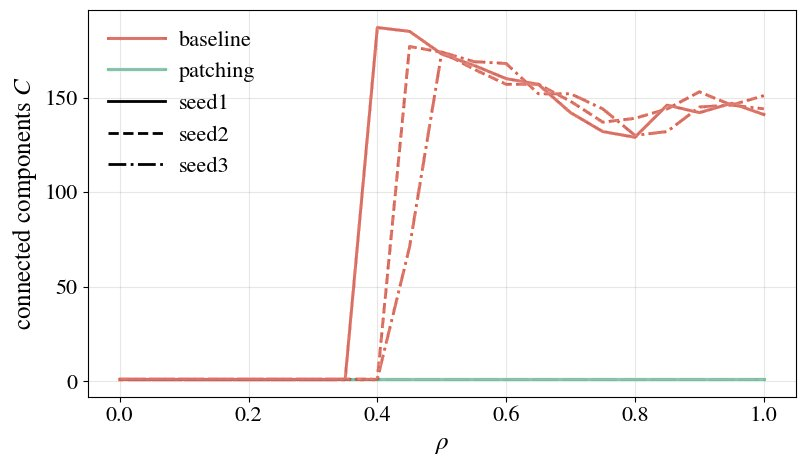}
        \caption{Connected components $C$ vs. $\rho$}
        \label{fig:wala_ddim_connectivty}
    \end{subfigure}
    \hfill
    \begin{subfigure}[b]{0.4\linewidth}
        \centering
        \includegraphics[width=\linewidth]{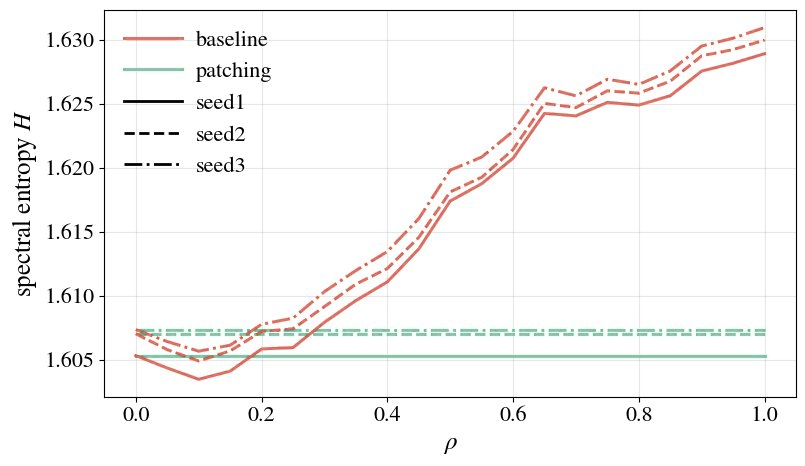}
        \caption{Spectral entropy $H$ vs. $\rho$}
        \label{fig:wala_ddim_spectral_entropy}
    \end{subfigure}
    \caption{The spectral-entropy probe $H$ co-moves with the connectivity transition. As we move from a healthy to an unhealthy run, the \textcolor[HTML]{C66E66}{baseline} case shows a smooth rise in $H$ and a sudden jump in $C$. \textcolor[HTML]{8CBCAA}{Patching} $\mathbf{Y}$ keeps both at healthy levels. The behavior is consistent across diffusion seeds. $H$ is diagnostic of the transition; the causal variable is the directional drift inside $\mathbf{Y}$ (\S\ref{sec:spectral_causality}).
    }
    \label{fig:connectivity_spectral_entropy_wala_side_by_sde}
\end{figure}

The directional commit lives in a high-dimensional, run-specific subspace and is not directly observable at test time. We therefore ask whether there is an observable scalar summary of $\mathbf{Y}$ that co-moves with this commit and can serve as a diagnostic of the transition. We considered several natural candidates along the $\rho$-sweep --- activation norm, the condition number, and entropy-based summaries of the singular spectrum (a comparison is in Appendix~\ref{app:more_spectral_metrics}). While most stay flat along $\rho$ or vary idiosyncratically across shapes and seeds, one candidate behaves consistently: the spectral entropy~\citep{Powell1979SpectralEntropy}
\begin{equation}
H(\mathbf{Y}) \;=\; -\sum_i p_i\log p_i,\qquad p_i\;=\;\sigma_i^2\big/\textstyle\sum_j\sigma_j^2,
\end{equation}
with $\sigma_i$ the singular values of $\mathbf{Y}$. Along the $\rho$-sweep, $H(\rho)$ rises smoothly while $C(\rho)$ jumps, and patching $\mathbf{Y}$ suppresses both (Fig.~\ref{fig:connectivity_spectral_entropy_wala_side_by_sde}). The pattern transfers to \textsc{Make-a-Shape} and holds across diverse shapes from GSO and SimJEB (Appendix~\ref{app:experiments}). $H$ is a robust diagnostic: a cheap, observable scalar that reliably tracks proximity to the transition across shapes, seeds, and architectures. It is not, however, the variable on which the model commits. The matched-magnitude controls of \S\ref{sec:spectral_causality} push $H$ both above and below the rescuing value without rescuing, while the directional surgery leaves $H$ essentially unchanged and rescues (Appendix~\ref{app:wala_H_not_cause}). $H$ co-moves with the directional commit but does not cause it.

The two roles of $\mathbf{Y}$'s spectrum are now distinct: the directional content of the spectrum is the surface on which the model commits (\S\ref{sec:spectral_causality}), and the entropy of the spectrum is an observable scalar that co-moves with this commit (\S\ref{sec:spectral_probe}). The next section uses this picture to build a deployable intervention.

\section{\texttt{PowerRemap}: a test-time intervention}
\label{sec:power_remap_main}

The directional surgery of \S\ref{sec:spectral_causality} localizes the lever but is not deployable: it requires the perturbation drift between a healthy and an unhealthy run, and therefore a healthy reference cloud which is unavailable at test time. We seek a single-pass intervention on $\mathbf{Y}$ that depresses this directional content without one. Both the lever and the probe live on the singular spectrum of $\mathbf{Y}$, so we operate on it directly: we modify the singular values while leaving the singular vectors fixed, so the feature directions written into the residual stream are preserved and only their energy distribution changes.

Concretely, let $\mathbf{Y} = U\Sigma V^\top$ with $\Sigma = \mathrm{diag}(\sigma_1, \ldots, \sigma_n)$ and $\sigma_1 \geq \cdots \geq \sigma_n \geq 0$. \texttt{PowerRemap} replaces $\Sigma$ with
\begin{equation}
\Sigma' = \mathrm{diag}(\sigma_1', \ldots, \sigma_n'), \qquad \sigma_i' = \frac{\sigma_i^\gamma}{\sigma_1^{\gamma-1}}, \qquad \gamma \geq 1,
\label{eq:powerremap}
\end{equation}
and writes $\mathbf{Y}' = U\Sigma' V^\top$ back into the residual stream at the site identified in \S\ref{sec:activation_patching}. A sweep applying \texttt{PowerRemap} at every cross-attention and MLP
site across \textsc{WaLa}'s U-ViT corroborates that the intervention is
site-specific: rescues are confined to $\mathbf{Y}_{4,7}$ and two
upstream feeders ($\mathbf{Y}_{3,7}$, $\mathrm{MLP}_{0,7}$) at the
first denoising step (Appendix~\ref{app:powerremap_sweep}). The strength $\gamma$ controls how sharply the tail is suppressed: $\gamma = 1$ recovers the identity, and larger $\gamma$ concentrates the spectrum onto the leading singular values. The effective $\gamma$ is model-dependent (WaLa: $\gamma = 100$; \textsc{Make-a-Shape}: $\gamma = 1.05$; Appendix~\ref{app:choice_gamma_remedy}), reflecting the architectural redundancy gap between the two models: WaLa exposes $\mathbf{Y}$ as one of $32$ cross-attention writes plus AdaLN, so $\mathbf{Y}$ tolerates aggressive compression; \textsc{Make-a-Shape} exposes $\mathbf{Y}$ as one of only $8$ writes, where the same operation must be milder to preserve the conditioning content the model cannot spare.

\subsubsection{Evaluation at scale}
\label{sec:eval_at_scale}

We assess whether Meltdown and the effectiveness of \texttt{PowerRemap}
generalize across diverse input geometries on two datasets, \emph{Google
Scanned Objects (GSO)}~\citep{downs2022googlescannedobjectshighquality}
and \textit{SimJEB}~\citep{Whalen_2021}, neither of which was used to
train \textsc{WaLa} or \textsc{Make-a-Shape}. The evaluation protocol
is detailed in Appendix~\ref{app:gso}.

For \textsc{WaLa} we evaluate every shape with a single global
$\gamma{=}100$. For \textsc{Make-a-Shape}, the milder $\gamma$ regime
identified in \S\ref{sec:power_remap_main} requires per-shape
calibration: we use category-stratified subsets ($130$ GSO,
$30$ SimJEB shapes) and select $\gamma$ per shape by grid search using
output connectivity $C{=}1$ as the criterion --- which requires no
ground-truth surface and is therefore deployable at test time.
Effective $\gamma$ values cluster tightly (median $1.10$,
$\sigma{=}0.13$ on GSO; median $1.05$, $\sigma{=}0.063$ on SimJEB;
Appendix~\ref{app:choice_gamma_remedy}), so the grid can be small in
practice. Subset sizes for \textsc{Make-a-Shape} reflect its $100$-step
DDIM schedule (vs.\ $8$ for \textsc{WaLa}), which puts full-corpus
$\gamma$-search outside our compute budget.

\begin{figure}[h!]
    \centering
    \includegraphics[width=0.65\linewidth]{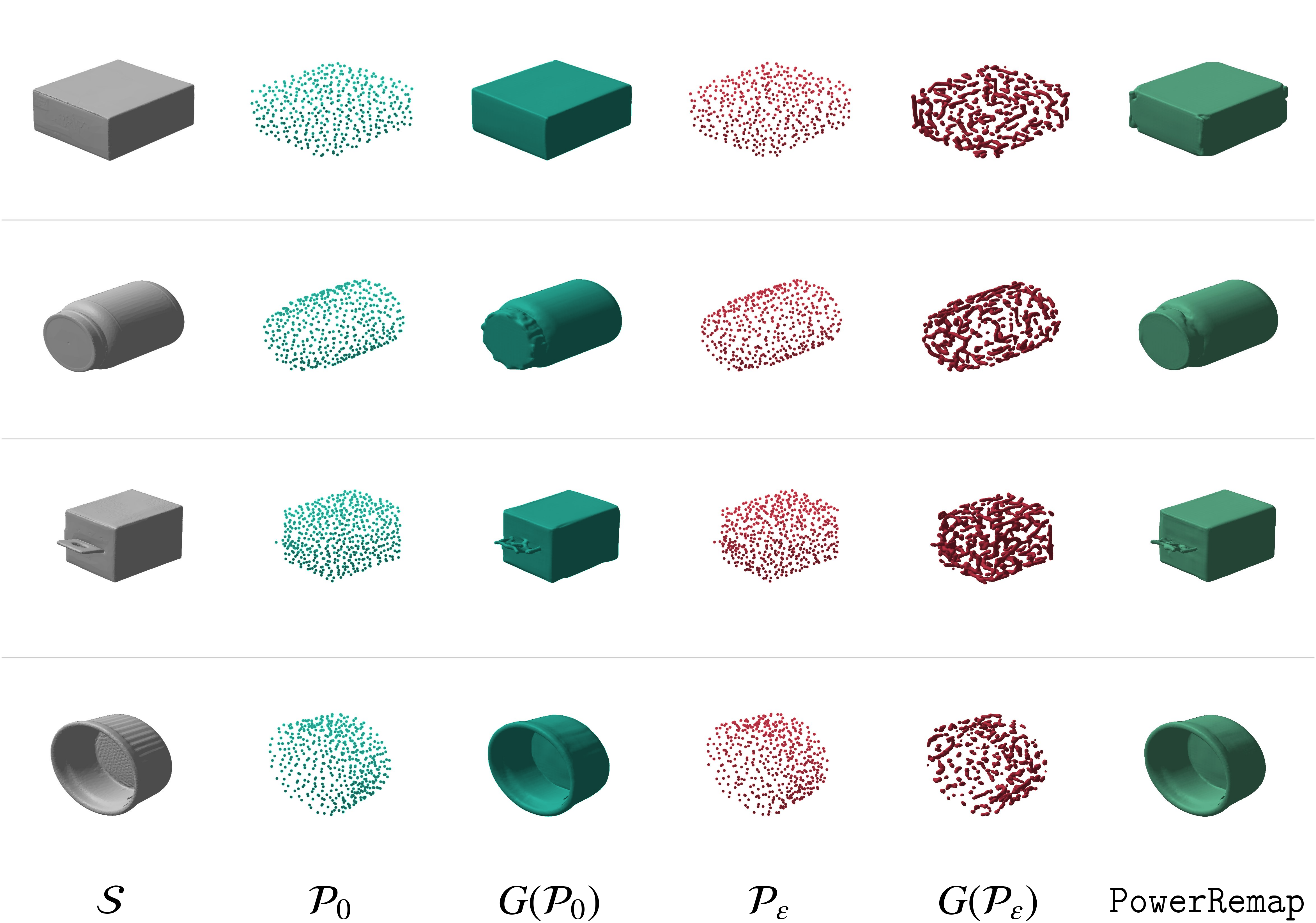}
\caption{%
\textbf{Qualitative \textsc{WaLa} results on \textit{Google Scanned Objects}}, one shape per row. For each shape, the \textcolor[HTML]{7A7A7A}{ground-truth surface $\mathcal{S}$} is sampled to produce a sparse \textcolor[HTML]{186C60}{healthy point cloud $\mathcal{P}_0$}, from which \textsc{WaLa} reconstructs a connected baseline \textcolor[HTML]{186C60}{$G(\mathcal{P}_0)$} with $C{=}1$. A tiny on-surface perturbation yields \textcolor[HTML]{A63444}{$\mathcal{P}_\varepsilon$}, yet the corresponding output \textcolor[HTML]{A63444}{$G(\mathcal{P}_\varepsilon)$} fractures into hundreds of disconnected pieces ($C{\gg}1$). We call this phenomenon Meltdown. \textcolor[HTML]{396E57}{\texttt{PowerRemap}}, applied to the same perturbed input, restores connectivity. Adversarial search (Alg.~\ref{alg:alg_adverserial_Meltdown_search}) finds Meltdown in $89.9\%$ of \textsc{WaLa} runs on GSO, of which \texttt{PowerRemap} rescues $98.3\%$ (Table~\ref{tab:gso_pr_category_top5_other}).%
}
    \label{fig:your_qualitative_panel_gso_wala}
\end{figure}

\paragraph{GSO.} GSO~\citep{downs2022googlescannedobjectshighquality}
is a diverse corpus of $1{,}030$ scanned household objects. We identify
Meltdown in $89.9\%$ of \textsc{WaLa} runs and $100\%$ on the
\textsc{Make-a-Shape} subset. \texttt{PowerRemap} stabilizes $98.3\%$
of \textsc{WaLa} failures and $84.6\%$ on the \textsc{Make-a-Shape}
subset (Table~\ref{tab:gso_pr_category_top5_other}, left; qualitative
examples in Fig.~\ref{fig:your_qualitative_panel_gso_wala}).

\paragraph{SimJEB.} SimJEB~\citep{Whalen_2021} is a benchmark of $381$
3D jet-engine bracket CAD models. We identify Meltdown in $92.4\%$ of
\textsc{WaLa} runs and $100\%$ on the \textsc{Make-a-Shape} subset.
\texttt{PowerRemap} stabilizes $97.7\%$ of \textsc{WaLa} failures and
$83.3\%$ on the \textsc{Make-a-Shape} subset
(Table~\ref{tab:gso_pr_category_top5_other}, right).

Additional experiments show that representative alternative interventions such input-cloud uniformization and noise injection do not rescue Meltdown, while \texttt{PowerRemap} does (Appendix~\ref{app:simpler_baselines}).
Further experiments show that Meltdown and the efficacy of
\texttt{PowerRemap} generalize to multi-object inputs
(Appendix~\ref{app:mult_objects}).

\begin{table}[h!]
\centering
\caption{Category-wise evaluation of \texttt{PowerRemap}. \emph{Left:} \textsc{WaLa} on the full GSO and SimJEB datasets — stabilization rate $98.3\%$ (GSO) and $97.7\%$ (SimJEB). \emph{Right:} \textsc{Make-A-Shape} on category-representative subsets — $84.6\%$ (GSO) and $83.3\%$ (SimJEB).}
\label{tab:gso_pr_category_top5_other}
\label{tab:simjeb_main_wala}
\label{tab:gso_pr_category_top5_MAS_main}
\label{tab:simjeb_Meltdown_mas}
\scriptsize
\setlength{\tabcolsep}{3pt}
\renewcommand{\arraystretch}{0.95}
\begin{minipage}[t]{0.49\linewidth}
\centering
\begin{tabular*}{\linewidth}{@{\extracolsep{\fill}}lrrr@{}}
\toprule
\multicolumn{4}{c}{\textbf{\textsc{WaLa}}} \\
\midrule
\textbf{Category} & \textbf{Shapes} & \textbf{Meltdown [\%]} & \textbf{Rescue [\%]} \\
\midrule
\multicolumn{4}{c}{\textit{GSO}} \\
\midrule
Shoe            & 254  & 97.2 & 99.6 \\
Consumer goods  & 248  & 97.6 & 99.2 \\
Unknown         & 216  & 88.4 & 95.8 \\
Other           & 112  & 92.9 & 99.0 \\
\midrule
\textbf{Total}  & \textbf{1030} & \textbf{89.9} & \textbf{98.3} \\
\midrule
\multicolumn{4}{c}{\textit{SimJEB}} \\
\midrule
Arch       &  37 &  89.2 & 100.0 \\
Beam       &  46 & 100.0 & 100.0 \\
Block      &  99 &  87.9 &  97.7 \\
Butterfly  &  43 &  93.0 &  95.0 \\
Flat       & 147 &  93.2 &  97.8 \\
Other      &   9 & 100.0 &  88.9 \\
\midrule
\textbf{Total} & \textbf{381} & \textbf{92.4} & \textbf{97.7} \\
\bottomrule
\end{tabular*}
\end{minipage}\hfill
\begin{minipage}[t]{0.49\linewidth}
\centering
\begin{tabular*}{\linewidth}{@{\extracolsep{\fill}}lrrr@{}}
\toprule
\multicolumn{4}{c}{\textbf{\textsc{Make-A-Shape}}} \\
\midrule
\textbf{Category} & \textbf{Shapes} & \textbf{Meltdown [\%]} & \textbf{Rescue [\%]} \\
\midrule
\multicolumn{4}{c}{\textit{GSO (subset)}} \\
\midrule
Consumer goods    &  60 & 100.0 & 90.0 \\
Bottles/cans/cups &  23 & 100.0 & 95.7 \\
Unknown           &  18 & 100.0 & 83.3 \\
Other             &  29 & 100.0 & 65.5 \\
\midrule
\textbf{Total}    & \textbf{130} & \textbf{100.0} & \textbf{84.6} \\
\midrule
\multicolumn{4}{c}{\textit{SimJEB (subset)}} \\
\midrule
Arch      &  2 & 100.0 &  50.0 \\
Beam      &  4 & 100.0 &  75.0 \\
Block     &  9 & 100.0 & 100.0 \\
Butterfly &  3 & 100.0 &  66.7 \\
Flat      & 11 & 100.0 &  90.9 \\
Other     &  1 & 100.0 &   0.0 \\
\midrule
\textbf{Total} & \textbf{30} & \textbf{100.0} & \textbf{83.3} \\
\bottomrule
\end{tabular*}
\end{minipage}
\end{table}

\section{Diffusion dynamics}
\label{sec:diffusion-dynamics}

\emph{Diffusion dynamics} refers to a collection of ideas describing the generative diffusion process using established theory from statistical physics \citep{raya2023spontaneous, biroli2024dynamical, Yu25noneq, ambrogioni2025information}, information theory \citep{ambrogioni2025information}, information geometry \citep{chen2023geometric, ventura2025manifolds}, random-matrix theory \citep{ventura2025manifolds}, and dynamical systems \citep{ambrogioni2025information}.
Key concepts from diffusion dynamics allow us to frame both the observed failure phenomenon and the intervention, ultimately connecting the mechanistic analysis to a theoretically established interpretation of the generative diffusion process.

\subsection{Preliminaries}

We introduce key ideas of diffusion dynamics adapted from \citet{raya2023spontaneous, biroli2024dynamical, ambrogioni2025information}.
The reverse-time diffusion can be viewed as a noisy gradient flow in a time-dependent potential $u(\cdot,s)$: 
\begin{equation}
\label{eq:potential}
\mathrm{d}\mathbf{X}_t = -\nabla_\mathbf{x} u(\mathbf{X}_t,s)\mathrm{d}t + g(s)\mathrm{d}\mathbf{W}_t,
\qquad
u(\mathbf{x},s) = -g^{2}(s)\log p(\mathbf{x},s) + \Phi(\mathbf{x},s),
\end{equation}
where $p(\cdot,s)$ is the forward marginal, $g$ is the noise scale, and $\Phi(\mathbf{x},s)=\int_0^\mathbf{x}f(\mathbf{z})\mathrm{d}\mathbf{z}$ integrates the forward drift $f$.
The potential $u$ is essentially a scaled and shifted marginal.
The critical points $x^*$ of this potential $\nabla u(x^*,s)=0$ are the \emph{attractors} of the dynamics.
Early in the generation ($t\approx0$), there is a global symmetric basin with a stable central fixed point, and the trajectories exhibit mean-reverting fluctuations around it. 
As noise decreases, the energy landscape deforms and, at a critical time $\tau^*$, the fixed point loses stability and the landscape \emph{bifurcates} into two basins. 
Such bifurcations repeat until at $t\approx T$ the potential has many fixed points aligning with the data modes (i.e., the data points under an exact score assumption).
These bifurcation times $\tau^*$ can be interpreted as \emph{decision} times where the sample trajectory is committed to a future attractor basin.

Around the degenerate critical point $x^*(\tau^*)$, two paths that are nearby for $t<\tau^*$ may diverge exponentially for $t>\tau*$ due to the Lyapunov exponent becoming positive (the smallest eigenvalue of $\nabla^2 u$ obtained from linearizing the reverse dynamics around the critical point). This can amplify tiny input differences and is the mechanism behind sending trajectories to different attractors.

This selection of one among many symmetry-equivalent states is called \emph{spontaneous symmetry breaking}. A canonical example is a ferromagnet: at high temperature ($t\approx T$) spins are disordered, while as $t \to 0$ they align. Any magnetization direction is a priori equivalent, yet each realization picks one. The underlying symmetry is visible only in the ensemble over many realizations.

\subsection{Application to Meltdown and intervention}

To test the diffusion dynamic perspective of the Meltdown phenomenon and the intervention, we perform several experiments predicted by this view. 
However, we must first introduce conditioning in the above diffusion dynamics view. For a fixed condition $\mathbf{C}$, this extension is trivial: simply modify the marginal $p(\cdot,s) = p(\cdot,s|\mathbf{C})$.
However, a family of conditions, like the univariate interpolation $\{\mathbf{C}(\rho) | \rho \in [0,1]\}$, introduces an additional dependence in the above formalism, and it is not obvious how to analyze the evident bifurcation around $\mathbf{C}^*$ instead of $\tau^*$. 
Fortunately, since the symmetry breaking originates \emph{locally} around the bifurcation time $\tau^*$ and point $x^*$, a small change in the condition $\mathrm{d}\mathbf{C}$ can be related to a small change in the initial condition $\mathrm{d}x_T$ through the total differential of the reverse path $\gamma : (t, x_T, \mathbf{C}) \mapsto x_t$. Qualitatively, this allows us to consider different $x_T$ for a fixed $\mathbf{C}$ in place of different $\mathbf{C}$ for a fixed $x_T$.

\begin{figure}[h!]
    \centering
    \begin{subfigure}[b]{0.4\linewidth}
        \centering
        \includegraphics[width=\linewidth]{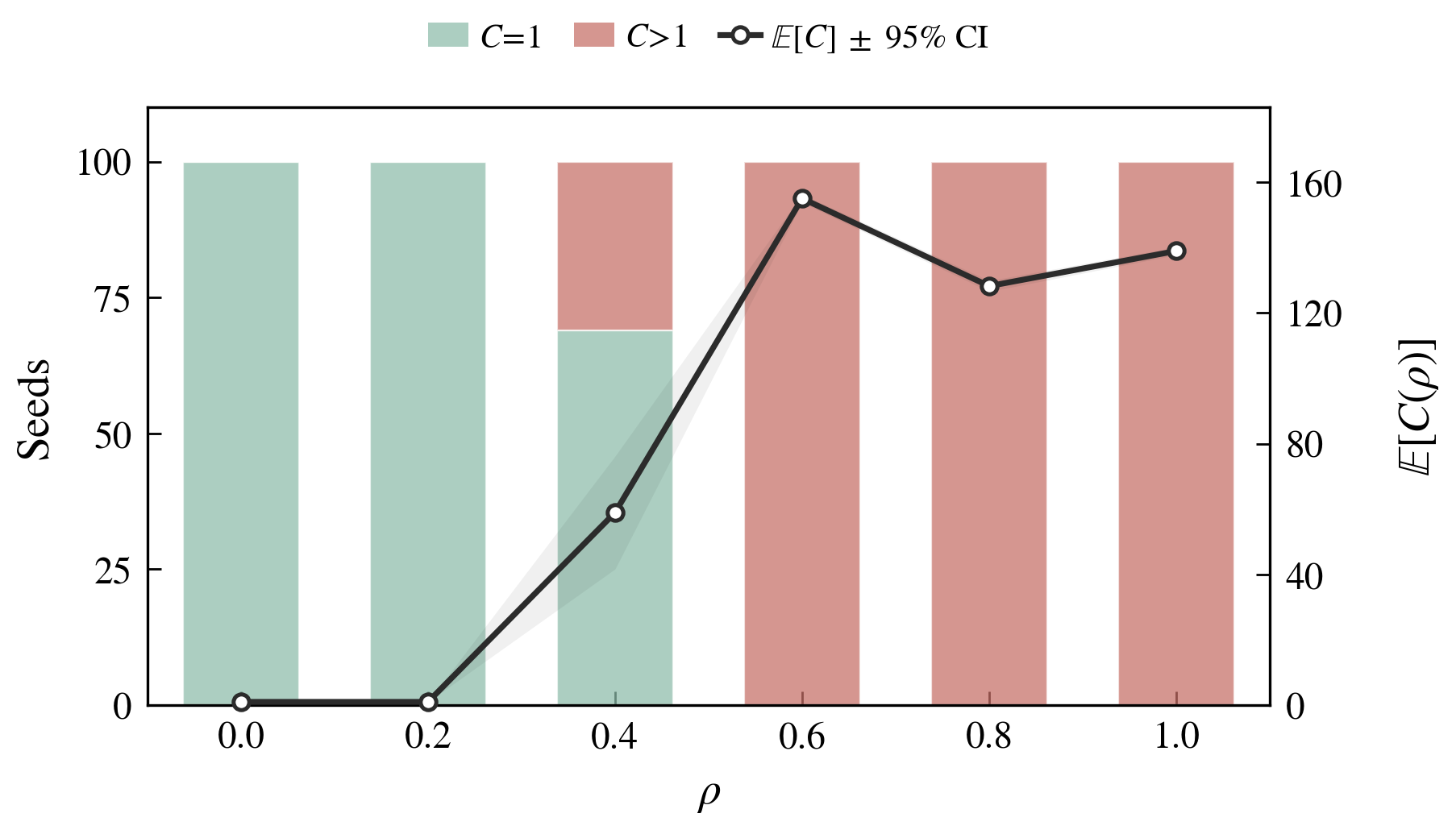}
        \caption{}
        \label{fig:ensemble}
    \end{subfigure}
    \hfill
    \begin{subfigure}[b]{0.3\linewidth}
        \centering
        \includegraphics[width=\linewidth]{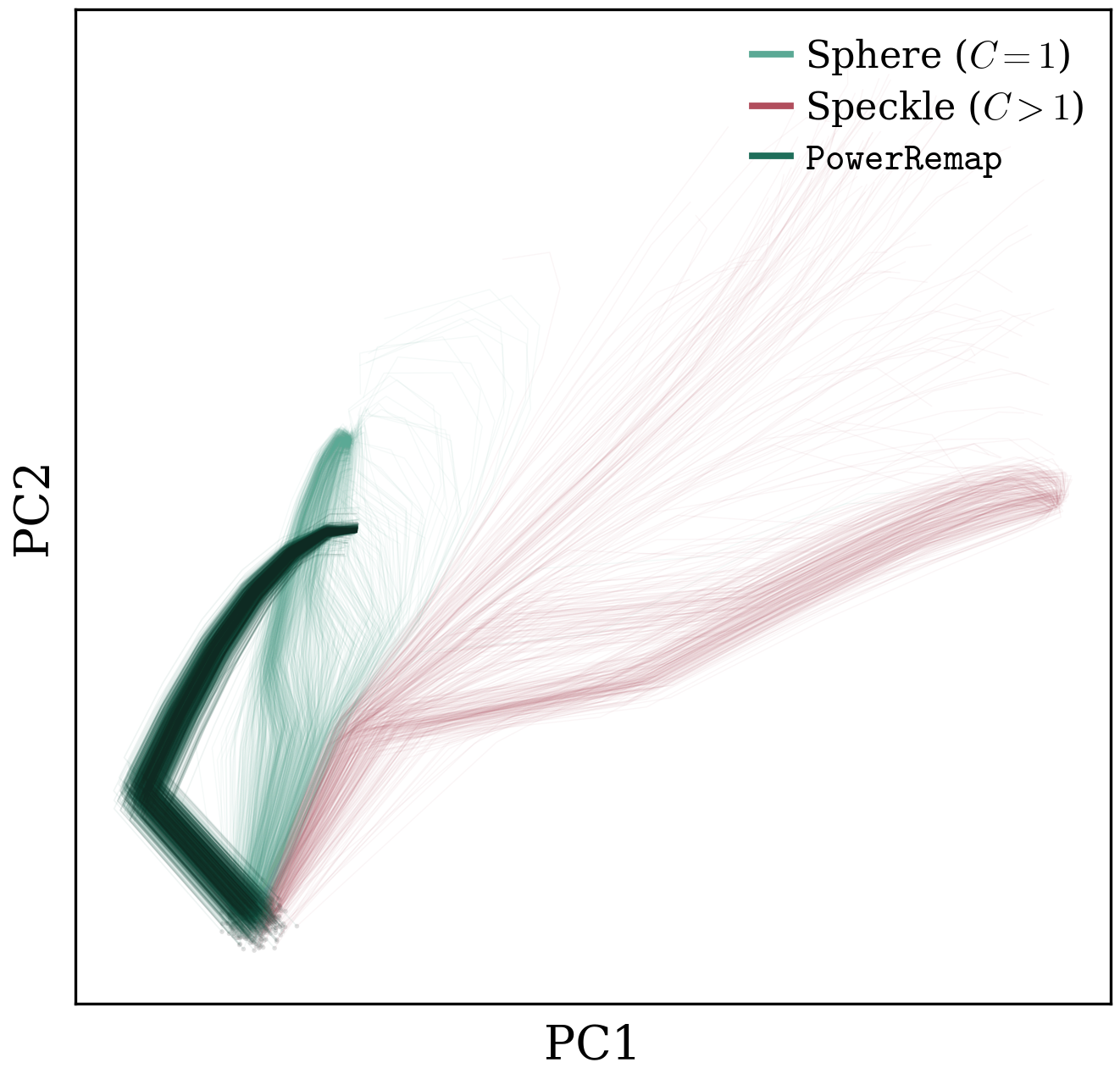}
        \caption{}
        \label{fig:trajectories}
    \end{subfigure}
    \caption{
    A collection of diffusion trajectories reveals additional insights about the Meltdown phenomenon.
    (a) In expectation over the initial noise, both the sphere and speckle shapes are produced at intermediate conditions, relaxing the sharp Meltdown behavior for a fixed initial noise.
    (b) Latent diffusion trajectories projected onto a 2D linear subspace spanned by the first two principal components of the final distribution of the baseline. The \texttt{PowerRemap} trajectories in green form a tight bundle following a different path that converges to a minor mode of the baseline distribution.
    }
    \label{fig:side_by_side}
\end{figure}

\paragraph{Ensemble.} Spontaneous symmetry breaking suggests that even though a single trajectory commits to a single attractor (sphere versus speckle), both ``symmetric'' configurations are visited over an ensemble of random trajectories. We record the trajectories for 100 initial conditions $x_T \sim \mathcal{N}(0, I)$ over the $\rho\in[0,1]$ range and plot the resulting shape connected component distribution in Figure \ref{fig:ensemble}. The extremes $\rho=0,1$ are far from a critical condition, and all trajectories converge to the respective attractors. However, at the intermediate conditions, the ensemble of trajectories visits both attractors, with the ratio of fractured shapes increasing steadily with $\rho$. In expectation, the component curve $\mathbb{E}_{x_T}[C(\rho)]$ exhibits a smooth behavior, relaxing the discrete jump in $C(\rho)$ for a single $x_T$.

\paragraph{Trajectories.} At intermediate $\rho{=}0.4$ we visualize $1000$ trajectories projected onto the first two principal components of $p(\cdot,0)$ (Fig.~\ref{fig:trajectories}). The first denoising step is mean-reverting~\citep{biroli2024dynamical, ventura2025manifolds}. The second step marks the symmetry breaking. We confirm this with Hartigan's dip test \citep{Hartigan1985TheDT} on the projection onto $\hat{\mathbf{u}}=(\mathbf{c}_{\mathrm{sphere}}-\mathbf{c}_{\mathrm{speckle}})/\|\cdot\|$: unimodality is not rejected at $t{=}T$ or after step~1 (Holm-adjusted $p{=}1.0$), but is rejected from step~2 onward ($p{=}4.5\times 10^{-3}$ at $t{=}5$, $p{<}10^{-4}$ thereafter). The intervention alters the first step, after which trajectories flow to a tight minor mode of the baseline.

\begin{figure}[h]
    \centering
    \includegraphics[width=\linewidth]{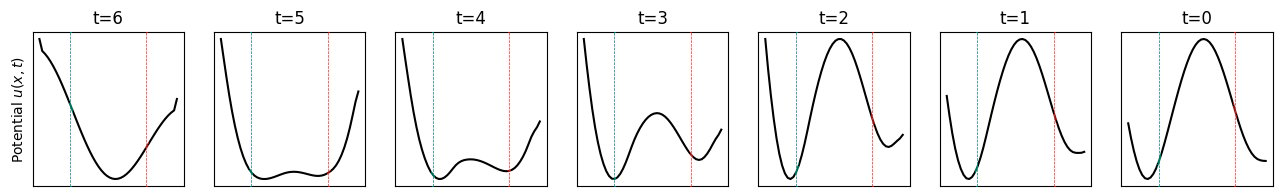}
    \caption{The potential $u$ (related to the marginal probability via Eq.~(\ref{eq:potential})) reveals the two diffusion stages separated by the bifurcation time $\tau^*\approx5$, where the single potential well flattens and splits into the two attractor basins. The particle's location just before this early bifurcation commits it to the final attractor and ultimately determines the generated shape. Small perturbations around this time become amplified, giving the appearance of discrete jumps that characterize the observed Meltdown.}
    \label{fig:potential}
\end{figure}

\paragraph{Potential.} We calculate the potential similar to the procedure introduced by \citet{raya2023spontaneous}. We select a pair of representative trajectories from each attractor and interpolate between them along a variance-preserving curve $x_t(\alpha) = \cos(\alpha)x^\text{sphere}_t + \sin(\alpha)x^\text{speckle}_t$ for $\alpha \in [-0.2\pi, 1.2\pi]$. Figure \ref{fig:potential} reveals the two diffusion stages separated by the bifurcation time $\tau^*\approx5$, where the single potential well flattens and splits into the two attractor basins.

We note that the mechanistic localization in \S\ref{sec:activation_patching} and the bifurcation analysis below describe two distinct moments of the reverse trajectory. The cross-attention lever $\mathbf{Y}_{4,7}$ writes into the residual stream at the first denoising step ($t{=}7$); the potential bifurcates around $\tau^*\approx 5$ (Fig.~\ref{fig:potential}), and Hartigan's dip test rejects unimodality from $t{=}5$ onward. The lever therefore acts \emph{upstream} of the basin split: it sets the trajectory's position at the moment the basins form, after which the standard diffusion-dynamics mechanism --- exponential separation under a positive Lyapunov exponent --- amplifies the difference across the separatrix. This sequencing also explains why the patching scan of Appendix~\ref{app:act_patching_extended} finds no within-block rescue site at $t\in\{0,1\}$: by then the trajectory lies on one side of the separatrix and single-cell interventions cannot transport it across.

\section{Discussion and Limitations}\label{sec:discussion}

\paragraph{What the analysis demonstrates.}
Three threads come together in this work. (i) A specific failure pattern---catastrophic fragmentation of reconstructions from sparse on-surface inputs---is identified, quantified, and shown to generalize across two open-weight architectures, two real-world datasets and two samplers. (ii) Targeted causal interventions isolate this failure to a single cross-attention write in the first denoising step, and matched-magnitude controls demonstrate that the committed variable is the directional content of the perturbation drift of the activation in a low-rank subspace. (iii) Interpreted through diffusion-dynamics theory, the localized commit sits immediately upstream of a symmetry-breaking bifurcation of the reverse-time potential, after which exponential separation under a positive Lyapunov exponent transports the trajectory across a basin separatrix. The two layers of description---a circuit-level lever and a trajectory-level bifurcation---are internally consistent, and motivate \texttt{PowerRemap}, a test-time spectral intervention that rescues 84.6--98.3\% of failures across architectures and datasets.

\paragraph{Practical implications.}
Sparse point-cloud reconstruction is non-trivial because the model must infer geometry the input does not directly resolve. Our results show that, under realistic input variation, the generative prior can move catastrophically between attractors. For safety-critical pipelines such as surgical navigation~\citep{Liu2024_neurosurgical} or perception for autonomous driving~\citep{zhang2023perception}, a model that occasionally produces hundreds of disconnected pieces from a near-identical sparse input is unsuitable regardless of average-case fidelity. Two practical takeaways follow. First, scalar correlates of the transition, such as the spectral entropy, provide a cheap online diagnostic that can be computed at test time without ground-truth surfaces. Second, single-site spectral interventions can stabilize a deployed model without retraining, with hyperparameters that cluster tightly across shapes (Appendix~\ref{app:choice_gamma_remedy}).

\paragraph{Why this particular signature?}
A natural follow-on question is why the particular cross-attention write in the first denoising step carries a directional commit for this specific failure. The first denoising step is when the latent is closest to noise and the model leans most heavily on conditioning~\citep{liu2024faster}. The early cross-attention writes are accordingly the highest-bandwidth conduits for conditioning information. Why the directional content of the cross-attention write concentrated in a low-rank subspace becomes primarily responsible for topology specifically is open. We conjecture this reflects an inductive bias acquired during training. The general pattern of diffusion transformers committing to coarse, high-level structure before resolving fine-grained detail has been observed in text-to-image models~\citep{tinaz2025emergenceevolutioninterpretableconcepts}. Our finding is consistent with a 3D analog in which the coarse structure being committed early is topology, and the commitment is encoded as a low-rank direction in conditioning space.

\paragraph{Limitations.}\hfill

\noindent\emph{Architectural and methodological scope.} Our analysis covers two point-cloud-conditioned DDPM-based diffusion transformers, \textsc{WaLa}~\citep{sanghi2024waveletlatentdiffusionwala} and \textsc{Make-A-Shape}~\citep{hui2024makeashapetenmillionscale3dshape}. Transfer to flow-matching backbones~\citep{hunyuan3domni2025, xia2026points}, to image-conditioned 3D generators, and to other latent-diffusion architectures is unverified. The selection of two open-weight large-scale point-cloud-conditioned 3D diffusion transformers reflects the current state of available diffusion transformers rather than methodological completeness.

\noindent\emph{Distributed signatures cannot be ruled out.} Single-cell activation patching identifies a sufficient lever for rescue, but the causal-asymmetry result in Appendix~\ref{app:act_patching_extended} (noising scan) shows that transplanting an unhealthy $\mathbf{Y}_{4,7}$ into an otherwise-healthy run does not by itself induce Meltdown. The committed signal is carried cumulatively by the residual stream, and the cross-attention site is a single-cell \emph{handle}, not a sufficient cause in isolation. Distributed commit signatures across multiple submodules are therefore not excluded by our analysis.

\noindent\emph{Trajectory-level claims rest on representative paths.} The bifurcation analysis in Section~\ref{sec:diffusion-dynamics} uses representative trajectories from a sphere example, and the unimodality tests use a fixed 2D projection. The qualitative picture transfers across shapes and seeds (Appendix~\ref{app:more_datapoints}), but we do not claim a quantitative bifurcation theorem for arbitrary shape spaces.

\noindent\emph{Per-shape calibration for \textsc{Make-A-Shape}.} On \textsc{WaLa}, a single global $\gamma=100$ suffices; on \textsc{Make-A-Shape}, per-shape selection over a small grid is required, using connectivity $C{=}1$ as the criterion. While this criterion is deployable at test time (no ground truth required), it adds inference cost. The tight clustering of effective $\gamma$ around 1.05--1.10 (Appendix~\ref{app:choice_gamma_remedy}) suggests this could be ameliorated by a learned shape-category default.

\noindent\emph{Adversarial nature of the failure.} Meltdown is identified by adversarial search (Algorithm~\ref{alg:alg_adverserial_Meltdown_search}). We show that arbitrary on-surface samplings \emph{can} induce the failure, that the failure is frequent (89.9--100\%) under adversarial search across two corpora, and that it concentrates in the low-areal-density regime (Appendix~\ref{app:density}) characteristic of real sparse-capture pipelines.

\paragraph{Future directions.}
Several extensions follow naturally. Extending the analysis to flow-matching and rectified-flow 3D backbones would test whether the same combined mechanistic--dynamical picture holds beyond DDPM and DDIM. Probing whether the directional signature is detectable at training time could enable preventive interventions during model development rather than test-time rescue. More broadly, the framework demonstrated here---activation patching with matched-magnitude directional controls, paired with a diffusion-dynamics interpretation of the resulting commit---appears to us applicable to other conditional generative settings where catastrophic, hard-to-anticipate mode commitments must be diagnosed and mitigated.

\section{Conclusion}

We have presented a mechanistic case study of a catastrophic failure mode in state-of-the-art point-cloud-conditioned 3D diffusion transformers. Tiny on-surface perturbations to a sparse input cloud can fracture the reconstructed output into hundreds of disconnected pieces---a failure we call \emph{Meltdown}---with prevalence 89.9--100\% under adversarial search across two open-weight architectures (\textsc{WaLa}, \textsc{Make-A-Shape}), two real-world datasets (GSO, SimJEB), and both DDPM and DDIM samplers. Activation patching localizes the failure to a single cross-attention write at the fourth block and first denoising step, and matched-magnitude directional controls demonstrate that the committed variable is the directional content of that activation concentrated in a low-rank subspace of its perturbation drift. Embedded in diffusion-dynamics theory, this localized commit sits immediately upstream of a symmetry-breaking bifurcation of the reverse-time potential. Motivated by this combined picture, we introduced \texttt{PowerRemap}, a test-time spectral control on the localized activation that stabilizes 84.6--98.3\% of failures across architectures and datasets.

Beyond this specific result, our work offers a template for combining mechanistic interpretability with diffusion-dynamics theory to study conditional diffusion transformers. Linking a single-cell cross-attention mechanism to a trajectory-level account of spontaneous symmetry breaking provides a level of explanatory coverage that, in our experience, neither approach achieves alone: the circuit identifies \emph{where} and \emph{with what} the model commits, while the dynamics explain \emph{why} a tiny local perturbation has a discontinuous global effect. We expect this combined methodology to be useful for other conditional generative models and other failure modes, particularly in safety-critical deployment regimes where catastrophic and hard-to-anticipate behaviors must be diagnosed and mitigated before they reach end users.

\clearpage
\bibliography{main}
\bibliographystyle{tmlr}

\clearpage
\appendix
\section{Background}\label{app:models}

In this section, we provide background information on the diffusion transformers \textsc{WaLa} \citep{sanghi2024waveletlatentdiffusionwala} and \textsc{Make-A-Shape} \citep{hui2024makeashapetenmillionscale3dshape} along the dimensions \textit{diffusion} (Appendix \ref{app:diffusion_background}) and \textit{transformer} (Appendix \ref{app:transformer_appendix}).

\subsection{Diffusion}
\label{app:diffusion_background}

\paragraph{Forward transition.} Diffusion generative models synthesize data by inverting a Markov chain that gradually corrupts an observation ${\mathbf x}_0\!\sim\!p_\text{data}$ with Gaussian noise over $T$ discrete timesteps \citep{sohl2015deep,ho2020ddpm}.  
The forward (noising) transition is
\begin{align}
q(\mathbf x_t \mid \mathbf x_{t-1}) &= 
\mathcal N\!\bigl(\mathbf x_t;\sqrt{1-\beta_t}\,\mathbf x_{t-1},\;\beta_t\mathbf I\bigr), \quad
t=1,\dots,T,                                                         \\
q(\mathbf x_t \mid \mathbf x_0)     &= 
\mathcal N\!\bigl(\mathbf x_t;\sqrt{\bar\alpha_t}\,\mathbf x_0,\;(1-\bar\alpha_t)\mathbf I\bigr), 
\quad\text{with}\quad 
\bar\alpha_t\!=\!\prod_{s=1}^{t}(1-\beta_s),                              \label{eq:q_xt_given_x0}
\end{align}
where the variance schedule $\{\beta_t\}_{t=1}^T\!\subset\!(0,1)$ is chosen so that $\mathbf x_T$ is nearly i.i.d. $\mathcal N( 0, I)$. Both architectures are associated with a cosine variance schedule.

\paragraph{Noise prediction objective.}
Instead of directly regressing $\mathbf x_0$, the denoising neural networks $\epsilon_\theta$ of \textsc{WaLa} and \textsc{Make-A-Shape} have been trained to predict the added noise:
\begin{equation}
\mathcal L_\text{simple}(\theta)=
\mathbb E_{t,\mathbf x_0,\boldsymbol\epsilon}\!
\Bigl[\bigl\lVert 
\boldsymbol\epsilon - \epsilon_\theta(\underbrace{\sqrt{\bar\alpha_t}\,\mathbf x_0 + \sqrt{1-\bar\alpha_t}\,\boldsymbol\epsilon}_{\mathbf x_t},\,t)
\bigr\rVert_2^2\Bigr],\qquad 
\boldsymbol\epsilon\sim\mathcal N( 0, I).            \label{eq:noise_objective}
\end{equation}
This "$\epsilon$‐parameterization" empirically stabilizes the training and is adopted by nearly all modern models \citep{ho2020ddpm}.

\paragraph{Denoising (DDPM).}
Given a trained $\epsilon_\theta$, the original Denoising Diffusion Probabilistic Model (DDPM) \citep{ho2020ddpm} samples via the stochastic reverse transition
\begin{align}
p_\theta(\mathbf x_{t-1}\!\mid\!\mathbf x_t)=
\mathcal N\!\bigl(&\mathbf x_{t-1};
\underbrace{\frac{1}{\sqrt{1-\beta_t}}
\Bigl(\mathbf x_t - \frac{\beta_t}{\sqrt{1-\bar\alpha_t}}\,
\epsilon_\theta(\mathbf x_t,t)\Bigr)}_{\boldsymbol\mu_\theta(\mathbf x_t,t)},
\,\sigma_t^2\mathbf I
\bigr),\qquad 
\sigma_t^2=\tilde\beta_t\!,                                   \label{eq:ddpm_reverse}
\end{align}
where $\tilde\beta_t=\beta_t\frac{1-\bar\alpha_{t-1}}{1-\bar\alpha_t}$.  
Iterating Eq.~\eqref{eq:ddpm_reverse} from $t\!=\!T$ to $1$ produces $\mathbf x_0$ in $T$ noisy steps.

\paragraph{Denoising (DDIM).}
\citet{song2021ddim} showed that the same model admits a \emph{deterministic} implicit sampler (DDIM) obtained by setting the variance term to zero:  
\begin{equation}
\mathbf x_{t-1} = 
\sqrt{\bar\alpha_{t-1}}\,
\underbrace{\Bigl(\frac{\mathbf x_t - \sqrt{1-\bar\alpha_t}\,\epsilon_\theta(\mathbf x_t,t)}
{\sqrt{\bar\alpha_t}}\Bigr)}_{\hat{\mathbf x}_0}
+\sqrt{1-\bar\alpha_{t-1}}\,
\epsilon_\theta(\mathbf x_t,t).\label{eq:ddim_update}
\end{equation}
Eq.~\eqref{eq:ddim_update} preserves the marginal $q(\mathbf x_{t-1}\mid\mathbf x_0)$, enabling user‐specified inference schedules (e.g., $t\!=\!T,\!\dots,\!1$ with $T\!\gg\!1$ for high fidelity or sparse subsets for speed) without retraining. Crucially, Eqs.~\eqref{eq:ddpm_reverse}--\eqref{eq:ddim_update} share the same $\epsilon_\theta$ trained via Eq.~\eqref{eq:noise_objective}. Hence one can \emph{train} with the log‐likelihood–consistent DDPM objective but sample using DDPM or DDIM.

\paragraph{Classifier-free guidance.} Both architectures employ classifier-free guidance (CFG) \citep{ho2022classifierfreediffusionguidance}.
CFG biases the denoising direction toward a user condition without requiring an external classifier. For a current latent $\mathbf x_t$ and the shared noise predictor $\epsilon_\theta$, we obtain two estimates at the same step $t$: the unconditional prediction $\epsilon_{\!\text{uncond}}$ (with the condition omitted) and the conditional prediction $\epsilon_{\!\text{cond}}$ (under the desired condition). We then form a guided estimate
\[
\tilde{\epsilon}
=\epsilon_{\!\text{uncond}}
+s\bigl(\epsilon_{\!\text{cond}}-\epsilon_{\!\text{uncond}}\bigr),\qquad s\ge 0,
\]
and substitute $\tilde{\epsilon}$ in place of $\epsilon_\theta(\mathbf x_t,t)$ in the DDPM/DDIM updates (Eqs.~\eqref{eq:ddpm_reverse}--\eqref{eq:ddim_update}). Setting $s=1$ at inference-time ignores updates from the unconditional stream, i.e.,  $\tilde{\epsilon}=\epsilon_{\!\text{cond}}$.

\subsection{Transformer} \label{app:transformer_appendix}

The noise–prediction objective~\eqref{eq:noise_objective} only specifies what to learn but leaves open how the denoiser~$\epsilon_\theta$ is parameterised.  
Classical DDPMs adopt a convolutional U-Net encoder–decoder \citep{ronneberger2015unet,ho2020ddpm}, whereas modern large‑scale models \citep{rombach2022highresolutionimagesynthesislatent, bao2023uvit,karras2023edm} replace the convolutional blocks with \emph{transformer} layers, yielding the U-ViT (U-shaped Vision Transformer) backbone. Both architectures, \textsc{WaLa} and \textsc{Make-A-Shape}, implement a diffusion generative model via transformer layers.

\subsubsection{Overview} \label{app:overview_transformer_wala_mas}
Both methods adopt a wavelet–latent diffusion pipeline in which 3D shapes are represented as multiscale wavelet coefficients and a U‑ViT‑style denoising backbone \citep{hoogeboom2023} is trained in the DDPM \citep{ho2020denoisingdiffusionprobabilisticmodels} framework.  The key difference lies in how the wavelet data are fed to the diffusion core.

\begin{enumerate}
    \item \textsc{WALA} first compresses the full wavelet tree with a convolutional VQ‑VAE (stage 1), mapping the diffusible wavelet tree to a latent grid. The latent grid is then modeled by a 32-layer U-ViT (stage 2), where each transformer layer runs self-attention and cross-attention, totaling 32 cross-attention calls.
    \item \textsc{Make-A-Shape} skips the auto‑encoder and instead packs selected wavelet coefficients into a compact grid. The U‑ViT backbone then downsamples this tensor to a bottleneck volume. The bottleneck is traversed by a 16-layer U-ViT core—8 self-attention layers immediately followed by 8 cross-attention layers— before up-sampling restores the packed grid.
\end{enumerate}

\subsubsection{Conditioning pathway (point-cloud)}

In general, both \textsc{Make-A-Shape} and \textsc{WaLa} share a common pipeline for conditioning on point clouds: a PointNet \citep{qi2017pointnetdeeplearningpoint} encoding followed by aggregation and injecting the resulting latent vectors into the U-ViT generator via ~\textit{(i)} affine modulation of normalization layers and ~\textit{(ii)} cross-attention.\\\\
In particular, \textsc{Make-A-Shape} injects the conditioning latent vectors into the U-ViT generator at three stages: ~(1) \textit{concatenation}: the latent vectors are aggregated and concatenated as additional channels of the input noise coefficients, ~(2) \textit{affine modulation}: the latent vectors are aggregated and subsequently utilized to condition the convolution (down-sampling) and de-convolution (up-sampling) layers via modulating the affine parameters of the group normalization layers, ~(3) \textit{cross-attention}: each condition latent vector is augmented with an element-wise positional encoding and then fed into a cross-attention module alongside the bottleneck volume.\\\\
\textsc{WaLa} injects the conditioning latent vectors into the U-ViT generator at two stages: ~(1) \textit{affine modulation}: the latent vectors are linearly projected via a global projection network and used to modulate the scale and bias parameters of GroupNorm layers in both the ResNet and attention blocks (AdaGN) \citep{esser2024}, ~(2) \textit{cross-attention}: each latent vector, augmented with an element-wise positional encoding, is employed as the key and value in cross-attention modules interleaved within each transformer block.

\clearpage
\section{Experiments}
\label{app:experiments}

This section shows that the observations and insights gained through studying the diffusion transformer \textsc{WaLa} \citep{sanghi2024waveletlatentdiffusionwala} under DDIM sampling on spheres robustly transfer (i) to the diffusion transformer \textsc{Make-A-Shape} \citep{hui2024makeashapetenmillionscale3dshape} (ii) sampling under DDPM and (iii) other shapes (GSO and SimJEB). Additionally, this section details the experimental setup to reproduce our results and empirically investigates further variables. In particular:

\begin{enumerate}
    \item \textbf{General} (\ref{app:experiments_general}): This section provides an overview on our experimental setup.
    \item \textbf{Sphere Experiments} (\ref{app:experiments_spheres}):
    \begin{enumerate}
        \item \textbf{General} (\ref{app:experiments_spheres_general}: This section provides an overview on the setup for the sphere experiments.
        \item \textbf{WaLa, DDIM} (\ref{app:experiments_sphere_wala_ddim}): This section reports the experimental setup for the sphere experiments in the main text.
        \item \textbf{WaLa, DDPM} (\ref{app:experiments_wala_ddpm}): This section reports additional results for \textsc{WaLa} under DDPM sampling.
        \item \textbf{Make-A-Shape, DDIM} (\ref{app:experiments_sphere_mas_ddim}): This section provides results for \textsc{Make-A-Shape} under DDIM sampling.
        \item \textbf{Make-A-Shape, DDPM} \ref{app:experiments_sphere_mas_ddpm}: This section provides results for \textsc{Make-A-Shape} under DDPM sampling.
    \end{enumerate}
    \item \textbf{Datasets} (\ref{app:gso}): This section details our experiments on GSO \citep{downs2022googlescannedobjectshighquality} and SimJEB \citep{Whalen_2021}.
    \item \textbf{Density Study} (\ref{app:density}): This section examines how the prevalence of Meltdown depends on the sparsity of the input point cloud.
    \item \textbf{Extended Activation Patching} (\ref{app:act_patching_extended}): This section provides activation-patching results on additional components beyond cross-attention.
    \item \textbf{More Datapoints} (\ref{app:more_datapoints}): This section provides further evidence that the  patterns observed in Section \ref{sec:activation_patching}-\ref{sec:patching_effect} generalize when evaluated on  more data points and random seeds.
    \item \textbf{Additional Spectral Metrics} (\ref{app:more_spectral_metrics}): This section assesses additional spectral metrics as potential indicators of Meltdown.
    \item \textbf{Multiple Objects} (\ref{app:mult_objects}): This section examines whether the Meltdown phenomenon and the effectiveness of \texttt{PowerRemap} extend beyond single-object inputs.
    \item \textbf{Examining \texttt{PowerRemap} strength} (\ref{app:choice_gamma_remedy}): This section empirically investigates the influence of the \texttt{PowerRemap} Strength $\gamma$ on reconstruction connectivity.
    \item \textbf{\texttt{PowerRemap} on Non-Meltdown Cases} (\ref{app:non_meltdown_cases}): This section empirically verifies that \texttt{PowerRemap} does not interfere with non-Meltdown cases.
\end{enumerate}

\subsection{General}\label{app:experiments_general}
This section provides a general overview on the experimental setup for all results reported in this work.

\paragraph{Restrict analysis to 
conditional stream.}
As the failure behavior, Meltdown, is independent of the unconditional stream, we exclusively investigate the conditional prediction stream. That is, we set the CFG scale $s=1.0$ and restrict our mechanistic analysis (e.g., activations) and diffusion dynamics analysis (e.g., latents) to the conditional stream.

\paragraph{Seeding.}
Randomness regarding a diffusion trajectory is controlled globally by seeding Python, NumPy, and PyTorch (\texttt{torch.backends.cudnn.deterministic=True}, \texttt{benchmark=False}) so that every evaluation at a given $\rho$ starts from the same terminal noise $x_T$.

\subsection{Sphere Experiments}
\label{app:experiments_spheres}

This section (i) provides a detailed account on our setup for the sphere experiments and (ii) reports additional results for Meltdown on \textsc{Make-A-Shape} and DDPM sampling.

\subsubsection{Setup}\label{app:experiments_spheres_general}

We detail the minimal, fully reproducible setup used to produce the sphere experiments for \textsc{WaLa} and \textsc{Make-A-Shape}. Throughout, the control parameter is $\rho\in[0,1]$, and the phenomenological order parameter is the number of connected components $C(\rho)$ in the generated mesh.

\paragraph{Conditioning clouds on the sphere.}
We work on $\mathcal{S}=\{x:\|x\|_2=1\}$ and fix $N=400$ points for \textsc{WaLa} and $N=1200$ for \textsc{Make-A-Shape}. The base cloud $\mathcal{P}(0)$ uses a golden-angle (Fibonacci) sphere distribution:
\[
g=\pi(3-\sqrt{5}),\quad
i\in\{0,\dots,N-1\},\quad
y_i=1-\frac{2(i+0.5)}{N},\quad
r_i=\sqrt{1-y_i^2},\quad
\theta_i=g\,i,
\]
\[
p_i(0)=\bigl(r_i\cos\theta_i,~y_i,~r_i\sin\theta_i\bigr).
\]
A second target cloud $\mathcal{P}(1)$ is produced by jittering each $p_i(0)$ with i.i.d.\ Gaussian noise $n_i\sim\mathcal{N}(0,0.1^2\mathbf{I}_3)$ and renormalizing to the unit sphere:
\[
\tilde{p}_i = \frac{p_i(0)+n_i}{\|p_i(0)+n_i\|_2},\qquad \mathcal{P}(1)=\{\tilde{p}_i\}_{i=1}^N.
\]
We then move each point \emph{along} the surface via per-point spherical linear interpolation (SLERP) between corresponding pairs:
\[
p_i(\rho)=\operatorname{slerp}\bigl(p_i(0),\tilde{p}_i;\rho\bigr)
= \frac{\sin((1-\rho)\omega_i)}{\sin\omega_i}\,p_i(0) + \frac{\sin(\rho\,\omega_i)}{\sin\omega_i}\,\tilde{p}_i,\quad
\omega_i=\arccos\!\bigl(\langle p_i(0),\tilde{p}_i\rangle\bigr).
\]
This yields the cloud path $\mathcal{P}(\rho)=\{p_i(\rho)\}_{i=1}^N$ used throughout.

\vspace{4pt}
\paragraph{Decoding and component counting.}
Given $\mathcal{P}(\rho)$, we compute a conditioning code via the model’s encoder and sample a latent with the diffusion sampler to yield $G(\mathcal{P}(\rho))$, i.e. a \texttt{mesh}. We report
\[
C(\rho)=\texttt{len}(\texttt{trimesh.split(mesh)}),
\]
i.e., the number of connected components in \texttt{trimesh}.

\paragraph{Grid over the control parameter.}
We sweep a uniform grid of $\rho$ values, i.e., $\rho\in\{0,0.05,\dots,1.0\}$.

\vspace{6pt}
\paragraph{Connectivity curve $C(\rho)$.}
We evaluate $C(\rho)$ on the uniform $\rho$ grid. For each $\rho$, we reseed the RNGs to reproduce the identical terminal noise $x_T$. The curve reported is the set
\[
\{(\rho,\,C(\rho))\}_{\rho\in\{0,0.05,\dots,1.0\}},
\]
from which the observed plateau at $C(\rho)=1$ and the subsequent jump to $C(\rho)>1$ over a narrow $\rho$-interval (a connectivity bifurcation) are directly obtained.

\paragraph{Spectral entropy curve $H(\rho)$.}
We evaluate the spectral entropy of the localized cross–attention write on the same uniform control grid $\rho\in\{0,0.05,\dots,1.0\}$ and with identical terminal noise across $\rho$.

For each $\rho$, we encode the cloud $\mathcal{P}(\rho)$, run a single sampling trace, and read out the token-wise cross-attention write at the chosen site, $\mathbf{Y}(\rho)$.
Let $\{\sigma_i(\rho)\}_i$ be the singular values of $\mathbf{Y}(\rho)$ (SVD of the matrix with shape tokens $\times$ features). We form normalized directional energies
\[
p_i(\rho)\;=\;\frac{\sigma_i(\rho)^2}{\sum_j \sigma_j(\rho)^2},
\qquad
H(\rho)\;=\;-\sum_i p_i(\rho)\,\log p_i(\rho),
\]
using the natural logarithm. The reported curve is the set
\[
\bigl\{\,(\rho,\;H(\rho))\,\bigr\}_{\rho\in\{0,0.05,\dots,1.0\}}.
\]

\subsubsection{\textsc{WaLa}, DDIM}
\label{app:experiments_sphere_wala_ddim}

\vspace{6pt}
\paragraph{Key hyperparameters.}
\begin{center}
\begin{tabular}{l l}
Model & \texttt{ADSKAILab/WaLa-PC-1B} \\
Sampler & DDIM ($\eta=0$) \\
Diffusion rescale & $8$ steps (\texttt{diffusion\_rescale\_timestep=8}), i.e., default \\
CFG weight & $1.0$ (\texttt{scale=1.0}), i.e., we consider only conditional stream \\
Points per cloud & $N=400$ \\
Cloud source & Unit sphere, golden-angle placement \\
Target cloud & Gaussian jitter $\sigma=0.1$ on $\mathbb{R}^3$, renormalize to $\mathbb{S}^2$ \\
Interpolation & Per-point SLERP, control $\rho\in[0,1]$ \\
$\rho$ grid & $21$ values: $0,0.05,\dots,1.0$ \\
Seeds & $0-1000$ for all RNG calls \\
Order parameter & $C(\rho)=\#$ connected components (\texttt{trimesh.split}) \\
Device & \texttt{cuda} (CPU is functionally equivalent but slower)
\end{tabular}
\end{center}

\newpage
\subsubsection{\textsc{WaLa}, DDPM}
\label{app:experiments_wala_ddpm}

\begin{figure*}[ht]
    \centering
    \begin{minipage}[b]{0.48\linewidth}
        \centering
        \includegraphics[width=\linewidth]{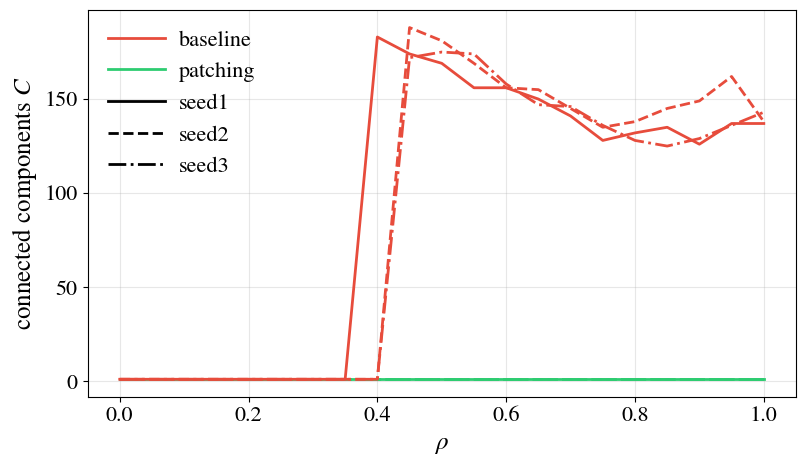}
        \vspace{-5pt} 
        \centerline{\small (a) Connected components $C$ vs. $\rho$}
    \end{minipage}
    \hfill 
    \begin{minipage}[b]{0.48\linewidth}
        \centering
        \includegraphics[width=\linewidth]{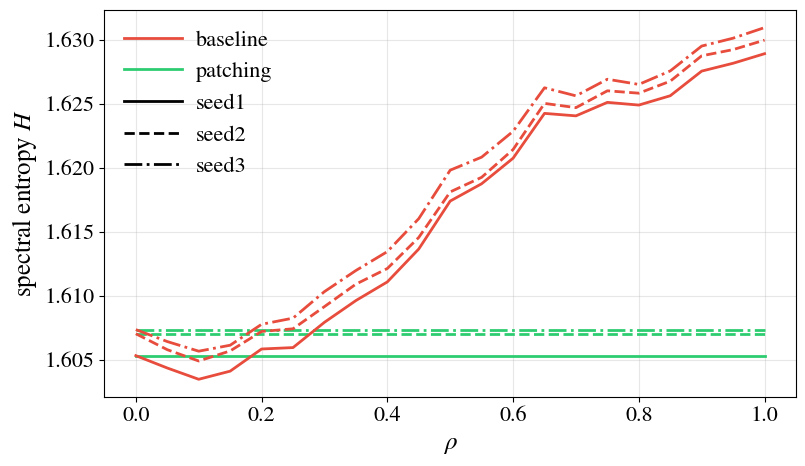}
        \vspace{-5pt} 
        \centerline{\small (b) Spectral entropy $H$ vs. $\rho$}
    \end{minipage}
    
    \vspace{10pt} 
    \caption{[\textsc{WaLa}, DDPM]. Our results from \textsc{WaLa} under DDIM sampling transfer to \textsc{WaLa} under DDPM sampling.}
    \label{fig:connectivity_spectral_entropy_wala_ddpm}
\end{figure*}

The activation-patching grid for \textsc{WaLa} under DDPM is equivalent to Figure \ref{fig:activation_patching_heatmap}, i.e., \textsc{WaLa} under DDIM. The corresponding curves can be found in Figure \ref{fig:connectivity_spectral_entropy_wala_ddpm}.

\vspace{6pt}
\paragraph{Key hyperparameters.}
\begin{center}
\begin{tabular}{l l}
Model & \texttt{ADSKAILab/WaLa-PC-1B} \\
Sampler & DDPM \\
Diffusion rescale & $8$ steps (\texttt{diffusion\_rescale\_timestep=8}) \\
CFG weight & $1.0$ (\texttt{scale=1.0}) (we consider only conditional stream) \\
Points per cloud & $N=400$ \\
Cloud source & Unit sphere, golden-angle placement \\
Target cloud & Gaussian jitter $\sigma=0.1$ on $\mathbb{R}^3$, renormalize to $\mathbb{S}^2$ \\
Interpolation & Per-point SLERP, control $\rho\in[0,1]$ \\
$\rho$ grid & $21$ values: $0,0.05,\dots,1.0$ \\
Seeds & 0 for all RNG calls \\
Order parameter & $C(\rho)=\#$ connected components (\texttt{trimesh.split}) \\
Device & \texttt{cuda} (CPU is functionally equivalent but slower)
\end{tabular}
\end{center}

\newpage
\subsubsection{\textsc{Make-A-Shape}, DDIM}
\label{app:experiments_sphere_mas_ddim}

\begin{figure}[h!]
    \centering
    \includegraphics[width=0.5\linewidth]{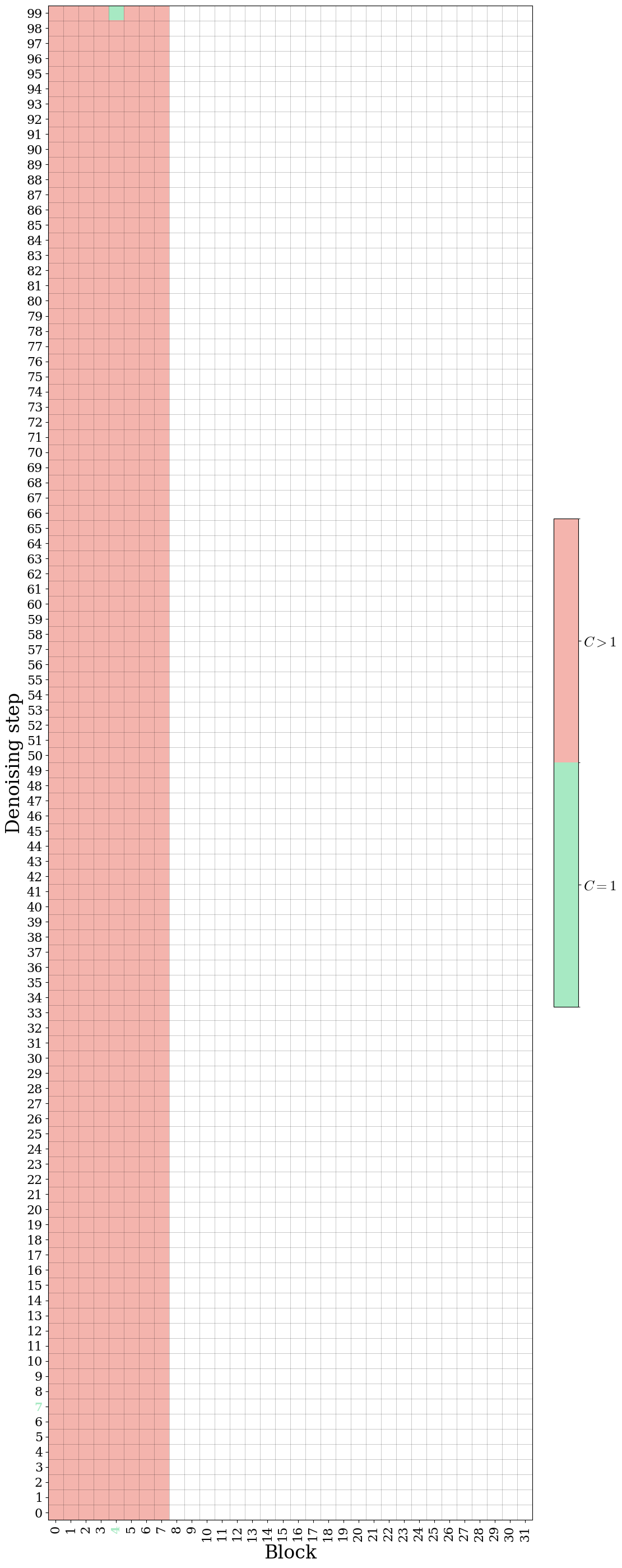}
    \caption{Activation-patching result for \textsc{Make-A-Shape}. Analogous to our result for \textsc{WaLa}, we find an early denoising cross-attention activation that controls Meltdown behavior.}
\label{fig:heatmap_activation_patching_mas}
\end{figure}

\begin{figure*}[h!]
    \centering
    \begin{minipage}[b]{0.48\linewidth}
        \centering
        \includegraphics[width=\linewidth]{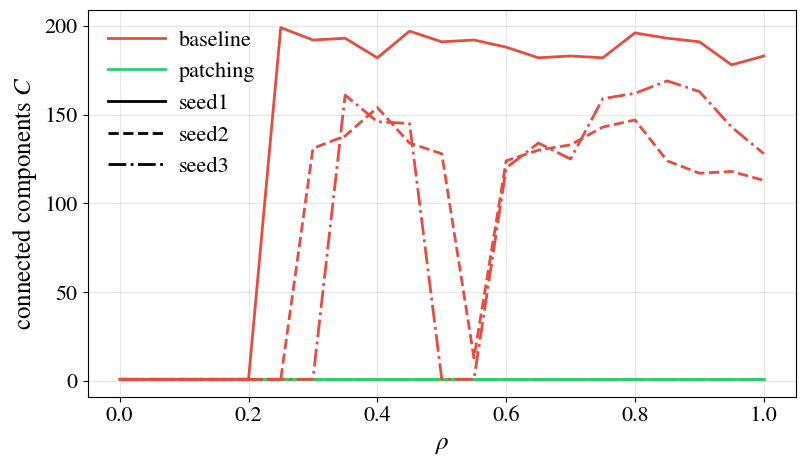}
        \vspace{-5pt} 
        \centerline{\small (a) Connected components $C$ vs. $\rho$}
    \end{minipage}
    \hfill 
    \begin{minipage}[b]{0.48\linewidth}
        \centering
        \includegraphics[width=\linewidth]{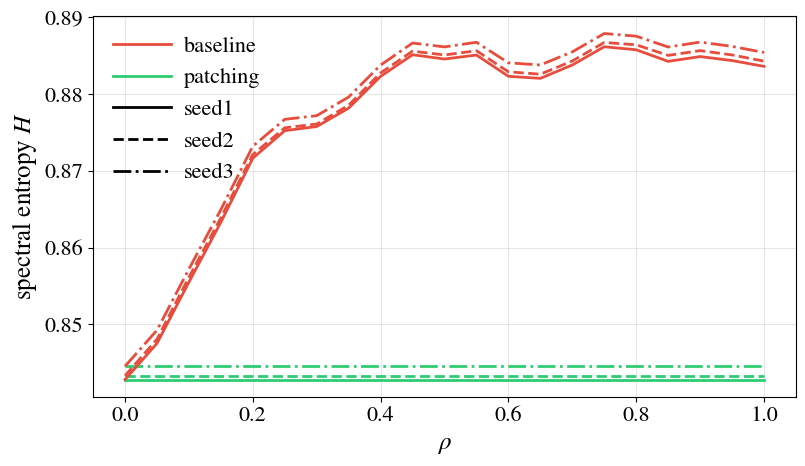}
        \vspace{-5pt} 
        \centerline{\small (b) Spectral entropy $H$ vs. $\rho$}
    \end{minipage}
    
    \vspace{10pt} 
    \caption{[\textsc{Make-A-Shape}, DDIM]. Our results from \textsc{WaLa} transfer to \textsc{Make-A-Shape}: As we move from a healthy to an unhealthy run, we observe that the \textcolor[HTML]{e74c3c}{baseline} case shows a smooth rise in spectral entropy and a sudden jump in connectivity. \textcolor[HTML]{2ecc71}{Patching} our $\mathbf{Y}$ keeps the spectral entropy at healthy levels and preserves connectivity. This behavior is consistent across diffusion seeds.}
    \label{fig:connectivity_spectral_entropy_mas_ddim}
\end{figure*}

We report the result for the activation search procedure for \textsc{Make-A-Shape} under DDIM in Figure \ref{fig:heatmap_activation_patching_mas}. The corresponding curves are depicted in Figure \ref{fig:connectivity_spectral_entropy_mas_ddim}. 

\vspace{6pt}
\paragraph{Key hyperparameters.}
\begin{center}
\begin{tabular}{l l}
Model & \texttt{ADSKAILab/Make-A-Shape-point-cloud-20m} \\
Sampler & DDIM \\
Diffusion rescale & $100$ steps (\texttt{diffusion\_rescale\_timestep=8}), i.e., default \\
CFG weight & $1.0$ (\texttt{scale=1.0}), i.e.,  we consider only conditional stream \\
Points per cloud & $N=1200$ \\
Cloud source & Unit sphere, golden-angle placement \\
Target cloud & Gaussian jitter $\sigma=0.1$ on $\mathbb{R}^3$, renormalize to $\mathbb{S}^2$ \\
Interpolation & Per-point SLERP, control $\rho\in[0,1]$ \\
$\rho$ grid & $21$ values: $0,0.05,\dots,1.0$ \\
Seeds & 0 for all RNG calls \\
Order parameter & $C(\rho)=\#$ connected components (\texttt{trimesh.split}) \\
Device & \texttt{cuda} (CPU is functionally equivalent but slower)
\end{tabular}
\end{center}

\newpage
\subsubsection{\textsc{Make-A-Shape}, DDPM}
\label{app:experiments_sphere_mas_ddpm}

\begin{figure*}[h!]
    \centering
    \begin{minipage}[b]{0.48\linewidth}
        \centering
        \includegraphics[width=\linewidth]{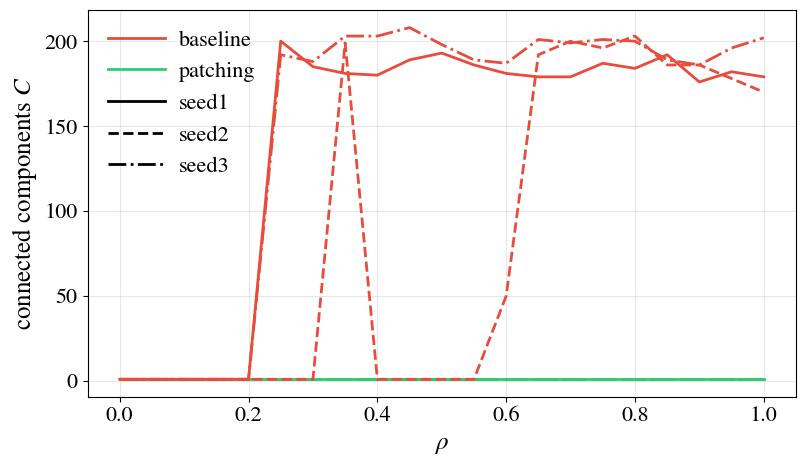}
        \vspace{-5pt} 
        \centerline{\small (a) Connected components $C$ vs. $\rho$}
    \end{minipage}
    \hfill 
    \begin{minipage}[b]{0.48\linewidth}
        \centering
        \includegraphics[width=\linewidth]{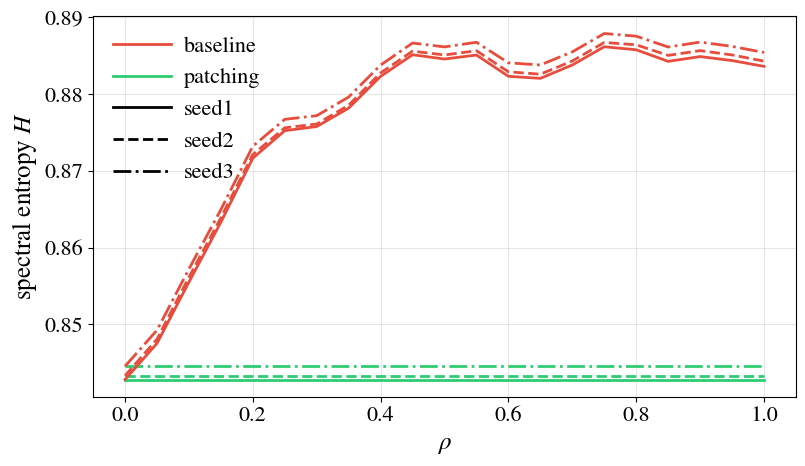}
        \vspace{-5pt} 
        \centerline{\small (b) Spectral entropy $H$ vs. $\rho$}
    \end{minipage}
    
    \vspace{10pt} 
    \caption{[\textsc{Make-A-Shape}, DDPM]. Our results from [\textsc{Make-A-Shape} under DDIM sampling transfer to \textsc{WaLa} under DDPM sampling.}
    \label{fig:connectivity_spectral_entropy_mas_ddpm}
\end{figure*}

The activation-patching grid for \textsc{Make-A-Shape} under DDPM is equivalent to Figure \ref{fig:heatmap_activation_patching_mas}, i.e., \textsc{Make-A-Shape} under DDIM. The corresponding curves can be found in Figure \ref{fig:connectivity_spectral_entropy_mas_ddpm}.

\vspace{6pt}
\paragraph{Key hyperparameters.}
\begin{center}
\begin{tabular}{l l}
Model & \texttt{ADSKAILab/Make-A-Shape-point-cloud-20m} \\
Sampler & DDPM \\
Diffusion rescale & $100$ steps (\texttt{diffusion\_rescale\_timestep=8}), i.e., default \\
CFG weight & $1.0$ (\texttt{scale=1.0}) (we consider only conditional stream) \\
Points per cloud & $N=1200$ \\
Cloud source & Unit sphere, golden-angle placement \\
Target cloud & Gaussian jitter $\sigma=0.1$ on $\mathbb{R}^3$, renormalize to $\mathbb{S}^2$ \\
Interpolation & Per-point SLERP, control $\rho\in[0,1]$ \\
$\rho$ grid & $21$ values: $0,0.05,\dots,1.0$ \\
Seeds & 0 for all RNG calls \\
Order parameter & $C(\rho)=\#$ connected components (\texttt{trimesh.split}) \\
Device & \texttt{cuda} (CPU is functionally equivalent but slower)
\end{tabular}
\end{center}

\clearpage
\newpage
\subsection{Dataset Evaluation}\label{app:gso}

This section provides (i) evidence that Meltdown exists across a variety of shapes, i.e., across the GSO \citep{downs2022googlescannedobjectshighquality} and SimJEB \citep{Whalen_2021} corpora,  and diffusion transformers, i.e., \textsc{WaLa} and \textsc{Make-A-Shape}. Furthermore it details our setup to evaluate our method \texttt{PowerRemap} on GSO and SimJEB as well as the results of these evaluations.

\paragraph{General.} We evaluate \texttt{PowerRemap} on the \textsc{WaLa} and \textsc{Make-A-Shape} architectures, using DDIM sampling \citep{song2021ddim}. For each object, we load the corresponding mesh as the ground‑truth surface $\mathcal{S}\subset\mathbb{R}^3$, center it and scale it to the unit cube.

\paragraph{Find Meltdown.}
We reuse the notation of \S\ref{sec:Meltdown}. Given a mesh $\mathcal{S}$ and generator $G$, we first determine a sparse point budget by searching the smallest $N$ over a grid for which a Poisson‑disk sample $A=\{a_i\}_{i=1}^N\!\subset\!\mathcal{S}$ yields a healthy output $C(G(A))=1$. We then define a surface‑constrained Meltdown path by jittering $A$ and projecting back to $\mathcal{S}$ to obtain $B=\{b_i\}_{i=1}^N$, and interpolate on‑manifold
\[
\mathcal{P}_\rho\;=\;\Pi_{\mathcal{S}}\!\big((1-\rho)A+\rho B\big),\quad \rho\in[0,1],
\]
where $\Pi_{\mathcal{S}}$ is nearest‑point projection. With a fixed random seed (reseeded before every inference), we sweep $\rho$ on a geometric grid to bracket a jump in connectivity, then refine by bisection to the smallest $\varepsilon$ such that $C\!\left(G(\mathcal{P}_\varepsilon)\right) \gg1$.

\begin{algorithm}[h!]
\caption{Adversarial Meltdown Search}
\label{alg:alg_adverserial_Meltdown_search}
\begin{algorithmic}[1]
    \Require surface $\mathcal{S}$, generator $G$, component counter $C(\cdot)$, seed $s=0$
    \State normalize $\mathcal{S}$; find smallest $N$ s.t. $A\!\sim\!\text{Poisson}(\mathcal{S},N)$ gives $C(G(A)){=}1$ (reseed $s$)
    \State $B\leftarrow \Pi_{\mathcal{S}}(A+\xi)$ \Comment{jitter \& project}
    \State sweep $\rho$ on a geometric grid; find $\rho_{\mathrm{lo}}{<}\rho_{\mathrm{hi}}$ with $C(\rho_{\mathrm{lo}}){=}1$, $C(\rho_{\mathrm{hi}}){>}1$
    \State $\varepsilon\leftarrow$ bisection$(\rho_{\mathrm{lo}},\rho_{\mathrm{hi}})$ with reseeding to $s$
    \State \textbf{Return} $(N,\varepsilon,\;C_0{=}C(G(A)),\;C_\varepsilon{=}C(G(\mathcal{P}_\varepsilon)))$
\end{algorithmic}
\end{algorithm}

\paragraph{Evaluate \texttt{PowerRemap}.} 
The task of reconstructing a global surface from a sparse point cloud has only two possible outcomes: success or failure. Thus, we assess the effectiveness \texttt{PowerRemap} by counting the number of times it succeeded in reducing $C_\varepsilon$ to 1, i.e., turning a speckle into a shape. Hence, we treat each shape as a Bernoulli trial under our adversarial search (Algorithm \ref{alg:alg_adverserial_Meltdown_search}). Each trial has an outcome  $p\in \{0,1\}$, where $p=1$ iff the reconstruction meets the criterion $C_\varepsilon=1$; otherwise $p=0$. We first identify baseline failures as those with $C_{0,\text{baseline}}=1$ and $C_{\varepsilon,\text{baseline}}>1$. We then apply \texttt{PowerRemap} only to these failures and count a remedy when $C_{\varepsilon,\texttt{PowerRemap}}=1$.

\subsubsection{GSO}

This section provides a quantitative  evaluation of the Meltdown phenomenon on the  GSO dataset~\citep{downs2022googlescannedobjectshighquality} and assesses the effectiveness of \texttt{PowerRemap} as a mitigation strategy. GSO \citep{downs2022googlescannedobjectshighquality} is a diverse corpus of 1,030 scanned household objects and was \emph{not} used to train either \textsc{WaLa} or \textsc{Make-a-Shape}. All results in this section are obtained by applying the protocol described in Appendix~\ref{app:gso} to the GSO dataset.

\paragraph{Evaluate \texttt{PowerRemap}.} For \textsc{WaLa}, we found Meltdown in 926/1,030 ($89.9\%$) shapes. Our method \texttt{PowerRemap} remedies failure in 910/926 (98.3\%) for  $\gamma=100$. Table \ref{tab:gso_pr_category_top5_other} (top) depicts the performance of our method across all shape categories. For \textsc{Make-A-Shape}, we consider a category-representative subset of 130 GSO shapes and find Meltdown in 130/130 ($100\%$) shapes. We evaluate \texttt{PowerRemap} over a $\gamma$-grid for the given subset, where 
\[
\gamma \in \{1.05, 1.1, 1.15, 1.2, 1.25, 1.3, 1.35, 1.4, 1.5, 2\}.
\]
For 110 out of 130 shapes ($84.6\%$), we find at least one value of $\gamma$ in this range that successfully remedies Meltdown. Among the rescued cases, the median effective value of $\gamma$ is $1.10$ with a standard deviation of $0.13$.  Table \ref{tab:gso_pr_category_top5_MAS_main} (top) depicts the performance of our method across the representative subset. In Appendix \ref{app:choice_gamma_remedy}, we empirically study the influence of the \texttt{PowerRemap} strength $\gamma$ on reconstruction connectivity, concluding that the optimal hyperparameter is model-dependent.

\clearpage
\subsubsection{SimJEB}\label{app:simjeb}

This section provides a quantitative  evaluation of the Meltdown phenomenon on the SimJEB dataset ~\citep{Whalen_2021} and assesses the effectiveness of \texttt{PowerRemap} as a mitigation strategy. SimJEB is a curated benchmark of 381 3D jet-engine bracket CAD models that was \emph{not} included in the training data of either \textsc{WaLa} or \textsc{Make-a-Shape}. All results in this section are obtained by applying the protocol described in Appendix~\ref{app:gso} to the SimJEB dataset.

\paragraph{Evaluate \texttt{PowerRemap}}

For \textsc{WaLa}, we found Meltdown in 352 out of 381 shapes ($92.4\%$). Our method \texttt{PowerRemap} remedies failure in 344/352 (97.7\%) for  $\gamma=100$. Table \ref{tab:gso_pr_category_top5_other} (bottom) depicts the performance of our method across all shape categories. For \textsc{Make-A-Shape}, we consider a category-representative subset of 30 SimJEB shapes and find Meltdown in 30/30 ($100\%$) shapes. We evaluate \texttt{PowerRemap} over a $\gamma$-grid for the given subset, where 
\[
\gamma \in \{1.05, 1.1, 1.15, 1.2, 1.25, 1.3, 1.35, 1.4, 1.5, 2\}.
\]
For 25 out of 30 shapes ($83.3\%$), we find at least one value of $\gamma$ in this range that successfully remedies Meltdown.
Across the rescued cases, the median effective $\gamma$ is 1.05 with a standard deviation 0.063.  Table \ref{tab:gso_pr_category_top5_MAS_main} (bottom) depicts the performance of our method across the representative subset. In Appendix \ref{app:choice_gamma_remedy}, we empirically study the influence of the \texttt{PowerRemap} strength $\gamma$ on reconstruction connectivity, concluding that the optimal hyperparameter is model-dependent.



\clearpage

\subsection{Density}\label{app:density}

In this section, we examine how the prevalence of Meltdown depends on the sparsity of the input point cloud. We quantify the sparsity of an input point cloud of size $N$ by its \emph{areal density}
\begin{equation}
    \eta = \frac{N}{A_{\mathcal{S}}},
\end{equation}
where $A_{\mathcal{S}}$ denotes the surface area of the underlying surface $\mathcal{S}$. Higher values of $\eta$ correspond to denser samplings of $\mathcal{S}$.

\begin{figure}[h!]
    \centering
    \includegraphics[width=0.6\linewidth]{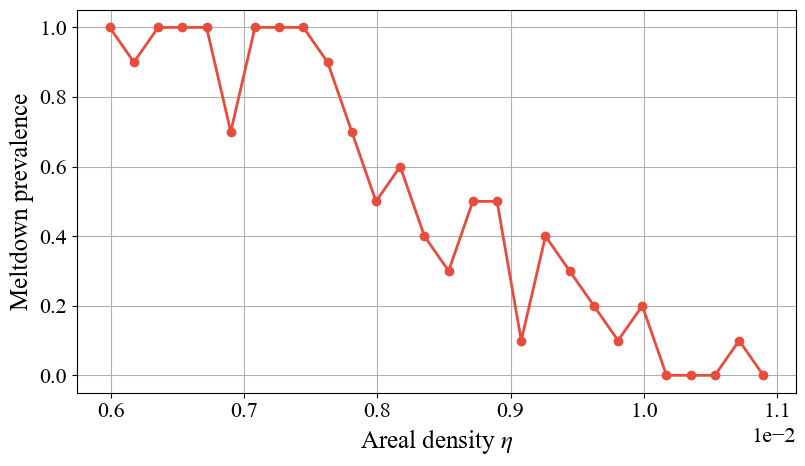}
    \caption{Incidence of Meltdown on SimJEB shape~492 as a function of areal density $\eta$. Meltdown events are especially common for low areal densities, underscoring the difficulty of robust surface reconstruction from sparse point clouds.}
    \label{fig:areal_density}
\end{figure}

In Figure~\ref{fig:areal_density}, we examine how the prevalence of Meltdown depends on $\eta$ for SimJEB shape~492. For each target areal density, we run Algorithm~\ref{alg:alg_adverserial_Meltdown_search} (parameterized by $\eta$) for 10 independent trials and record how often a Meltdown configuration is identified. We observe that Meltdown is particularly frequent in the low-$\eta$ regime.

\clearpage

\subsection{More Datapoints}\label{app:more_datapoints}

\begin{figure*}[t]
    \centering
    \begin{minipage}[b]{0.48\linewidth}
        \centering
        \includegraphics[width=\linewidth]{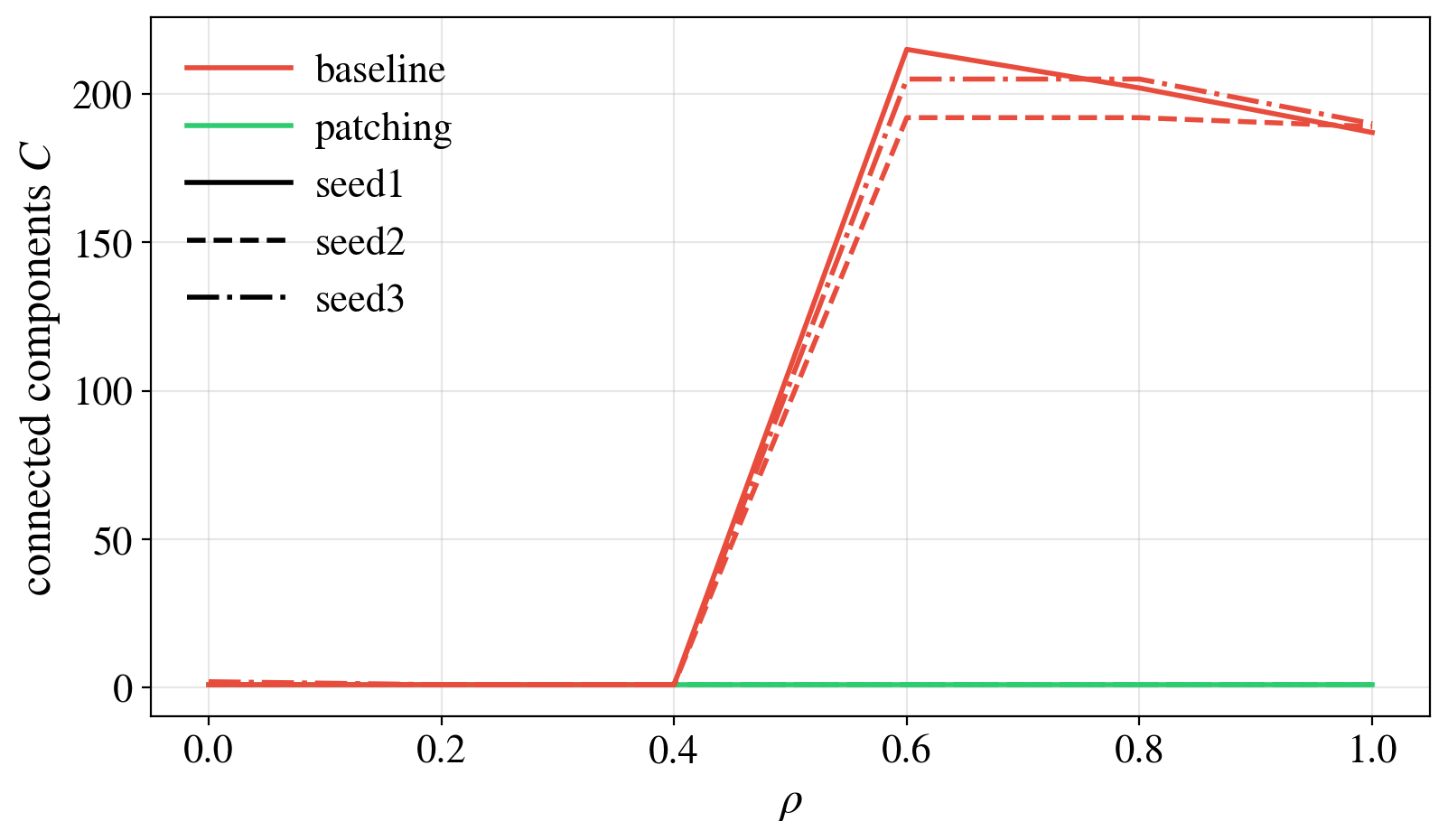}
        \vspace{-5pt}
        \centerline{\small (a) Shape 1 of Figure \ref{fig:your_qualitative_panel_gso_wala} (GSO)}
    \end{minipage}
    \hfill
    \begin{minipage}[b]{0.48\linewidth}
        \centering
        \includegraphics[width=\linewidth]{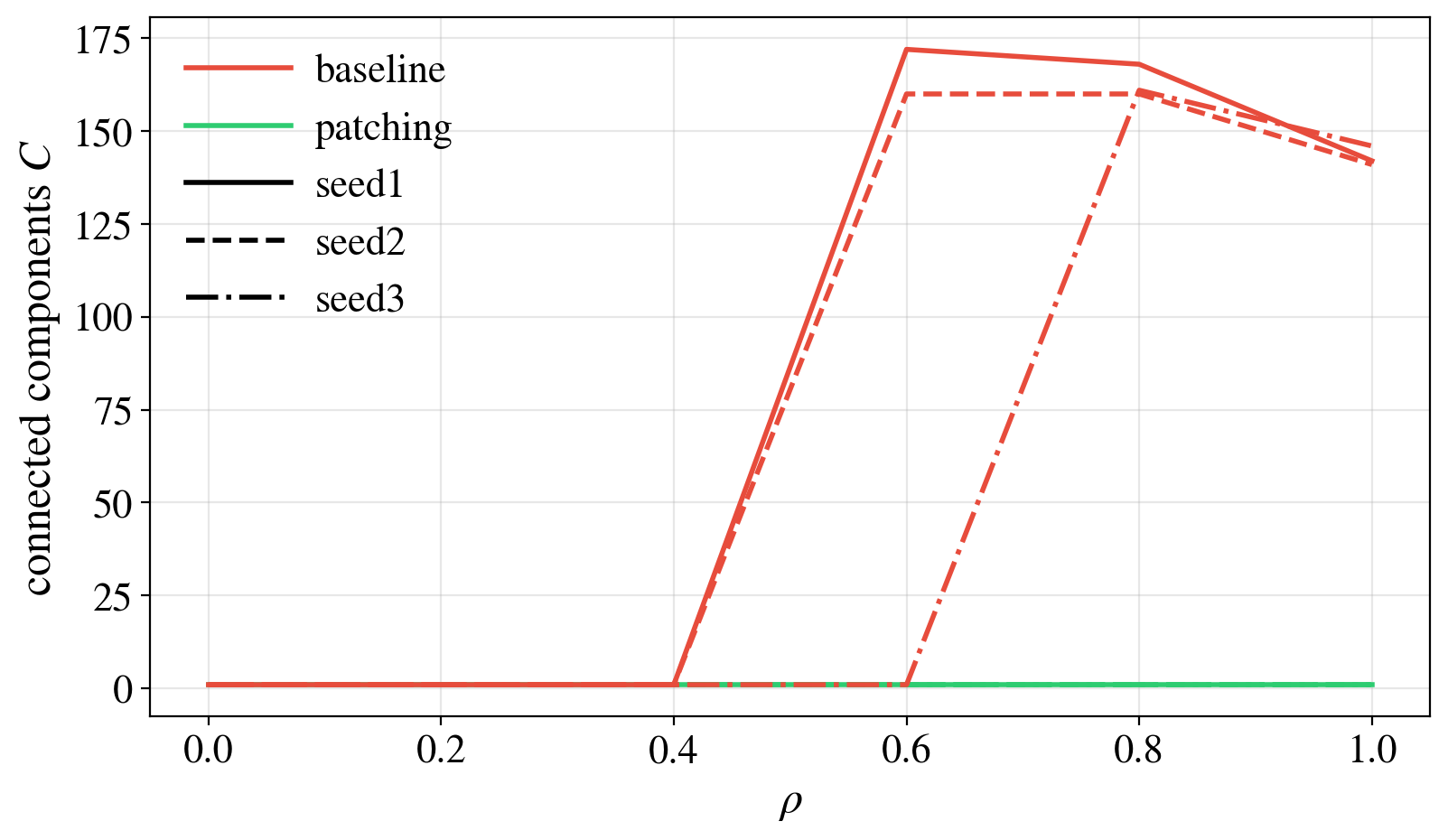}
        \vspace{-5pt}
        \centerline{\small (b) Shape 2 of Figure \ref{fig:your_qualitative_panel_gso_wala} (GSO)}
    \end{minipage}

    \vspace{1em} 

    \begin{minipage}[b]{0.48\linewidth}
        \centering
        \includegraphics[width=\linewidth]{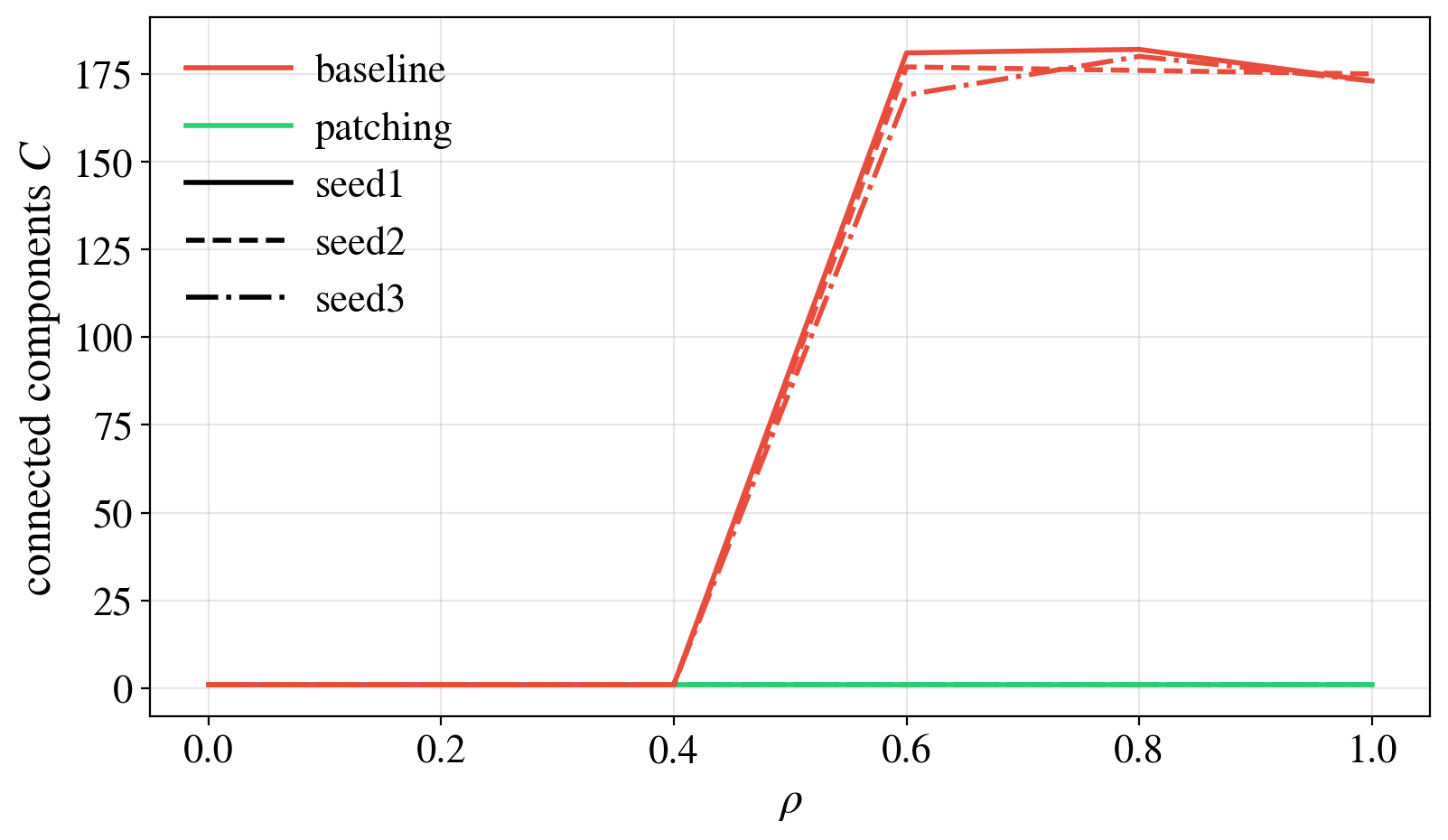}
        \vspace{-5pt}
        \centerline{\small (c) Shape 3 of Figure \ref{fig:your_qualitative_panel_gso_wala} (GSO)}
    \end{minipage}
    \hfill
    \begin{minipage}[b]{0.48\linewidth}
        \centering
        \includegraphics[width=\linewidth]{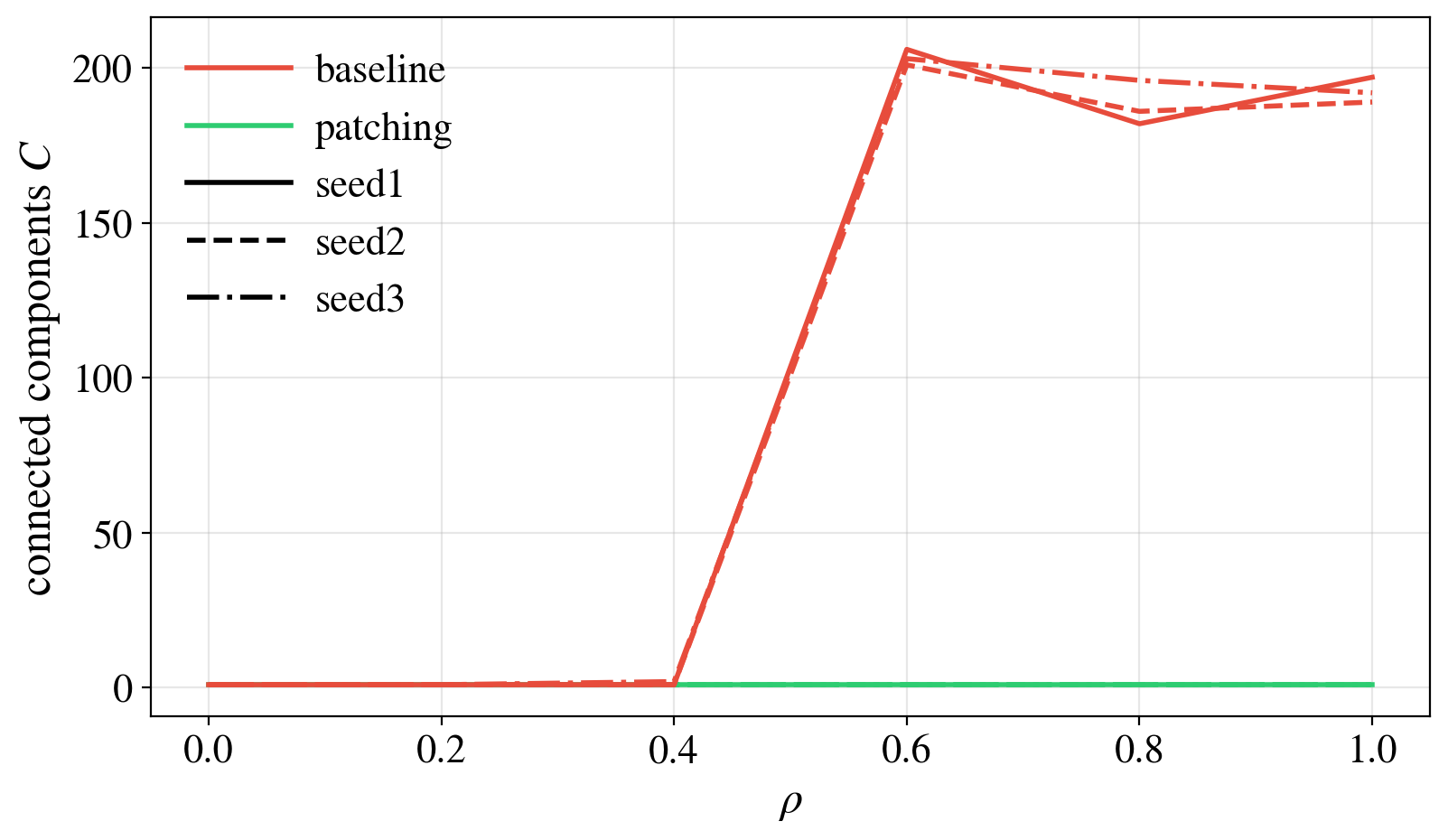}
        \vspace{-5pt}
        \centerline{\small (d) Shape 4 of Figure \ref{fig:your_qualitative_panel_gso_wala} (GSO)}
    \end{minipage}

    \vspace{1em} 

    \begin{minipage}[b]{0.48\linewidth}
        \centering
        \includegraphics[width=\linewidth]{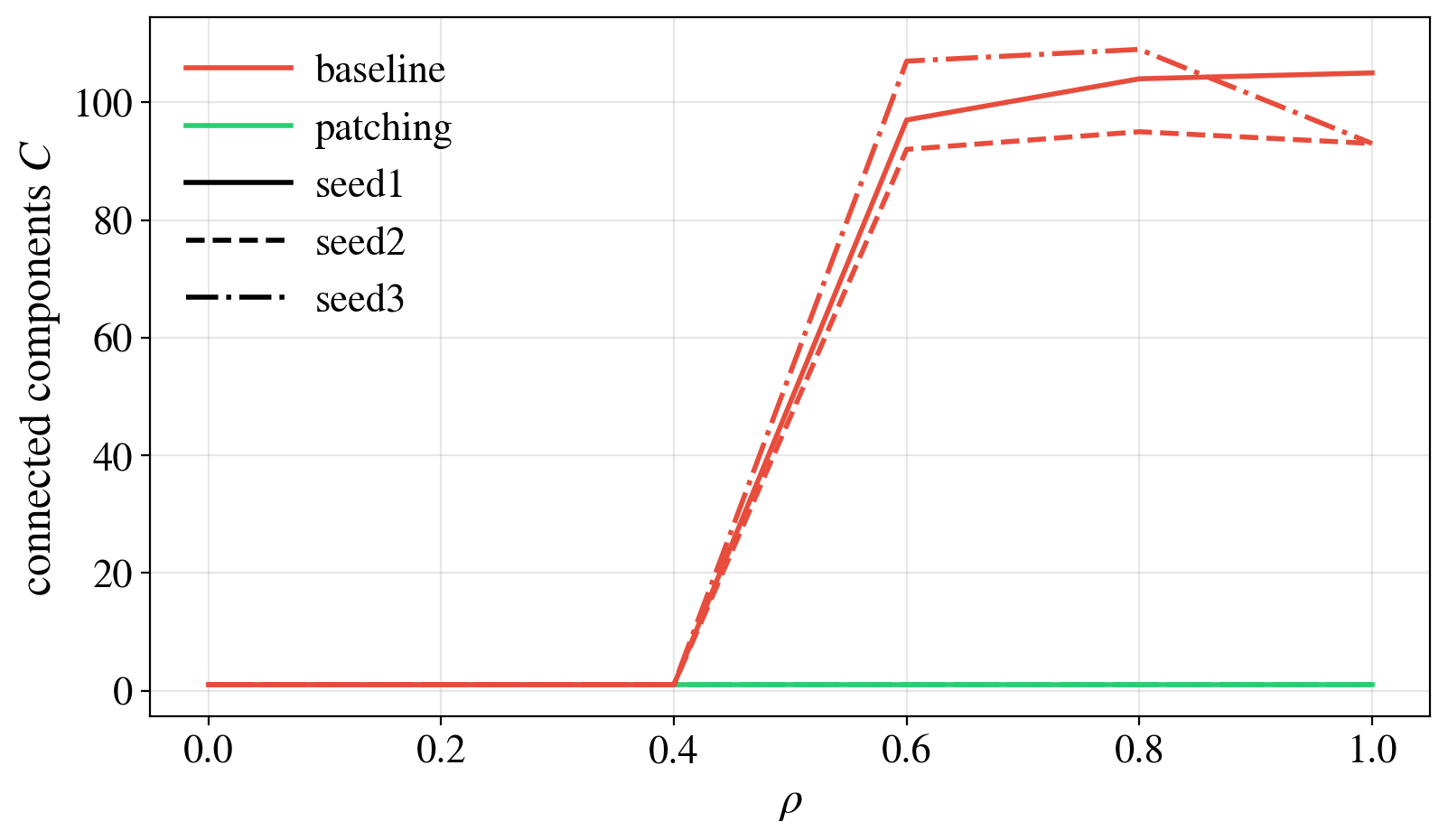}
        \vspace{-5pt}
        \centerline{\small (e) Shape 492 (SimJEB)}
    \end{minipage}
    \hfill
    \begin{minipage}[b]{0.48\linewidth}
        \centering
        \includegraphics[width=\linewidth]{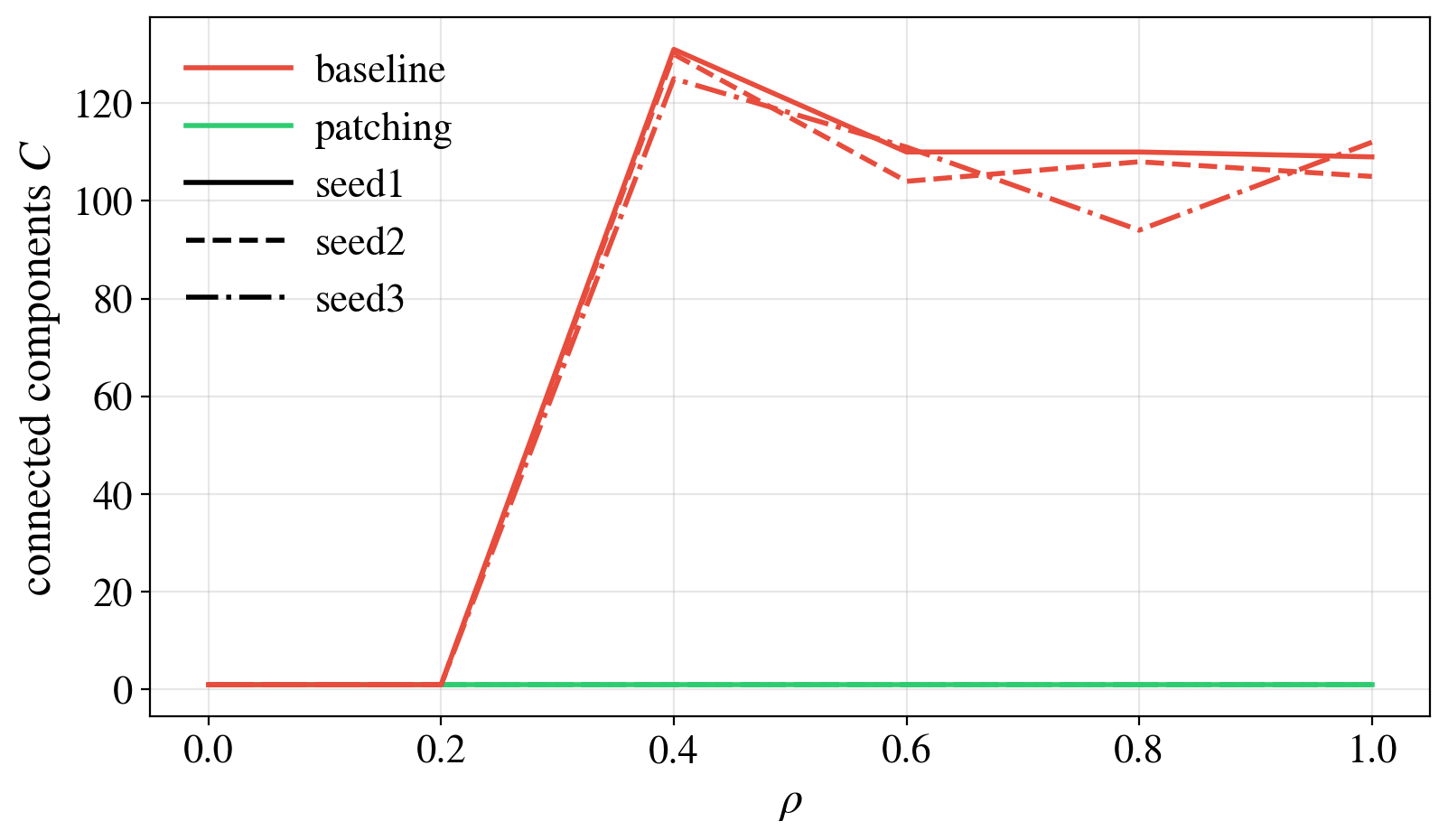}
        \vspace{-5pt}
        \centerline{\small (f) Shape 525 (SimJEB)}
    \end{minipage}

    \caption{Connected components $C$ vs. $\rho$.}
    \label{fig:connectivity_gso_simjeb}
\end{figure*}

\begin{figure*}[t]
    \centering
    \begin{minipage}[b]{0.48\linewidth}
        \centering
        \includegraphics[width=\linewidth]{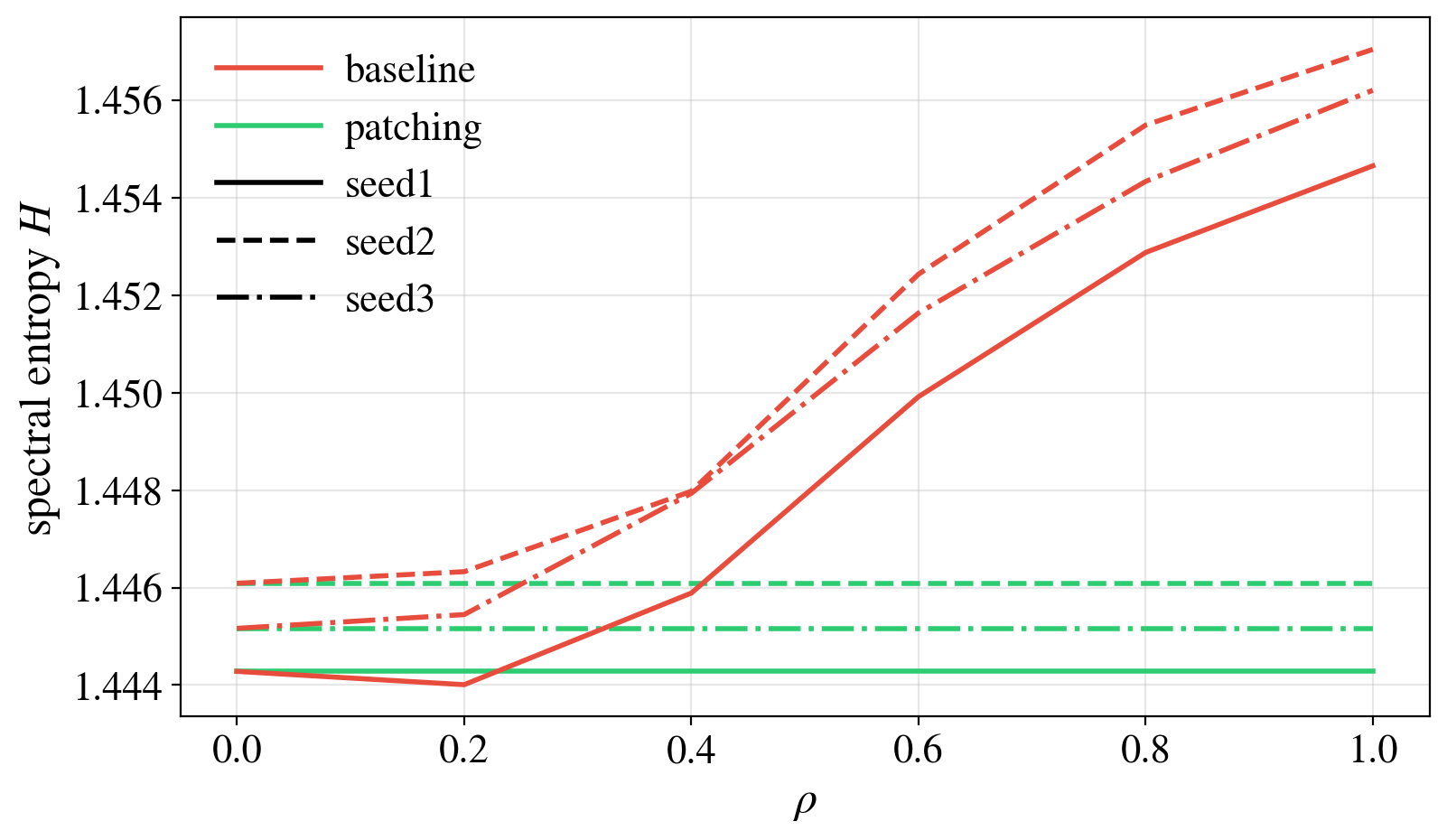}
        \vspace{-5pt}
        \centerline{\small (a) Shape 1 of Figure \ref{fig:your_qualitative_panel_gso_wala} (GSO)}
    \end{minipage}
    \hfill
    \begin{minipage}[b]{0.48\linewidth}
        \centering
        \includegraphics[width=\linewidth]{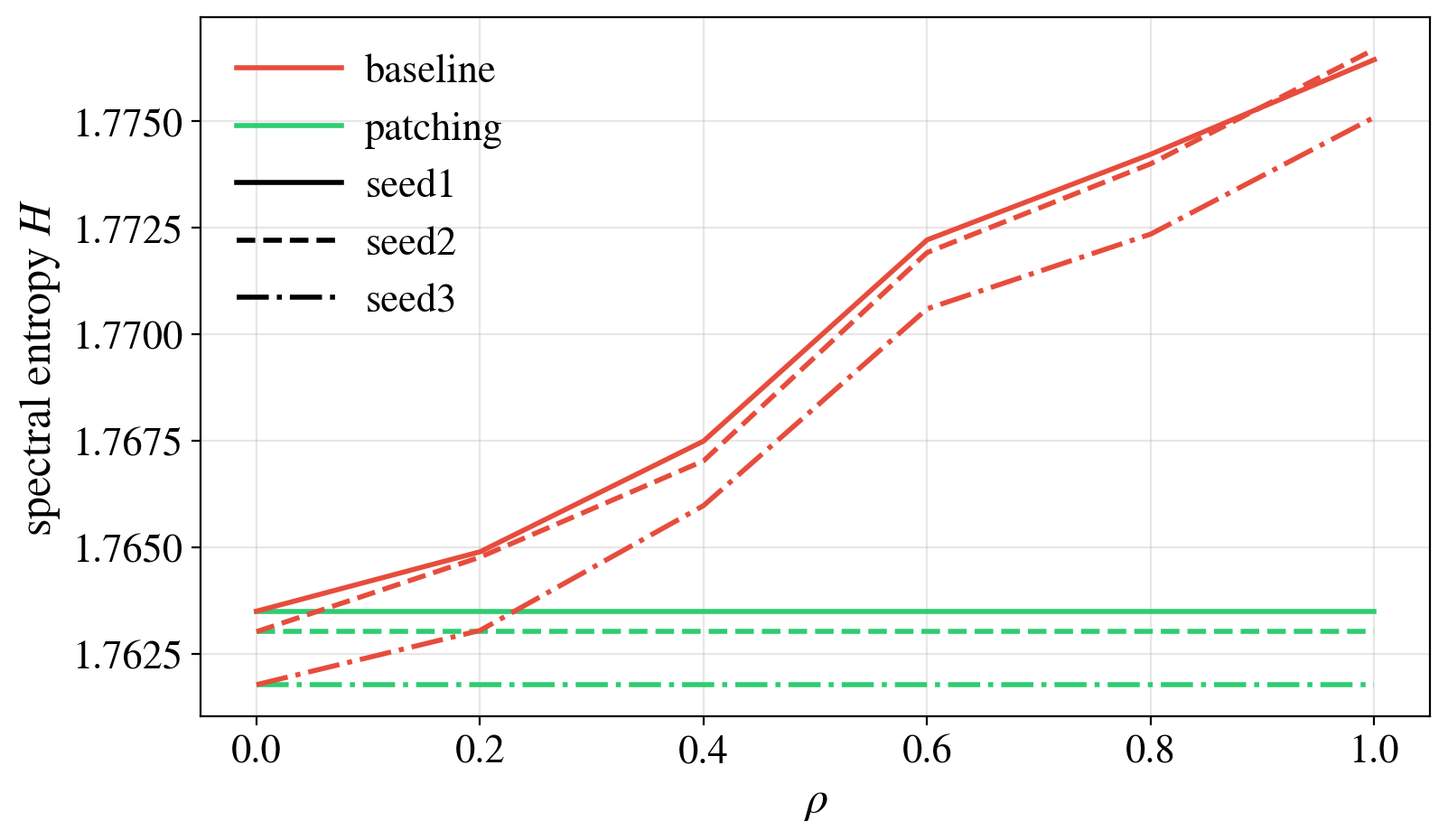}
        \vspace{-5pt}
        \centerline{\small (b) Shape 2 of Figure \ref{fig:your_qualitative_panel_gso_wala} (GSO)}
    \end{minipage}

    \vspace{1em}

    \begin{minipage}[b]{0.48\linewidth}
        \centering
        \includegraphics[width=\linewidth]{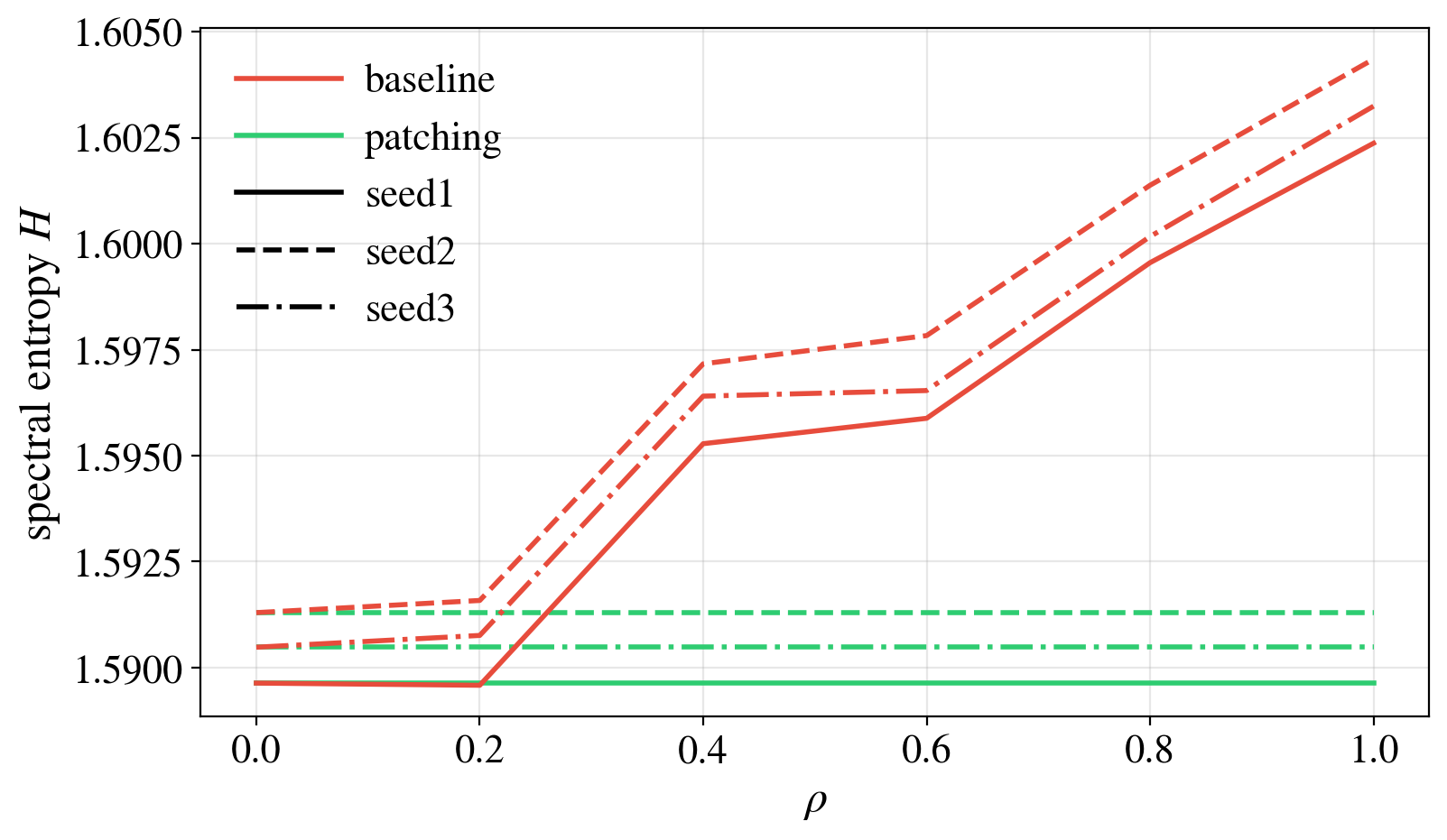}
        \vspace{-5pt}
        \centerline{\small (c) Shape 3 of Figure \ref{fig:your_qualitative_panel_gso_wala} (GSO)}
    \end{minipage}
    \hfill
    \begin{minipage}[b]{0.48\linewidth}
        \centering
        \includegraphics[width=\linewidth]{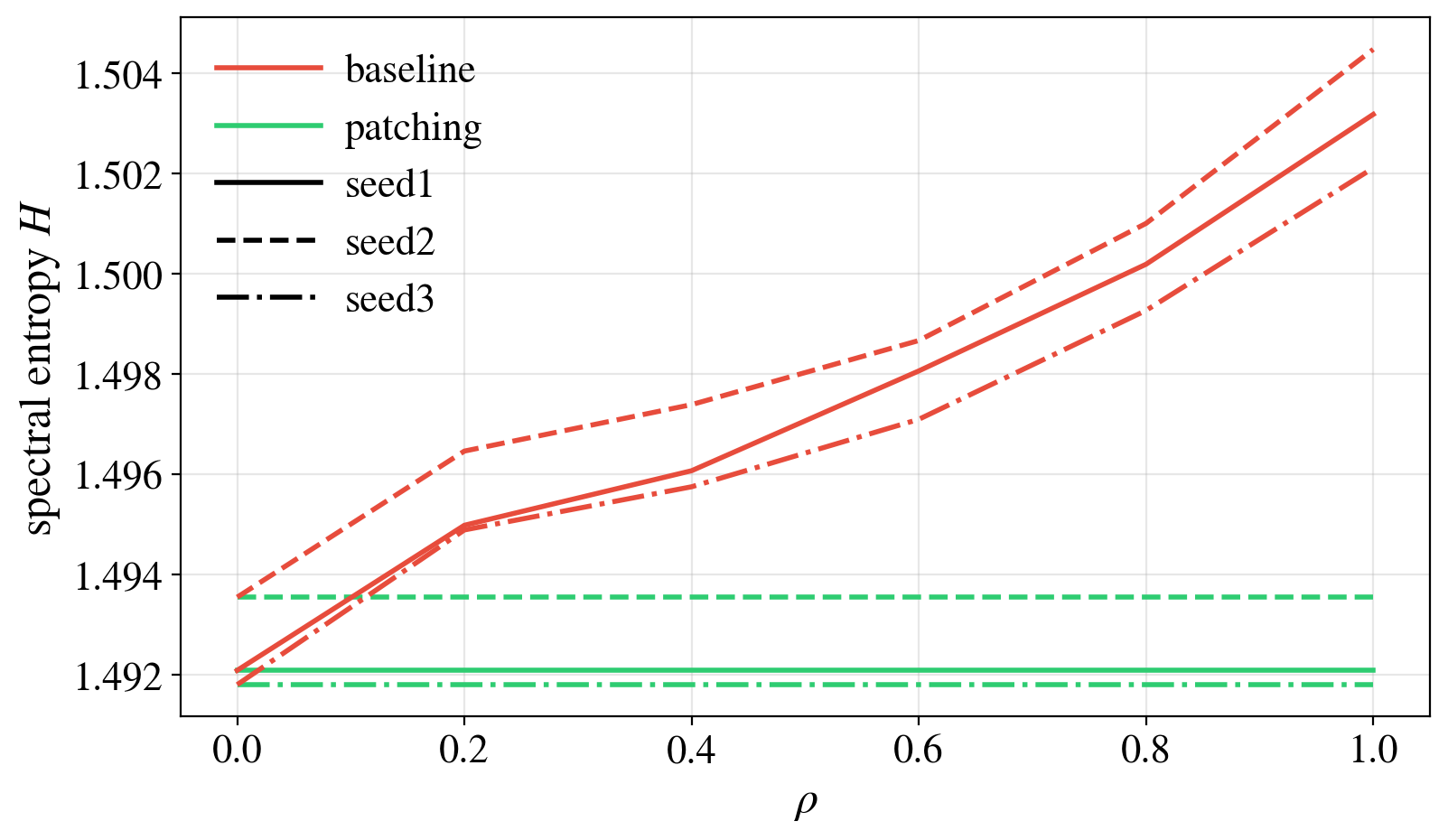}
        \vspace{-5pt}
        \centerline{\small (d) Shape 4 of Figure \ref{fig:your_qualitative_panel_gso_wala} (GSO)}
    \end{minipage}

    \vspace{1em}

    \begin{minipage}[b]{0.48\linewidth}
        \centering
        \includegraphics[width=\linewidth]{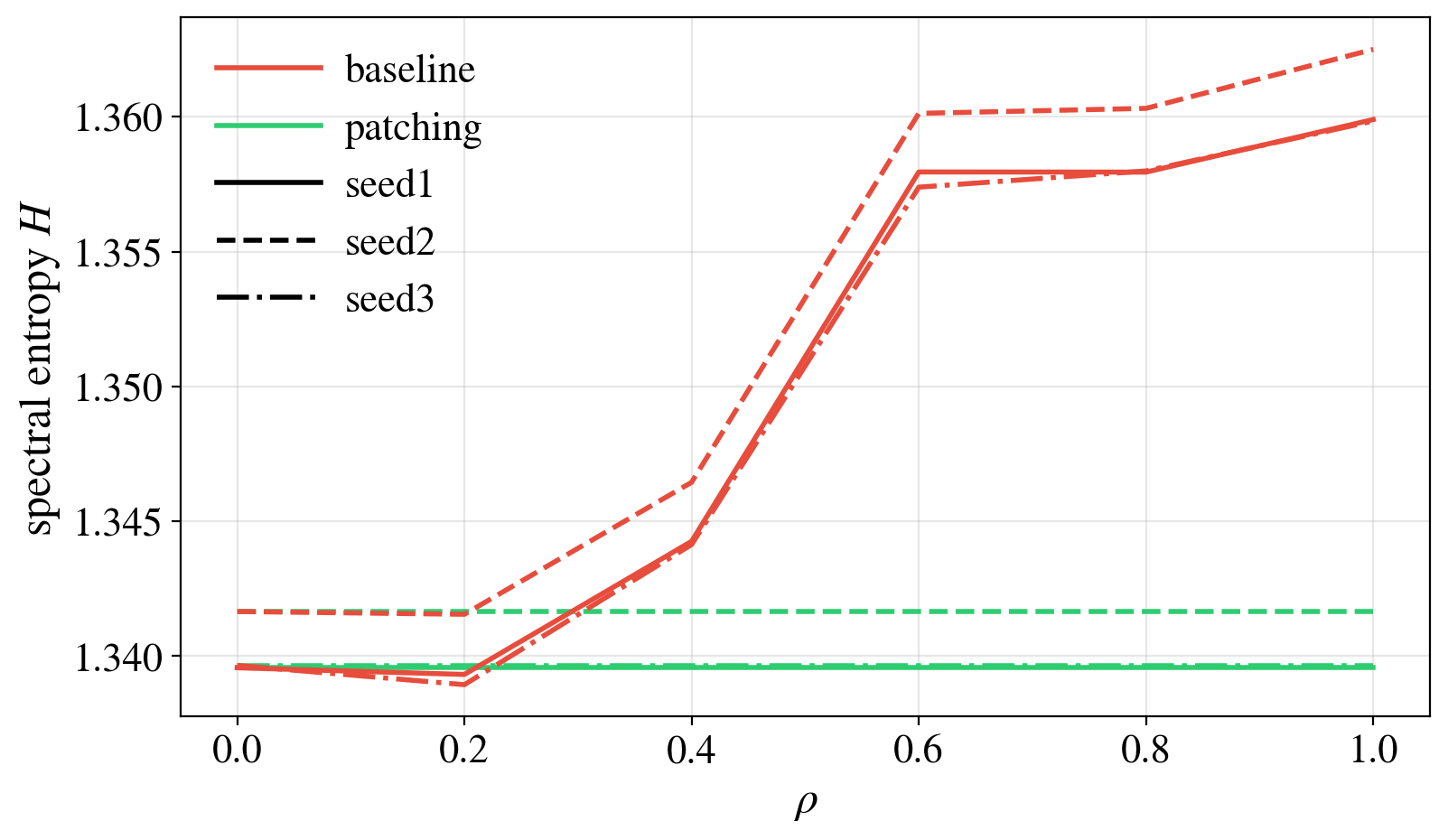}
        \vspace{-5pt}
        \centerline{\small (e) Shape 492 (SimJEB)}
    \end{minipage}
    \hfill
    \begin{minipage}[b]{0.48\linewidth}
        \centering
        \includegraphics[width=\linewidth]{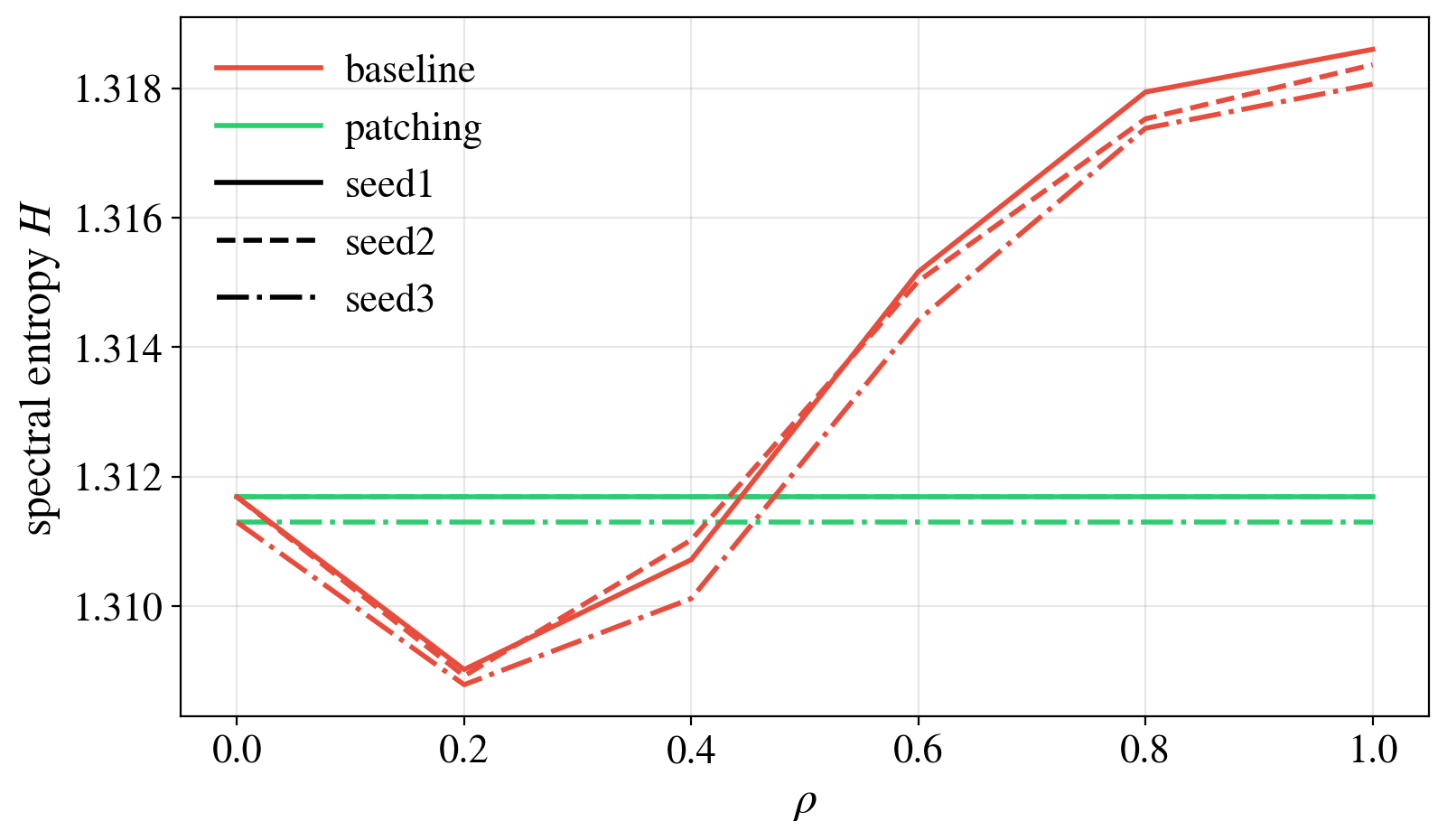}
        \vspace{-5pt}
        \centerline{\small (f) Shape 525 (SimJEB)}
    \end{minipage}

    \caption{Spectral entropy $H$ vs. $\rho$.}
    \label{fig:spectralEntropy_gso_simjeb}
\end{figure*}

\begin{figure*}[t]
    \centering
    \begin{minipage}[b]{0.48\linewidth}
        \centering
        \includegraphics[width=\linewidth]{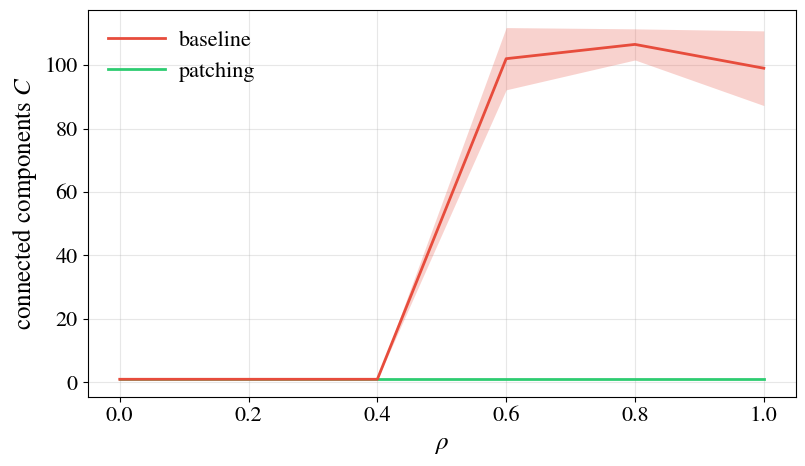}
        \vspace{-5pt}
        \centerline{\small (a) Connected components $C$ vs. $\rho$}
    \end{minipage}
    \hfill
    \begin{minipage}[b]{0.48\linewidth}
        \centering
        \includegraphics[width=\linewidth]{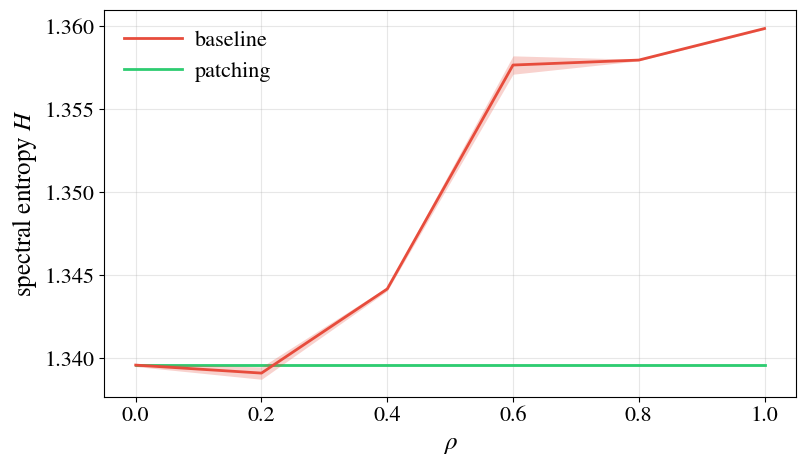}
        \vspace{-5pt}
        \centerline{\small (b) Spectral entropy $H$ vs. $\rho$}
    \end{minipage}
    
    \caption{Patterns at population level for SimJEB shape 492, using 150 diffusion seeds.}
    \label{fig:C_H_population_492}
\end{figure*}

In this section, we provide additional evidence that the  patterns observed in Section \ref{sec:activation_patching}-\ref{sec:patching_effect} generalize when evaluated on  more data points and random seeds. Figure \ref{fig:connectivity_gso_simjeb} and Figure \ref{fig:spectralEntropy_gso_simjeb} show that the behavior transfers to diverse shapes from the GSO \citep{downs2022googlescannedobjectshighquality} and SimJEB \citep{Whalen_2021} corpora as well as diffusion seeds for the \textsc{WaLa} model. Figure \ref{fig:C_H_population_492} shows that the average behavior over a population of 150 diffusion seeds for SimJEB shape 492 is consistent with the observations reported in Section \ref{sec:patching_effect} for the \textsc{WaLa} model.

\clearpage
\subsection{Additional Spectral Metrics}\label{app:more_spectral_metrics}

\begin{figure*}[t]
    \centering
    \begin{minipage}[b]{0.48\linewidth}
        \centering
        \includegraphics[width=\linewidth]{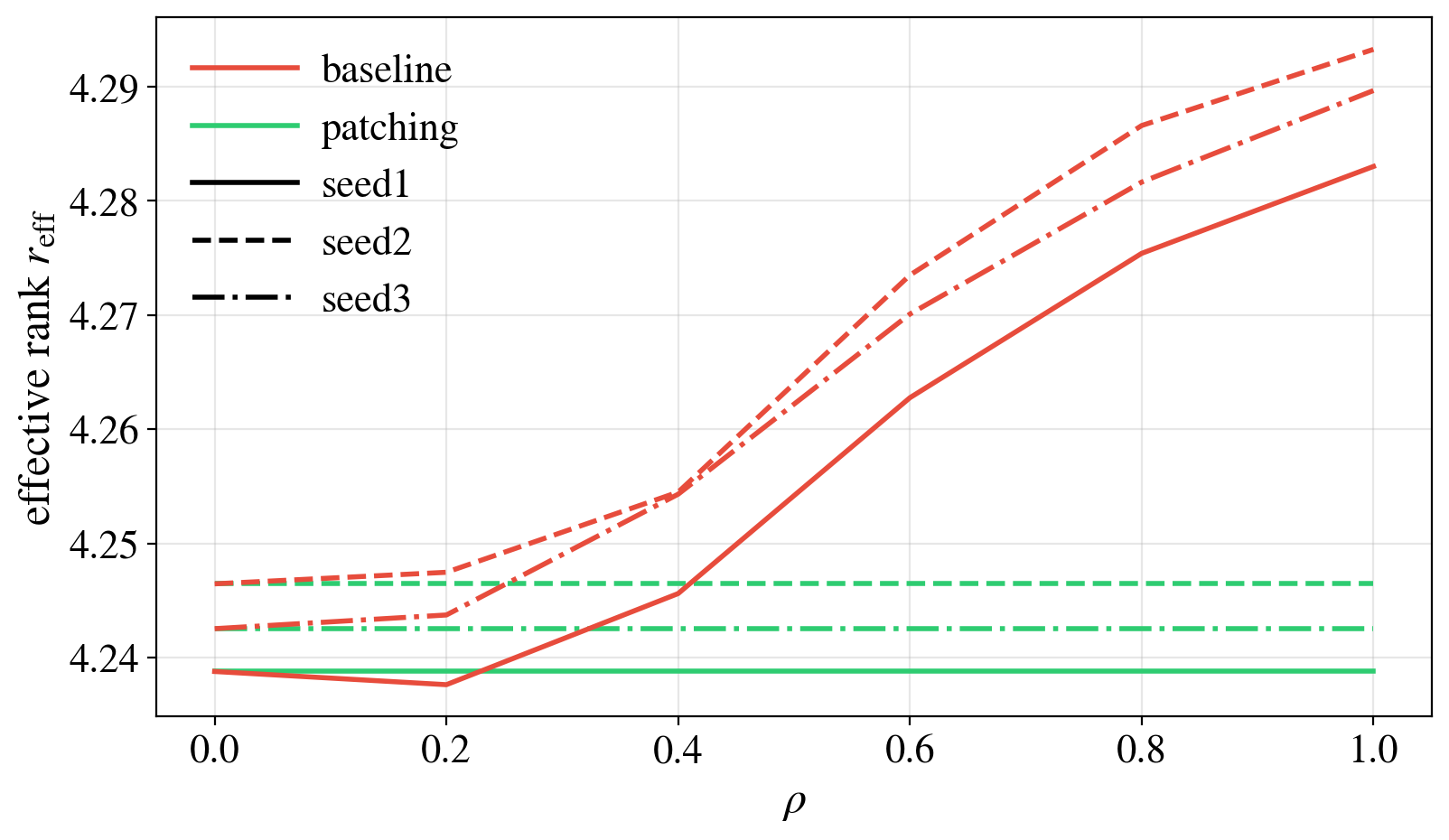}
        \vspace{-5pt}
        \centerline{\small (a) Shape 1 of Figure \ref{fig:your_qualitative_panel_gso_wala} (GSO)}
    \end{minipage}
    \hfill
    \begin{minipage}[b]{0.48\linewidth}
        \centering
        \includegraphics[width=\linewidth]{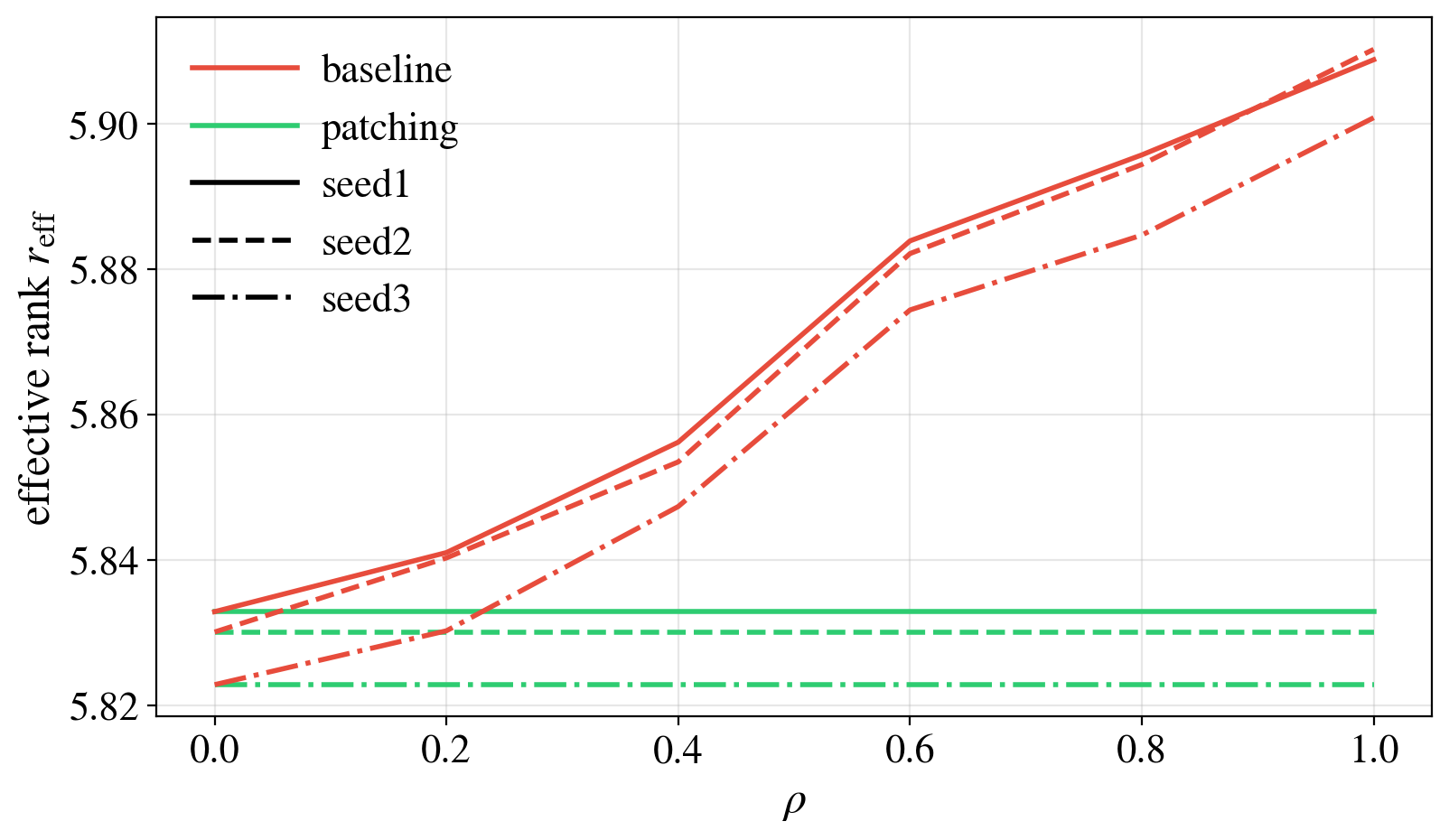}
        \vspace{-5pt}
        \centerline{\small (b) Shape 2 of Figure \ref{fig:your_qualitative_panel_gso_wala} (GSO)}
    \end{minipage}

    \vspace{1em}

    \begin{minipage}[b]{0.48\linewidth}
        \centering
        \includegraphics[width=\linewidth]{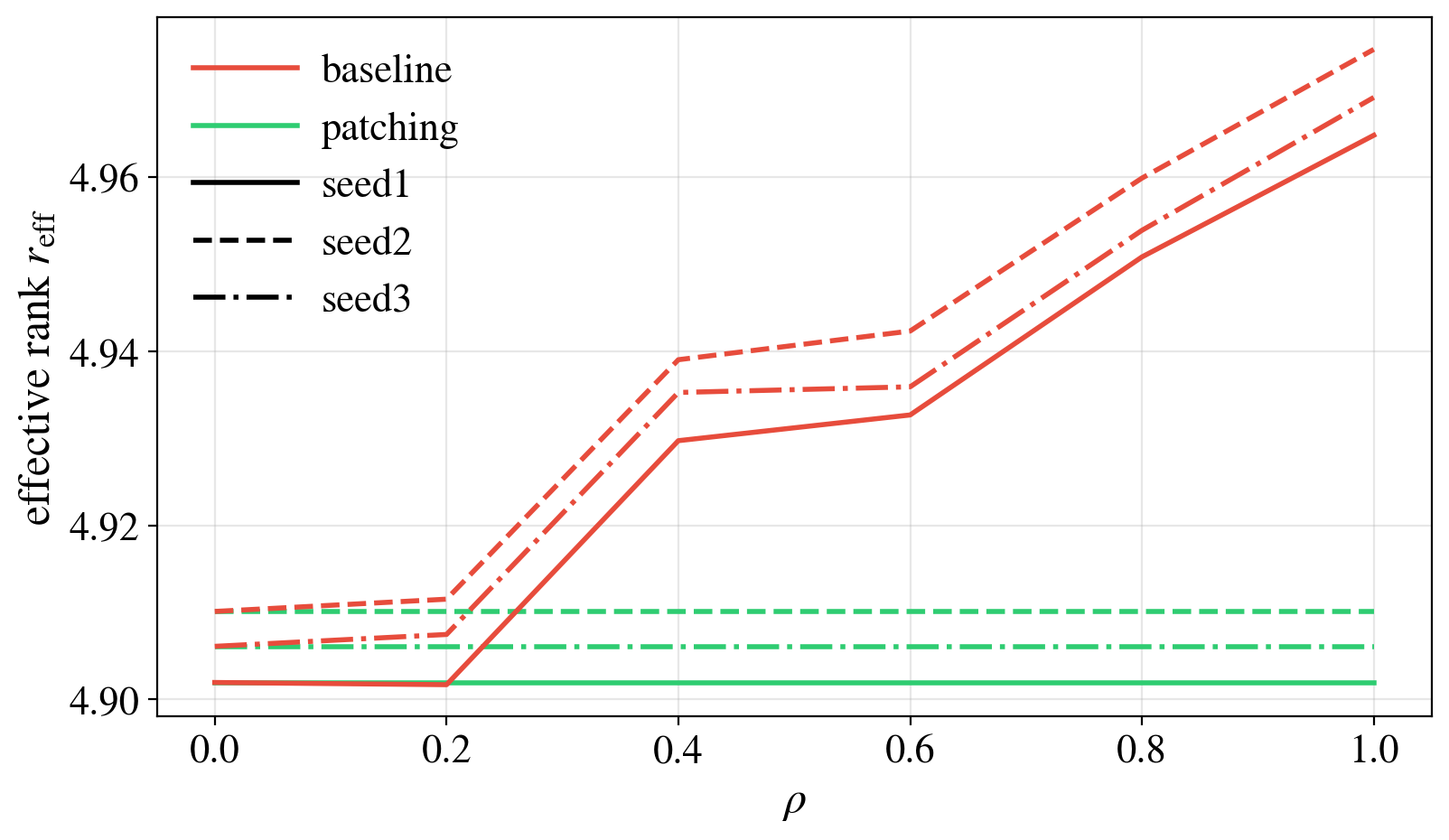}
        \vspace{-5pt}
        \centerline{\small (c) Shape 3 of Figure \ref{fig:your_qualitative_panel_gso_wala} (GSO)}
    \end{minipage}
    \hfill
    \begin{minipage}[b]{0.48\linewidth}
        \centering
        \includegraphics[width=\linewidth]{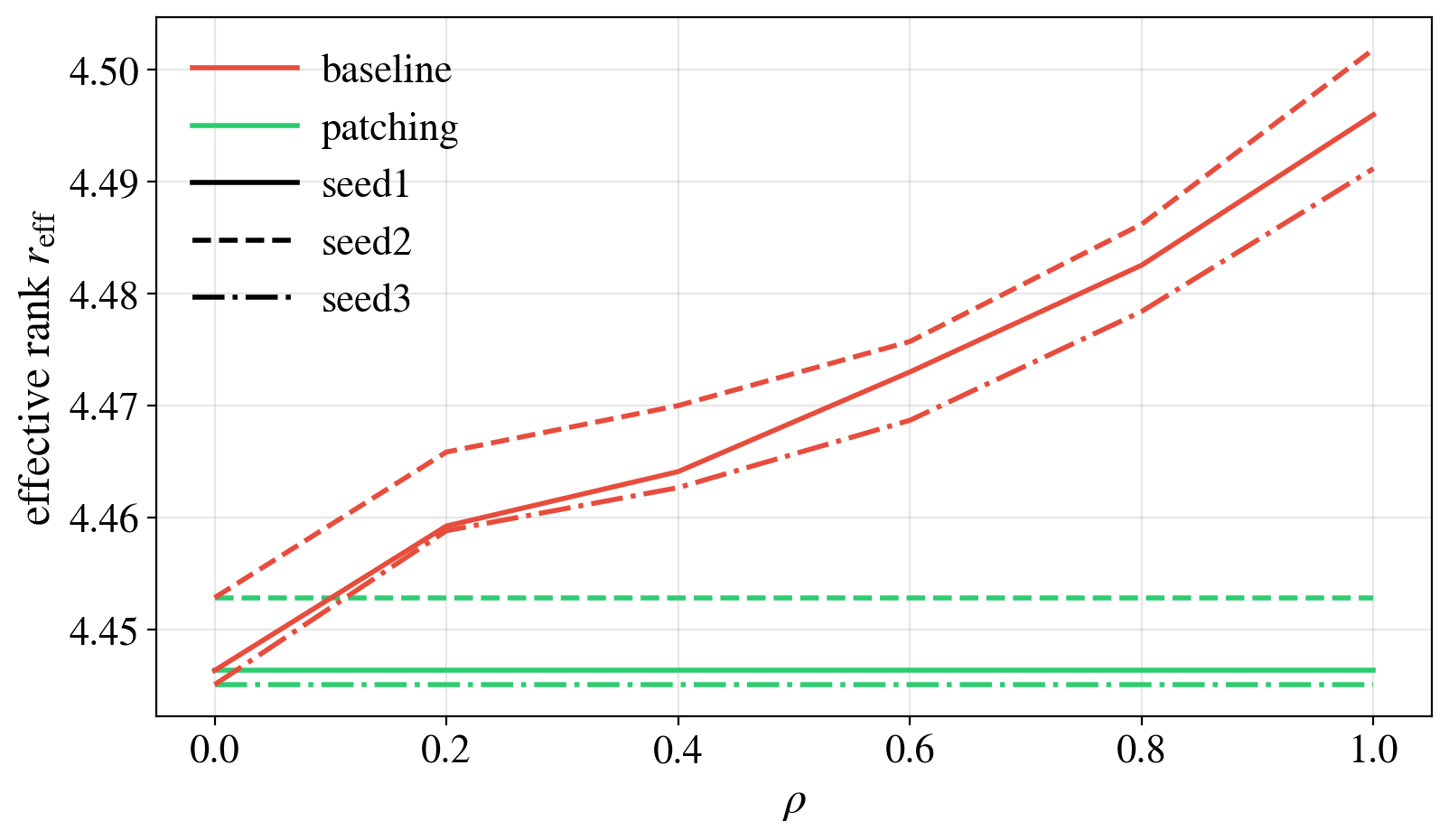}
        \vspace{-5pt}
        \centerline{\small (d) Shape 4 of Figure \ref{fig:your_qualitative_panel_gso_wala} (GSO)}
    \end{minipage}

    \vspace{1em}

    \begin{minipage}[b]{0.48\linewidth}
        \centering
        \includegraphics[width=\linewidth]{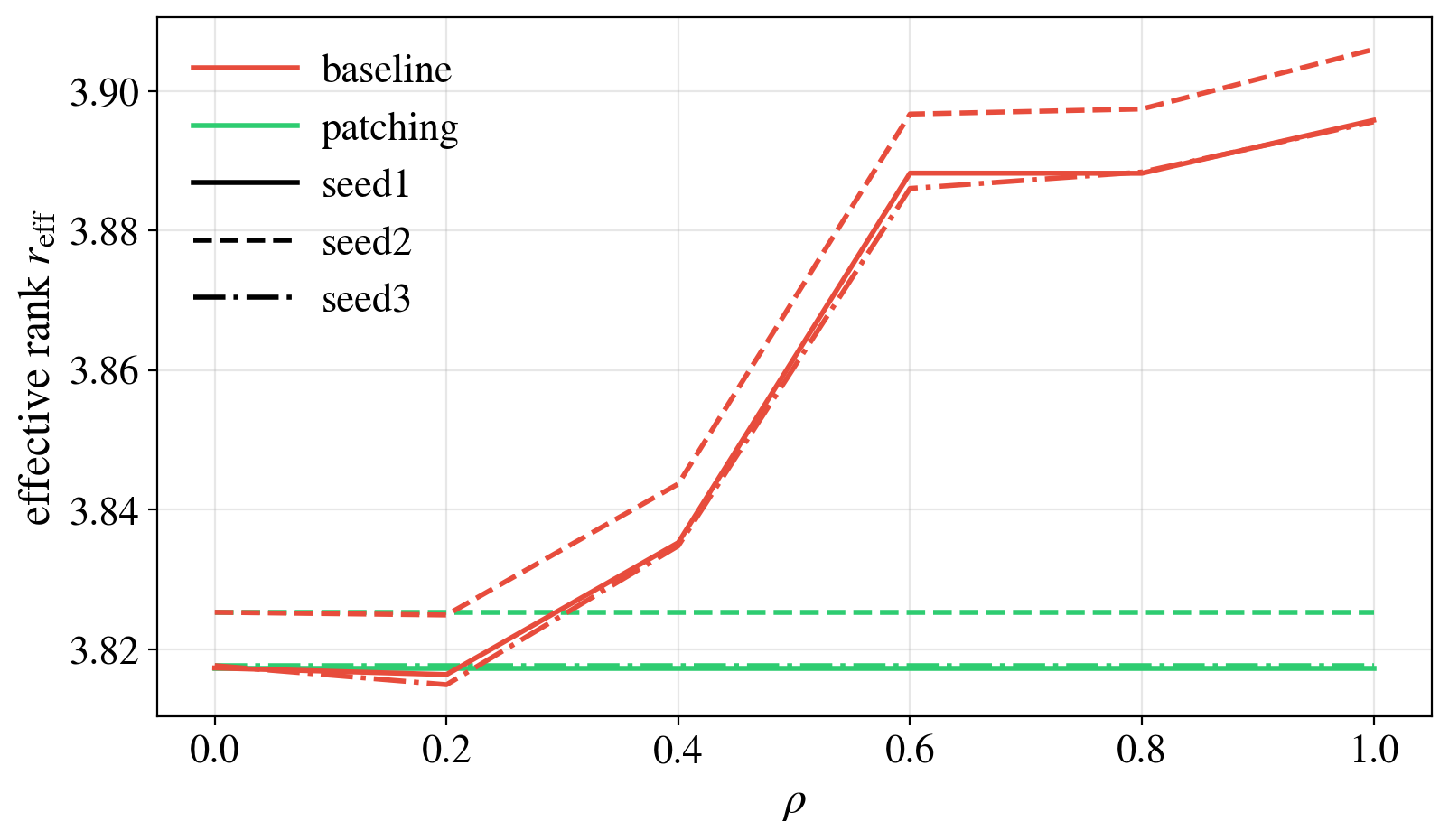}
        \vspace{-5pt}
        \centerline{\small (e) Shape 492 (SimJEB)}
    \end{minipage}
    \hfill
    \begin{minipage}[b]{0.48\linewidth}
        \centering
        \includegraphics[width=\linewidth]{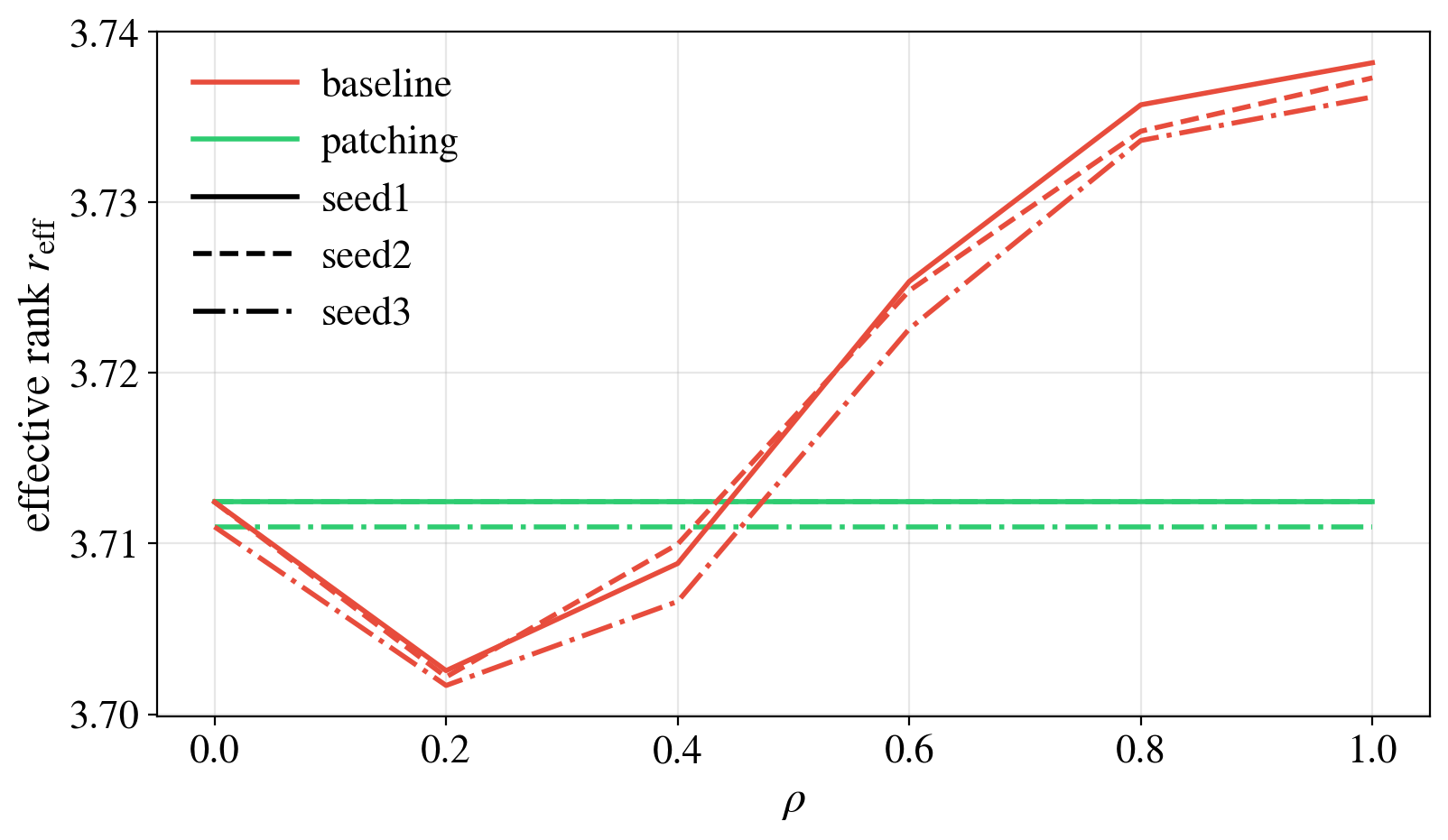}
        \vspace{-5pt}
        \centerline{\small (f) Shape 525 (SimJEB)}
    \end{minipage}

    \caption{Effective rank $r_{\text{eff}}$ vs. $\rho$.}
    \label{fig:erank_gso_simjeb}
\end{figure*}

\begin{figure*}[t]
    \centering
    \begin{minipage}[b]{0.48\linewidth}
        \centering
        \includegraphics[width=\linewidth]{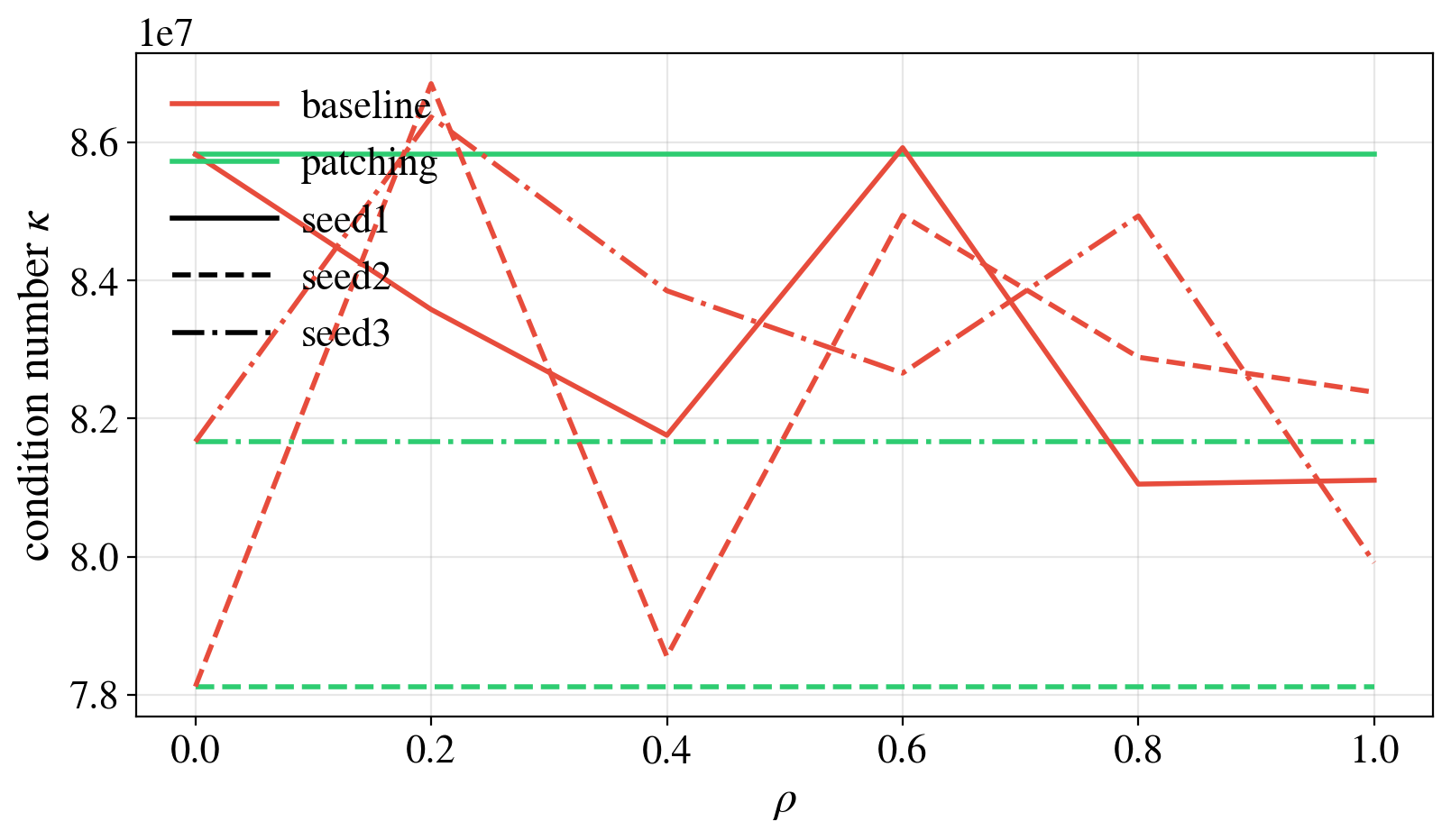}
        \vspace{-5pt}
        \centerline{\small (a) Shape 1 of Figure \ref{fig:your_qualitative_panel_gso_wala} (GSO)}
    \end{minipage}
    \hfill
    \begin{minipage}[b]{0.48\linewidth}
        \centering
        \includegraphics[width=\linewidth]{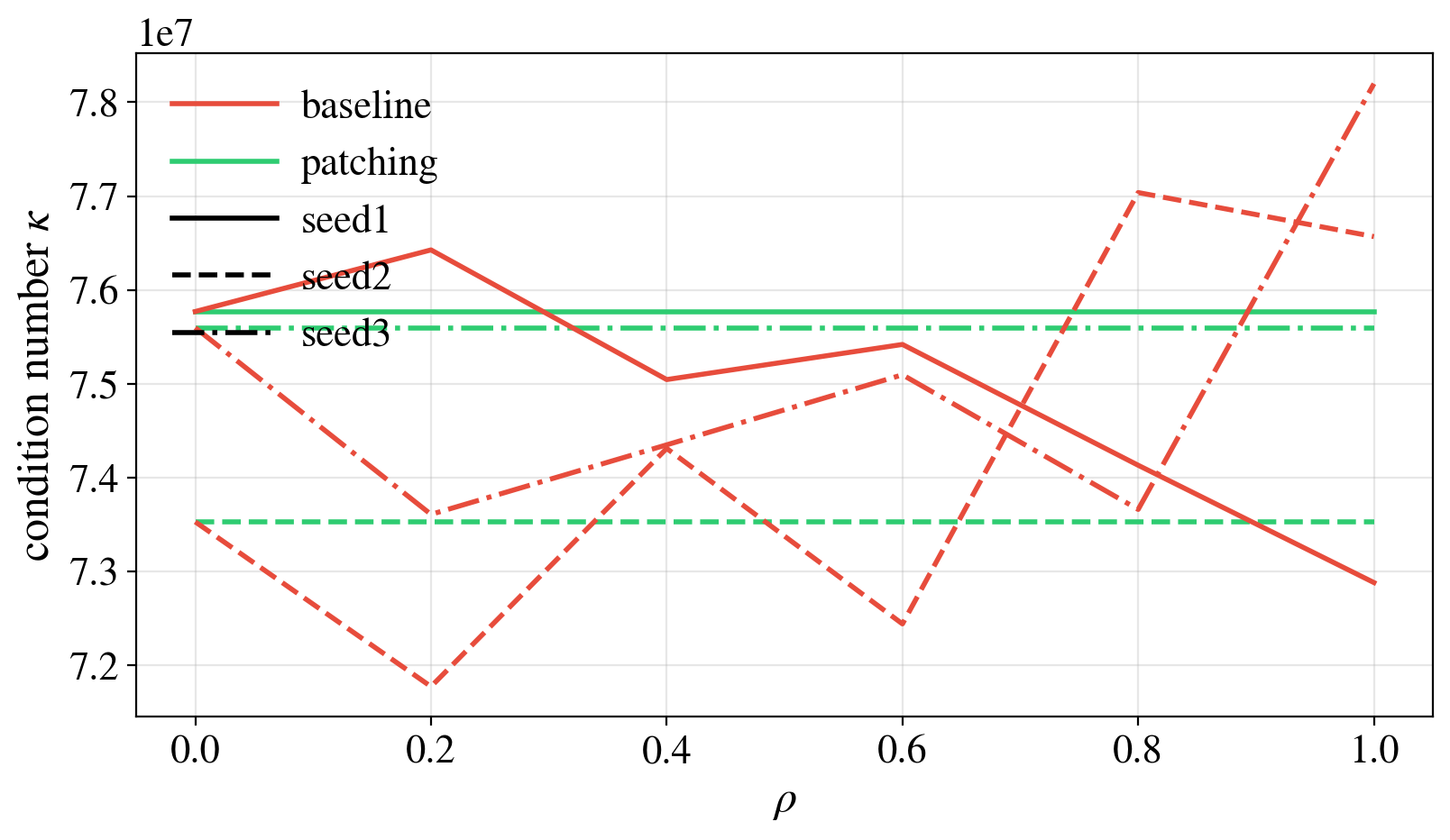}
        \vspace{-5pt}
        \centerline{\small (b) Shape 2 of Figure \ref{fig:your_qualitative_panel_gso_wala} (GSO)}
    \end{minipage}

    \vspace{1em}

    \begin{minipage}[b]{0.48\linewidth}
        \centering
        \includegraphics[width=\linewidth]{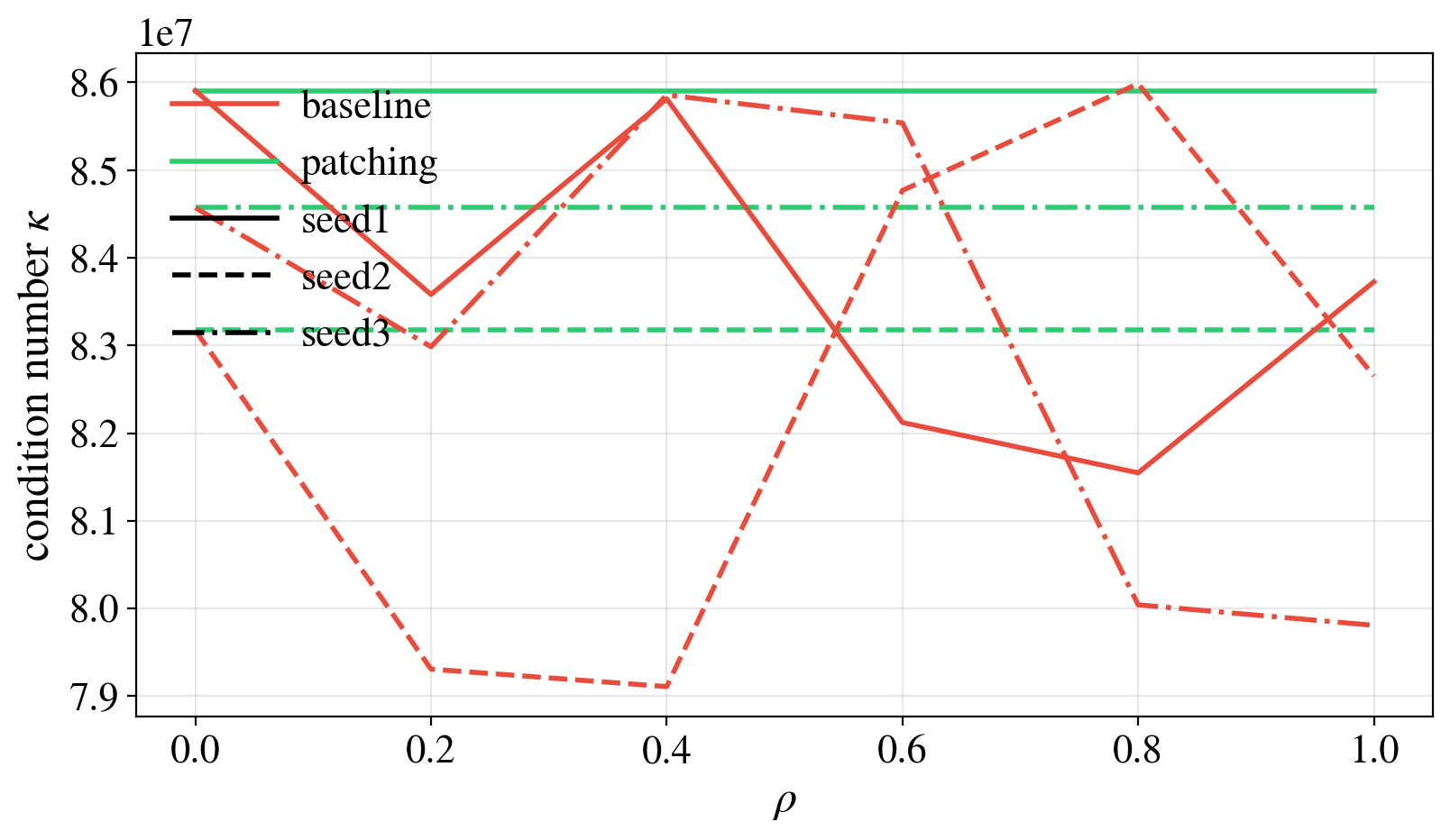}
        \vspace{-5pt}
        \centerline{\small (c) Shape 3 of Figure \ref{fig:your_qualitative_panel_gso_wala} (GSO)}
    \end{minipage}
    \hfill
    \begin{minipage}[b]{0.48\linewidth}
        \centering
        \includegraphics[width=\linewidth]{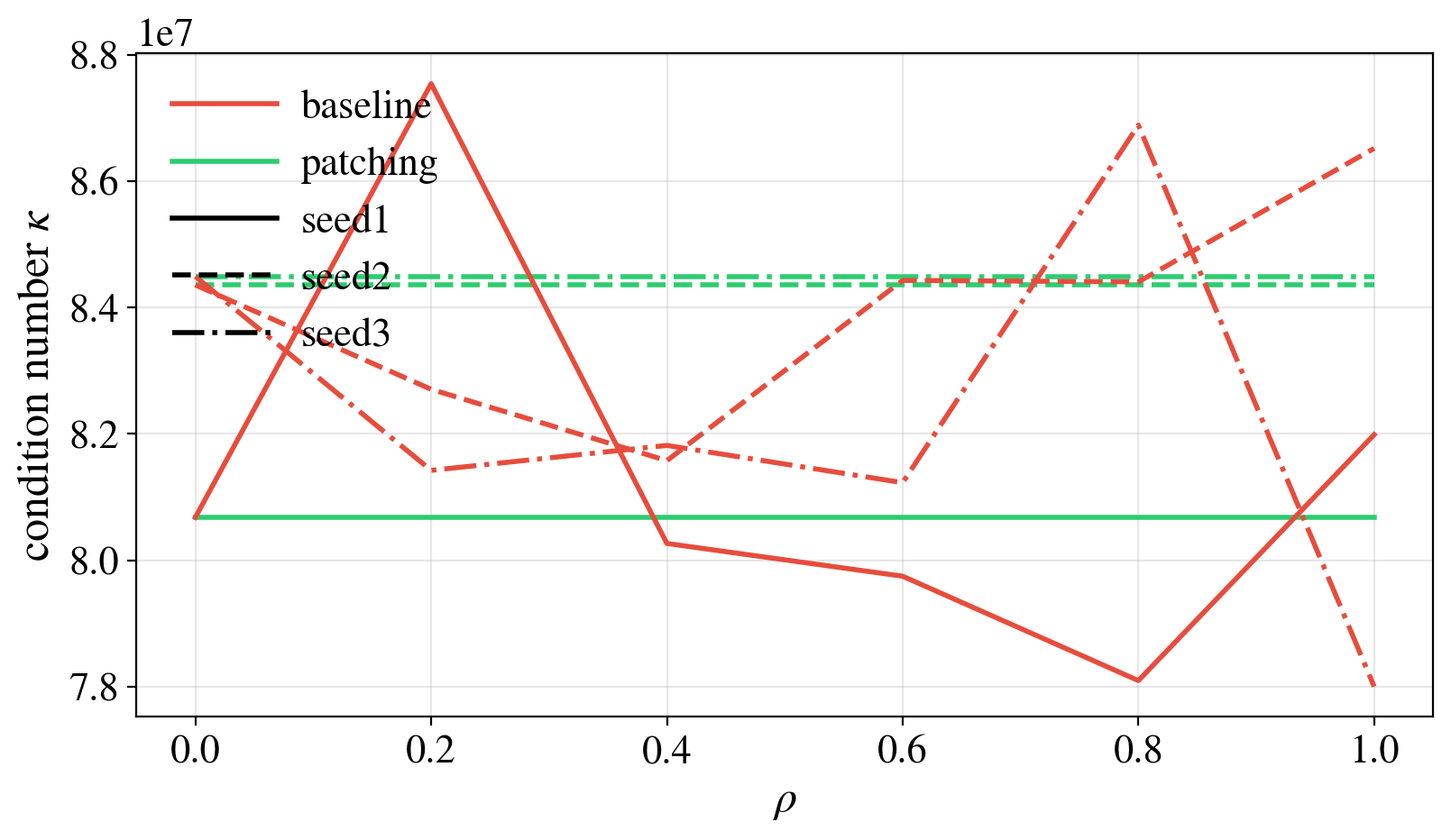}
        \vspace{-5pt}
        \centerline{\small (d) Shape 4 of Figure \ref{fig:your_qualitative_panel_gso_wala} (GSO)}
    \end{minipage}

    \vspace{1em}

    \begin{minipage}[b]{0.48\linewidth}
        \centering
        \includegraphics[width=\linewidth]{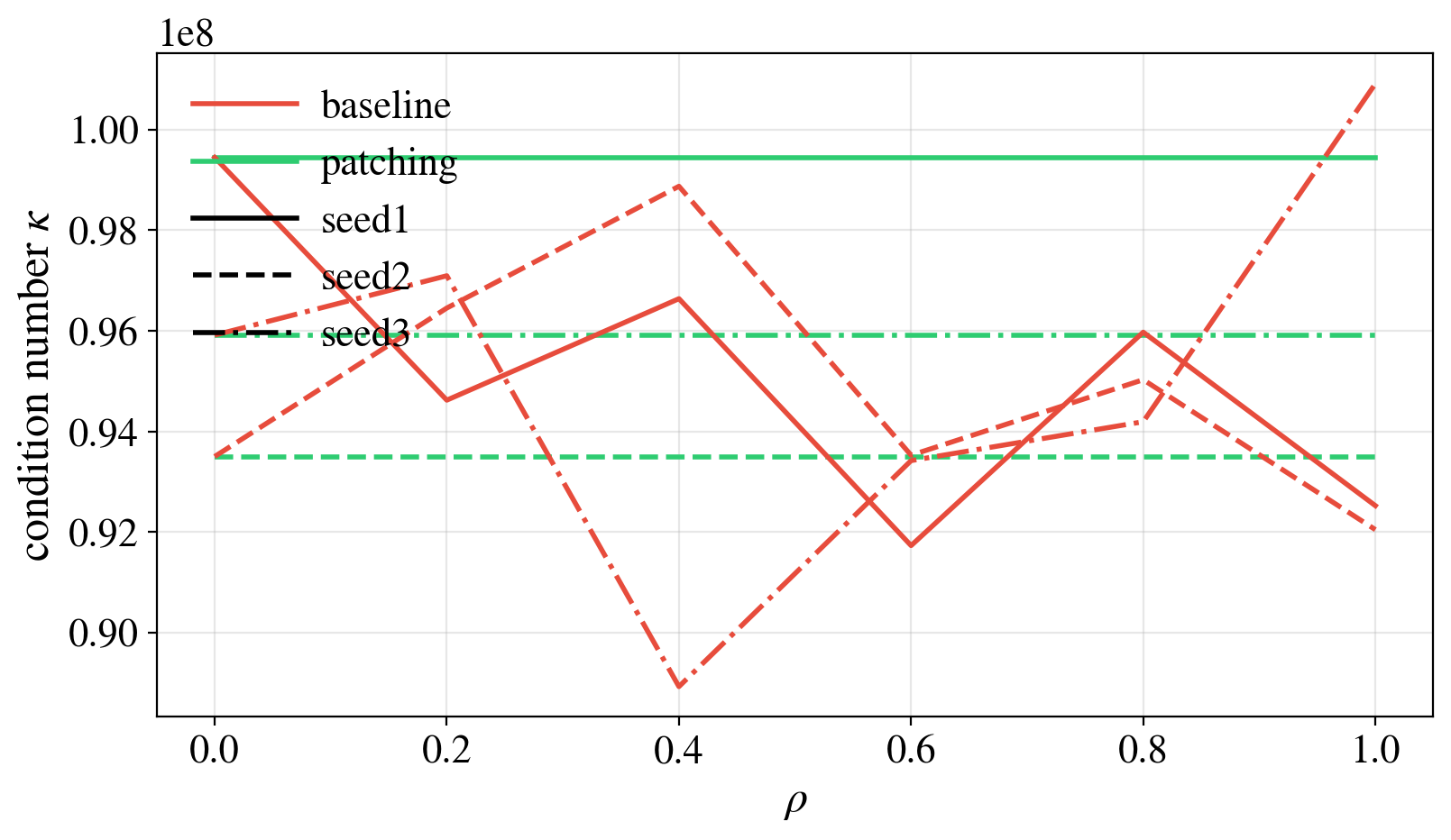}
        \vspace{-5pt}
        \centerline{\small (e) Shape 492 (SimJEB)}
    \end{minipage}
    \hfill
    \begin{minipage}[b]{0.48\linewidth}
        \centering
        \includegraphics[width=\linewidth]{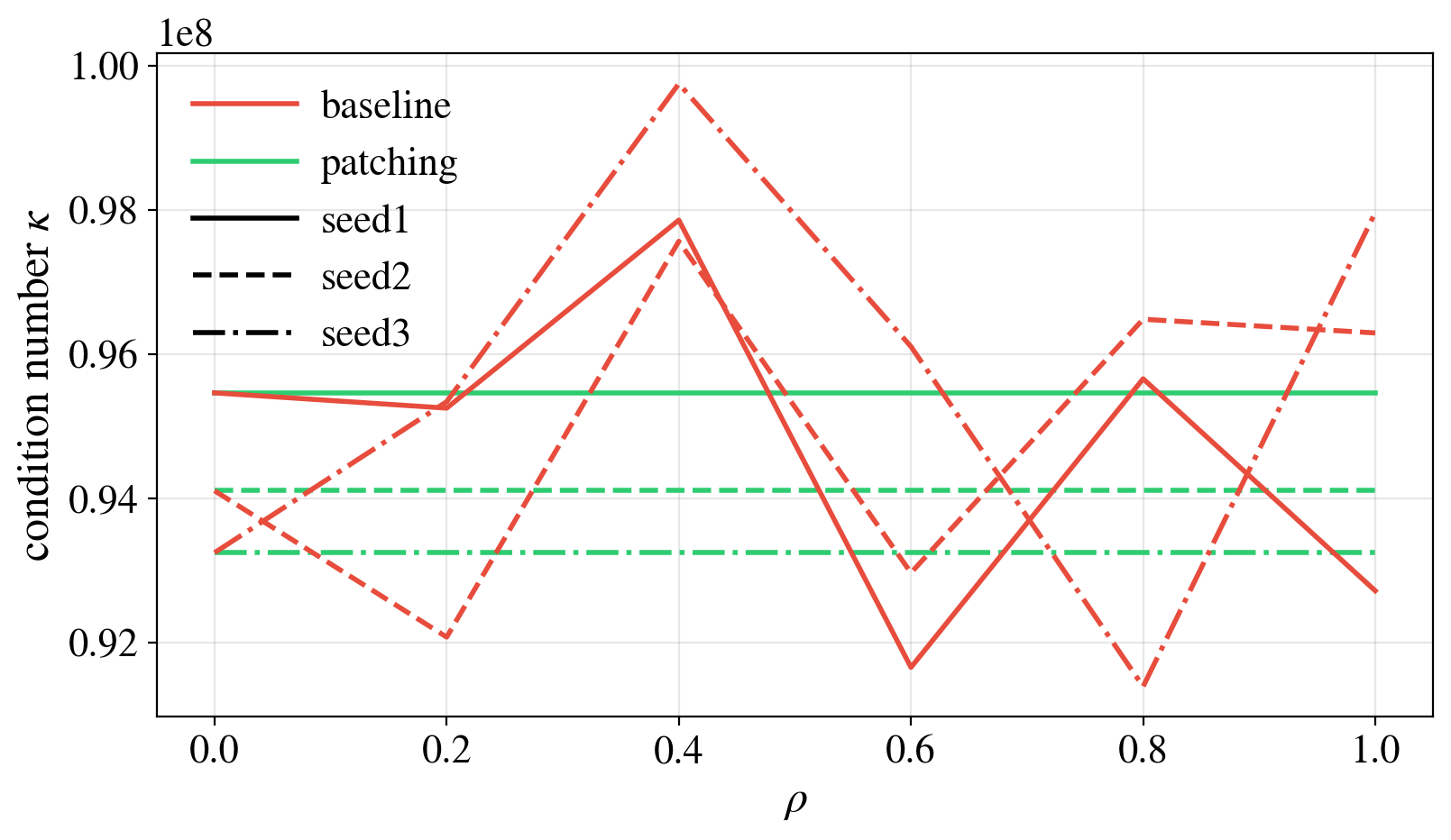}
        \vspace{-5pt}
        \centerline{\small (f) Shape 525 (SimJEB)}
    \end{minipage}

    \caption{Condition number $\kappa$ vs. $\rho$.}
    \label{fig:kappa_gso_simjeb}
\end{figure*}

In this section, we analyze  additional spectral metrics to assess their suitability as indicators of Meltdown for the \textsc{WaLa} model. In particular, Figure \ref{fig:erank_gso_simjeb} reports the  effective rank $r_{\text{eff}}=\exp(H)$ and Figure \ref{fig:kappa_gso_simjeb} the condition number $\kappa=\sigma_{\max}/\sigma_{\min}$ as alternatives to spectral entropy for a diverse set of shapes. We observe that the effective rank—which is a monotonic transformation of spectral entropy—provides an equally informative indicator of Meltdown. By contrast, the condition number exhibits no apparent correlation with the failure phenomenon, suggesting that it is not a suitable diagnostic metric in this setting.

\clearpage
\subsection{Multiple Objects}\label{app:mult_objects}
\begin{figure}[h!]
    \centering
    \includegraphics[width=0.6\linewidth]{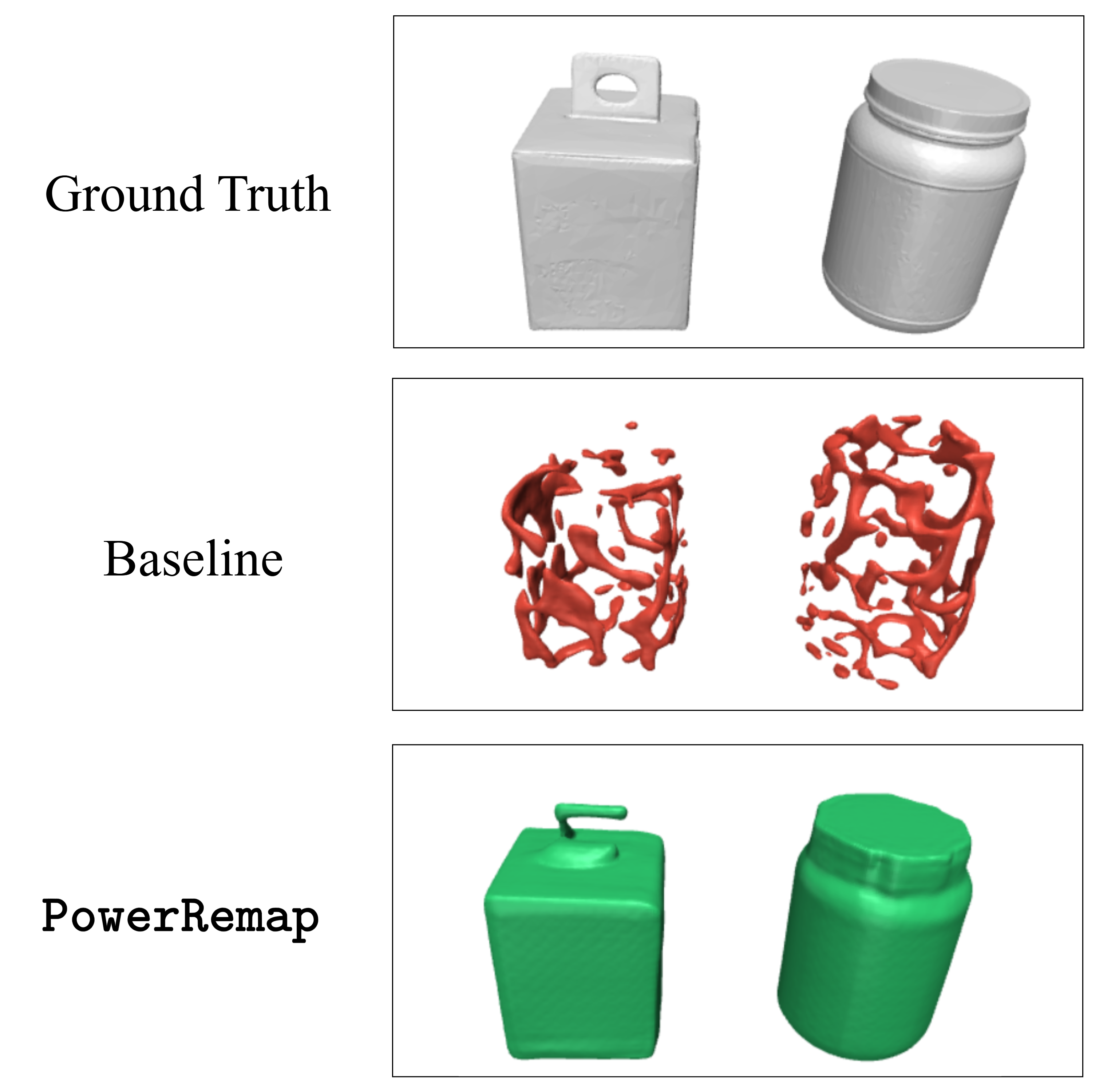}
    \caption{Qualitative evaluation of multi-object inputs. Meltdown persists in scenes with multiple objects, leading to severe degradation of the reconstruction, whereas \texttt{PowerRemap} effectively prevents this failure mode and yields a stable reconstruction of all objects.}
    \label{fig:multi_shape}
\end{figure}

We further assess whether the Meltdown phenomenon and the effectiveness of \texttt{PowerRemap} extend beyond single-object inputs. Figure~\ref{fig:multi_shape} provides a qualitative evaluation on a scene containing multiple objects for the \textsc{WaLa} model. We observe that Meltdown still occurs in this multi-object setting, while using \texttt{PowerRemap} reliably suppresses the failure and preserves a plausible reconstruction of all objects in the scene.

\clearpage
\subsection{Examining \texttt{PowerRemap} strength $\gamma$ on Reconstruction Connectivity} \label{app:choice_gamma_remedy}

We empirically investigate the influence of the \texttt{PowerRemap} strength $\gamma$ on reconstruction connectivity. Overall, we conclude that the optimal hyperparameter $\gamma$ is model-dependent.

\begin{figure}[h!]
    \centering
    \includegraphics[width=0.6\linewidth]{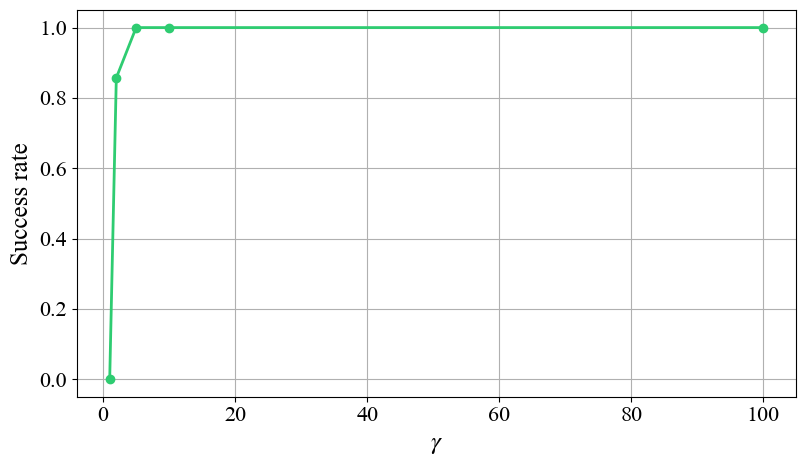}
    \caption{\texttt{WaLa}. We find that a \texttt{PowerRemap} strength of $\gamma >2$ remedies Meltdown.}
    \label{fig:wala_gamma_success}
\end{figure}

\paragraph{\textsc{WaLa}}
We run Algorithm~\ref{alg:alg_adverserial_Meltdown_search} on SimJEB (\textsc{WaLa}) shape~492 
for 10 independent random seeds and select the Meltdown configuration with $C_{\varepsilon} > 1$. We then apply \texttt{PowerRemap} to this configuration over the hyperparameter 
grid $\gamma \in \{1,2, 5, 10, 100\}$, where $\gamma=1$ denotes the identity mapping. As can be seen in Figure \ref{fig:wala_gamma_success}, our \texttt{PowerRemap} method achieves a high success rate for $\gamma >2$.

\begin{figure}[h!]
    \centering
    \includegraphics[width=0.6\linewidth]{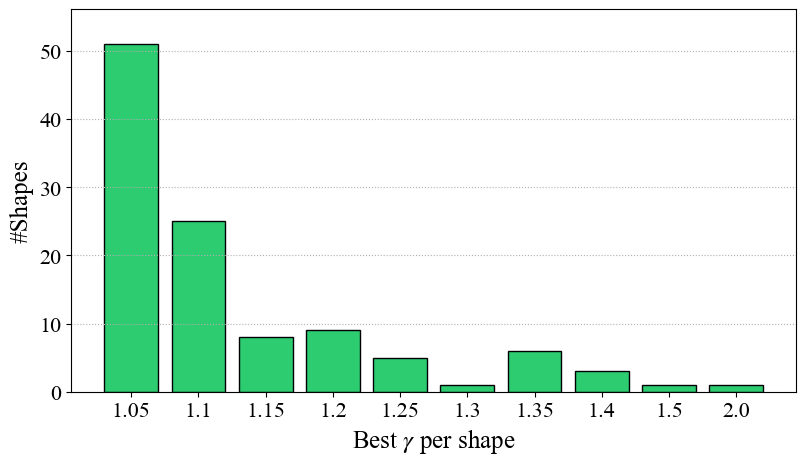}
    \caption{\textsc{Make-a-Shape}. For the representative subset in Table \ref{tab:gso_pr_category_top5_MAS_main} (top), we find that $\gamma$ values around 1.05 are effective to remedy Meltdown.}
    \label{fig:mas_best_gamma}
\end{figure}

\paragraph{\textsc{Make-a-Shape}}
For \textsc{Make-a-Shape}, we investigate the distribution of \texttt{PowerRemap} strengths $\gamma$ for the 130-shape subset of GSO as discussed in Table \ref{tab:gso_pr_category_top5_MAS_main} (top). We find that $\gamma$ values around 1.05 are effective to remedy Meltdown, as illustrated in Figure \ref{fig:mas_best_gamma}.

\clearpage
\subsection{Examining \texttt{PowerRemap} on Non-Meltdown Cases}
\label{app:non_meltdown_cases}

We empirically verify that \texttt{PowerRemap} does not interfere with  non-Meltdown runs. 
Specifically, we run Algorithm~\ref{alg:alg_adverserial_Meltdown_search} on SimJEB (\textsc{WaLa}) shape~492 
for 10 independent random seeds and select a configuration with $C_0 = 1$ (a single connected 
component). We then apply \texttt{PowerRemap} to this configuration over the hyperparameter 
grid $\gamma \in \{2, 5, 10, 100\}$. As can be seen in Table~\ref{tab:non_Meltdown_powerrmap}, the reconstructed surface retains 
$C_0 = 1$ in all cases and we do not observe any change in the final topology. The results in Table~\ref{tab:non_Meltdown_powerrmap} indicate that \texttt{PowerRemap} is effectively 
topologically neutral on this non-Meltdown instance.

\begin{table}[h!]
\centering
\caption{Evaluation of \texttt{PowerRemap} on a non-Meltdown SimJEB shape (shape 492). 
Across 10 random seeds and $\gamma \in \{2, 5, 10, 100\}$, \texttt{PowerRemap} preserves 
the original topology ($C_0 = 1$) in all cases.}
\label{tab:non_Meltdown_powerrmap}
\begin{tabular}{lrrr}
\toprule
\textbf{Shapes} & \textbf{Seeds} & \textbf{Meltdown occurs [\%]} & \textbf{Topology preserved [\%]} \\
\midrule
SimJEB 492          & 10             & 0.0                           & 100.0                            \\
\midrule
\textbf{Total}      & \textbf{10}    & \textbf{0.0}                  & \textbf{100.0}                   \\
\bottomrule
\end{tabular}
\end{table}

\clearpage

\section{Spectral analysis of the cross-attention write $\mathbf{Y}$}
\label{app:Y_spectral_analysis}

This appendix details the protocols and full numerical results behind the
\textsc{Make-a-Shape} and \textsc{WaLa} paragraphs in the main text. The two
models are analyzed with the same target (the block-4, first-denoising-step
cross-attention write $\mathbf{Y}$), but their architectural redundancy
differs, and the ablations that cleanly localize Meltdown differ accordingly.
We first describe the shared protocol
(\S\ref{app:Y_protocol}), then establish the architectural contrast
(\S\ref{app:Y_architecture}), then present the \textsc{Make-a-Shape}
experiment where direct spectral ablation is informative
(\S\ref{app:mas_spectral_ablation}), and finally the \textsc{WaLa}
experiments where direct ablation is confounded and a drift-targeted
protocol is required (\S\ref{app:wala_specificity}).

\subsection{Shared protocol}
\label{app:Y_protocol}

For each model, we identify the cross-attention write at block $4$ on the
first denoising step --- $\mathbf{Y}\equiv\mathbf{Y}_{4,7}$ for \textsc{WaLa}
($T{=}8$ DDIM steps) and $\mathbf{Y}\equiv\mathbf{Y}_{4,99}$ for
\textsc{Make-a-Shape} ($T{=}100$ DDIM steps), following the localization in
the main text. All interventions are single-forward-pass: we register a
forward hook on the corresponding sub-module
(\texttt{unet.blocks[4].cross\_attn.proj} in \textsc{WaLa};
\texttt{unet.middle\_blocks[4][0].att} in \textsc{Make-a-Shape}) that
overwrites the output at the target denoising step and lets the rest of
the reverse process run unmodified.

Each intervention is evaluated on $100$ diffusion seeds ($\{0,\ldots,99\}$)
at the model-specific phase-transition point: $\rho_{\mathrm{melt}}{=}0.4$
for \textsc{WaLa} and $\rho_{\mathrm{melt}}{=}0.5$ for
\textsc{Make-a-Shape}. Rescue is declared when the output mesh has a
single connected component ($C{=}1$); for \textsc{WaLa}, we additionally
require the chamfer distance to the healthy reference to be at least
$0.01$ below the unintervened baseline, to distinguish genuine sphere
recovery from trivial rank collapse. The baselines (unintervened at
$\rho_{\mathrm{melt}}$, averaged over the $100$ seeds) are:
\textsc{WaLa}: $\bar{C}{=}143.9$, $\overline{\text{chamfer}}{=}0.1122$,
rescue rate $0/100$; \textsc{Make-a-Shape}: $\bar{C}{=}124.9$, rescue
rate $25/100$ (stochasticity at the bifurcation is larger for MAS; see
main text \S\ref{sec:diffusion-dynamics} on ensemble behavior).

\subsection{Architectural contrast}
\label{app:Y_architecture}

The two models expose the encoded condition $\mathbf{C}$ to the denoising
backbone through different numbers of sites.

\paragraph{\textsc{WaLa}.} The denoiser is a
\texttt{Latent\_UVIT} (\texttt{network.\_orig\_mod.unet}) whose main
body is a \texttt{ModuleList} of $32$ sequential
\texttt{Cross\_DiTBlock}s. Each block contains, in order of the residual
stream update:
\begin{itemize}
\item a self-attention layer (\texttt{attn}) whose pre-norm
  (\texttt{LayerNorm(1152, elementwise\_affine=False)}) is AdaLN-modulated;
\item an MLP ($1152\to 4608\to 1152$) whose pre-norm is also
  AdaLN-modulated;
\item a cross-attention layer (\texttt{cross\_attn}, the site we study)
  reading $\mathbf{C}$ as keys/values.
\end{itemize}
The \texttt{adaLN\_modulation} submodule in each block is
\texttt{Linear(1152,6912)}, producing $6\times 1152$ scale/shift/gate
parameters ---  scale, shift, and gate for both the attention and the
MLP norms --- from $\mathbf{C}$-derived features (main text
\S\ref{sec:transformer}). Across all $32$ blocks, the condition therefore
enters the residual stream through $32$ cross-attention writes
\emph{and} $32$ AdaLN-modulation sites. A single
\texttt{cross\_attn} output is one of these $64$ depth-wise injection
points.

\paragraph{\textsc{Make-a-Shape}.} The denoiser is a
\texttt{Condition\_UVIT} (\texttt{network.unet}). Its bottleneck is
structured as two parallel module lists of length $8$:
\texttt{self\_middle\_blocks} (self-attention) and
\texttt{middle\_blocks} (cross-attention to $\mathbf{C}$). Each
\texttt{Transformer\_Block} in \texttt{middle\_blocks} contains an
attention layer (reading $\mathbf{C}$ as keys/values) and an MLP with
FiLM-style scale and shift produced by two \texttt{Linear(256,512)}
submodules from a $256$-dimensional embedding.
The condition therefore enters the backbone explicitly through the
$8$ cross-attention layers of \texttt{middle\_blocks}; no per-block
$\mathbf{C}$-dependent AdaLN is present.

\paragraph{Implication.} Zeroing the output of a single
\texttt{cross\_attn} removes $1/32$ of the direct cross-attention
pathways in \textsc{WaLa} and $1/8$ in \textsc{Make-a-Shape}, with the
former further buffered by $32$ independent AdaLN-modulation sites.
This $4\times$ (or larger, counting AdaLN) redundancy asymmetry is the
structural reason a single-write ablation is informative in
\textsc{Make-a-Shape} but confounded in \textsc{WaLa}, as the
experiments below confirm.

\subsection{\textsc{Make-a-Shape}: spectral ablation of $\mathbf{Y}$}
\label{app:mas_spectral_ablation}

\paragraph{Protocol.} For each seed we SVD-decompose
$\mathbf{Y}{=}U\Sigma V^\top$ at the target site and replace
$\mathbf{Y}$ with one of two rank-modified versions for
$m\in\{1,5,10,20,50,100\}$:
\begin{itemize}
\item \textbf{keep-top-$m$}: retain only the $m$ largest components,
  $\mathbf{Y}^{\text{keep}}_{(m)}{=}\sum_{i\leq m}\sigma_i u_i v_i^\top$
  (zeros the tail);
\item \textbf{zero-top-$m$}: retain only the tail,
  $\mathbf{Y}^{\text{zero}}_{(m)}{=}\sum_{i>m}\sigma_i u_i v_i^\top$
  (zeros the top).
\end{itemize}
These are run at every $m$ across all $100$ seeds ($1{,}200$ runs
plus $100$ unintervened baseline).

\paragraph{Results.} Rescue counts out of $100$ seeds (baseline:
$25/100$):

\begin{table}[h]
\centering
\small
\caption{\textsc{Make-a-Shape} spectral ablation at
$\rho_{\mathrm{melt}}{=}0.5$; $100$ seeds. Mean
connected-component count $\bar{C}$ in parentheses.}
\label{tab:mas_spectral_ablation}
\begin{tabular}{lcccccc}
\toprule
ablation $\backslash\ m$ & $1$ & $5$ & $10$ & $20$ & $50$ & $100$ \\
\midrule
keep-top-$m$ (zero tail) & $26$ $(127.5)$ & $\mathbf{99}$ $(2.75)$ & $\mathbf{99}$ $(2.67)$ & $61$ $(64.1)$ & $25$ $(124.8)$ & $25$ $(125.0)$ \\
zero-top-$m$ (zero top)  & $\phantom{0}8$ $(156.8)$ & $\phantom{0}4$ $(165.3)$ & $\phantom{0}5$ $(165.5)$ & $\phantom{0}8$ $(158.9)$ & $\phantom{0}8$ $(158.9)$ & $\phantom{0}8$ $(158.7)$ \\
\bottomrule
\end{tabular}
\end{table}

\paragraph{Three observations.} First, keep-top-$m$ is strongly
non-monotonic: at $m{=}1$ rescue ($26/100$) matches baseline
($25/100$) --- the rank-$1$ write carries too little to condition the
denoiser; at $m\in\{5,10\}$ rescue jumps to $99/100$ with
$\bar{C}{\approx}2.7$ (essentially a clean sphere); at $m\in\{50,100\}$
rescue returns \emph{exactly} to baseline with $\bar{C}$ matching the
unintervened distribution, meaning the intervention has no effect at
these ranks. This last fact implies the effective rank of $\mathbf{Y}$ at
this site is ${\leq}50$: retaining the top-$50$ of the spectrum is
equivalent to retaining the entire spectrum.

Second, zero-top-$m$ is uniformly destructive: rescue is $4$--$8\%$,
strictly below baseline, with $\bar{C}\in[156.8,\,165.5]$ exceeding
even the unintervened $\bar{C}{=}124.9$. Removing the top of the
spectrum makes outcomes worse than doing nothing.

Third, combining these: because keep-top-$50$ already reproduces
baseline, zero-top-$m$ at $m{\geq}50$ effectively sets
$\mathbf{Y}{=}\mathbf{0}$, and yields only $8/100$ rescues. In
\textsc{Make-a-Shape}, zeroing $\mathbf{Y}$ is actively harmful. This
is consistent with the architectural analysis in
\S\ref{app:Y_architecture}: $\mathbf{Y}$ is one of only $8$
cross-attention writes into the backbone and has no redundant
AdaLN fallback.

\paragraph{Interpretation.} The two ablations mirror each other on
opposite halves of the spectrum. Removing the tail while keeping an
informative top rescues Meltdown; removing the top (with or without
the tail) is uniformly worse than doing nothing. The causal
asymmetry is clean: the \emph{tail} of $\mathbf{Y}$ carries the
Meltdown-inducing signal, while the \emph{dominant} directions carry
the conditioning information the model cannot spare.

\subsection{\textsc{WaLa}: direct spectral ablation is non-informative}
\label{app:wala_direct_ablation}

\paragraph{Keep-top-$m$ on $\mathbf{Y}$'s own basis.} Using the same
truncation protocol as \S\ref{app:mas_spectral_ablation} for
$m\in\{1,2,3,5,10,15,20,25,30,40,50,75,100,150,200\}$:

\begin{table}[h]
\centering
\small
\caption{\textsc{WaLa} keep-top-$m$ on $\mathbf{Y}$'s own basis at
$\rho_{\mathrm{melt}}{=}0.4$; $100$ seeds. Baseline
$\bar{C}{=}143.9$, $\overline{\text{chamfer}}{=}0.1122$.}
\label{tab:wala_keep_top_m}
\begin{tabular}{lccccccccccc}
\toprule
$m$ & $1$ & $2$ & $3$ & $5$ & $10$ & $20$ & $30$ & $50$ & $100$ & $200$ \\
\midrule
rescue       & $62$   & $73$   & $0$     & $0$     & $0$     & $0$     & $0$     & $0$     & $0$     & $0$     \\
$\bar{C}$    & $1.06$ & $25.7$ & $142.8$ & $143.6$ & $144.0$ & $144.0$ & $143.6$ & $143.7$ & $143.9$ & $144.0$ \\
$\bar{\text{cham}}$ & $.100$ & $.100$ & $.113$ & $.112$ & $.112$ & $.112$ & $.112$ & $.112$ & $.112$ & $.112$ \\
\bottomrule
\end{tabular}
\end{table}

At every $m{\geq}3$, rescue is $0/100$ and both $\bar{C}$ and mean
chamfer are indistinguishable from the unintervened baseline. The
cases $m{\in}\{1,2\}$ rescue at $62/100$ and $73/100$, but these are
degenerate: at $m{=}1$ the mean chamfer is $0.0996$, closer to the
$\mathbf{Y}{=}0$ output than to the healthy reference ($0.0887$); a
rank-$1$ write carries essentially no content and the behaviour
reduces to the full $\mathbf{Y}{=}0$ ablation. The
\textsc{Make-a-Shape}-style tail-removal protocol therefore does not
transfer.

\paragraph{Zeroing $\mathbf{Y}$ entirely.} The fully-ablated condition
($\mathbf{Y}{=}\mathbf{0}$) \emph{does} rescue in \textsc{WaLa}. As
argued in \S\ref{app:Y_architecture}, this rescue follows from
architectural redundancy --- the other $31$ cross-attention writes
plus the $32$ AdaLN-modulation sites suffice to propagate the
condition --- rather than from any property of Meltdown. The direct
empirical contrast with \textsc{Make-a-Shape}, where the same
intervention is destructive (\S\ref{app:mas_spectral_ablation}),
makes this redundancy interpretation concrete.

\subsection{\textsc{WaLa}: targeted drift surgery}
\label{app:wala_drift_surgery}

To localize past the redundancy confound we target not $\mathbf{Y}$
itself but the \emph{drift} induced by $\rho$. Let
$\mathbf{Y}_c\,{=}\,\mathbf{Y}(\rho_{\mathrm{melt}})$,
$\mathbf{Y}_0\,{=}\,\mathbf{Y}(0)$, and
$d\mathbf{Y}\,{=}\,\mathbf{Y}_c - \mathbf{Y}_0$. Let $V_m^{d\mathbf{Y}}$
denote the top-$m$ right-singular subspace of $d\mathbf{Y}$. The
targeted surgery subtracts the portion of the drift that lives in its
own top-$m$ subspace:
\[
\mathbf{Y}_{\text{post}} \;=\; \mathbf{Y}_c \;-\; \Pi_{V_m^{d\mathbf{Y}}}\,(\mathbf{Y}_c - \mathbf{Y}_0).
\]
Rescue exhibits a sharp rank threshold:

\begin{center}
\small
\begin{tabular}{lcccc}
\toprule
$m$ & $5$ & $10$ & $20$ & $50$ \\
\midrule
rescue count            & $0/100$ & $76/100$ & $100/100$ & $100/100$ \\
$\bar{C}$               & $143.6$ & $24.0$   & $1.00$    & $1.00$    \\
$\bar{\text{chamfer}}$  & $0.1122$& $0.1008$ & $0.0887$  & $0.0887$  \\
\bottomrule
\end{tabular}
\end{center}

At $m{\geq}20$, the rescued chamfer matches the healthy reference
($0.0887$) to four decimals.

\subsection{\textsc{WaLa}: magnitude-matched direction controls}
\label{app:wala_specificity}

The drift surgery at $m{=}20$ perturbs $\mathbf{Y}$ with Frobenius
magnitude $\Delta_{20}\,{\approx}\,1.385\times 10^{6}$. A concern is
that any perturbation of comparable magnitude might rescue by pushing
the latent past an expendability threshold, independent of
direction. To test this, we construct four controls that match or
compare against this magnitude but differ in directional content.

\paragraph{Controls.} Let $d\mathbf{Y}\,{=}\,U_d\Sigma_d V_d^\top$.

\begin{description}
\item[\textbf{C1 (drift-orthogonal random subspace).}]
$\mathbf{Y}_{\text{post}}\,{=}\,\mathbf{Y}_c - U_{d,1:m}\,\Sigma_{d,1:m}\, R^\top$,
where $R\in\mathbb{R}^{d\times m}$ is an orthonormal basis of a
randomly sampled $m$-dimensional subspace of the orthogonal
complement of $V_{d,1:m}$, produced by QR on a Gaussian matrix
projected off $V_{d,1:m}$. This preserves the left factors $U_d$, the
singular values $\Sigma_d$, and $\Delta_m$ exactly; only the
feature-space directions $V$ are rotated. The orthogonality residual
$\|V_{d,1:m}^\top R\|$ across all $1{,}200$ runs has mean
$5.6\times 10^{-7}$ and maximum $1.6\times 10^{-6}$. Three independent
$R$ draws per $(\text{seed},m)$.
\item[\textbf{C2 (scalar attenuation).}]
$\mathbf{Y}_{\text{post}}\,{=}\,\alpha\mathbf{Y}_c$ with
$\alpha\,{=}\,\max\{0,\,1-\Delta_m/\|\mathbf{Y}_c\|_F\}$. Preserves the
\emph{shape} of $\mathbf{Y}_c$'s spectrum exactly (every singular
value is scaled by $\alpha$); represents the ``shrink uniformly
toward zero'' direction. One run per $(\text{seed},m)$.
\item[\textbf{C3 (isotropic Gaussian noise).}]
$\mathbf{Y}_{\text{post}}\,{=}\,\mathbf{Y}_c + \epsilon_m G$,
$G\sim\mathcal{N}(0,I)$ i.i.d.\ entrywise, $\epsilon_m$ chosen so
$\|\epsilon_m G\|_F\,{=}\,\Delta_m$. No directional structure. Three
independent $G$ draws per $(\text{seed},m)$.
\item[\textbf{C4 (keep-top-$m$ on $\mathbf{Y}$'s own basis).}]
The intervention of \S\ref{app:wala_direct_ablation}, repeated here
for completeness. Its Frobenius perturbation
$\|\mathbf{Y}_c-\mathbf{Y}^{\text{keep}}_{(m)}\|_F$ is determined by
$\mathbf{Y}_c$'s own tail energy and is not explicitly matched to
$\Delta_m$; we include C4 because it is the architecture-aligned
structural ablation, not because it is magnitude-matched.
\end{description}

For C1--C3, the Frobenius match is exact by construction up to the
orthogonality residual of C1: targeted, C2, and C3 all share
$\Delta_m\,{=}\,1.38465\times 10^6$ at $m{=}20$ to machine precision,
while C1 deviates by ${<}1$ part in $10^{5}$.

\paragraph{Results.}

\begin{table}[h]
\centering
\small
\caption{\textsc{WaLa} rescue counts out of $100$ seeds.
Randomized controls (C1, C3) are aggregated over $3$ replicates per
$(\text{seed},m)$. All controls fail at every rank tested.}
\label{tab:wala_specificity}
\begin{tabular}{lcccc}
\toprule
variant $\backslash\ m$ & $5$ & $10$ & $20$ & $50$ \\
\midrule
targeted drift surgery                              & $0$ & $76$ & $\mathbf{100}$ & $\mathbf{100}$ \\
C1: drift-orth.\ random subspace ($n\,{=}\,300$/rank) & $0$ & $0$  & $0$            & $0$            \\
C2: scalar attenuation ($n\,{=}\,100$/rank)         & $0$ & $0$  & $0$            & $0$            \\
C3: Gaussian noise ($n\,{=}\,300$/rank)             & $0$ & $0$  & $0$            & $0$            \\
C4: keep-top-$m$ on $\mathbf{Y}$ ($n\,{=}\,100$/rank)& $0$ & $0$  & $0$            & $0$            \\
\bottomrule
\end{tabular}
\end{table}

Aggregated across the four ranks: C1 rescues $0/1200$, C2 rescues
$0/400$, C3 rescues $0/1200$, C4 rescues $0/400$; across the entire
control suite $0/3200$ runs produce a sphere, while the targeted
surgery reaches $100/100$ at $m{\geq}20$. Rescue at $\mathbf{Y}$ is
therefore direction-specific, not magnitude-driven. C2 in particular
rules out the ``$\mathbf{Y}{=}0$ just shrinks the write'' hypothesis:
at matched magnitude, the uniform-shrink direction never rescues.

\subsection{\textsc{WaLa}: $H$ is a correlate, not a cause}
\label{app:wala_H_not_cause}

Main-paper Section~\ref{sec:patching_effect} identifies the spectral
entropy $H(\mathbf{Y})$ as a scalar that rises smoothly with $\rho$
alongside the discontinuous jump in $C$. The matched-magnitude
controls allow us to test whether $H$ is the causal variable.
Post-intervention values of $H(\mathbf{Y}_{\text{post}})$ at $m{=}20$
(mean over $100$ seeds; standard deviations all
${<}5\times 10^{-3}$):

\begin{center}
\small
\begin{tabular}{lcc}
\toprule
variant                                       & $H$ at $m{=}20$ & rescue \\
\midrule
targeted drift surgery                        & $1.613$ & $100/100$ \\
scalar attenuation (C2)                       & $1.639$ & $0/100$   \\
random subspace (C1)                          & $1.779$ & $0/100$   \\
Gaussian noise (C3)                           & $1.884$ & $0/100$   \\
keep-top-$20$ on $\mathbf{Y}$ (C4)            & $1.582$ & $0/100$   \\
\bottomrule
\end{tabular}
\end{center}

Because C2 preserves the shape of the spectrum exactly, its
$H$-value ($1.639$) is the corrupt-baseline entropy
$H(\mathbf{Y}_c)$; this is stable to four decimals across $m$ as
expected. The targeted surgery reduces $H$ by only $0.026$ and
rescues. C4 drives $H$ \emph{below} the rescuing value
($1.582<1.613$) yet fails in all $100$ seeds. C1 and C3 push $H$
well \emph{above} the corrupt baseline ($1.779$, $1.884$) and also
fail. Matched-magnitude interventions therefore move $H$ in both
directions without rescuing, while the successful intervention
leaves $H$ essentially unchanged. $H$ cannot be the causal variable:
it tracks proximity to the transition, but the directional content
of $\mathbf{Y}$ decides which basin the trajectory commits to.

\subsection{Joint interpretation}
\label{app:Y_joint_summary}

Both models localize Meltdown to the cross-attention write at
$(\text{block}\ 4, \text{step}\ T)$, and in both, \emph{spectral}
structure of $\mathbf{Y}$ carries the causal signal. The protocols
differ because the architectures differ.

In \textsc{Make-a-Shape}, $\mathbf{Y}$ is one of only $8$
cross-attention writes and has no $\mathbf{C}$-dependent AdaLN
fallback. Zeroing the write is actively
destructive. Directly splitting the spectrum of $\mathbf{Y}$ with SVD
reveals that its \emph{tail} carries the Meltdown-inducing component
while its \emph{top} carries the indispensable conditioning signal;
this is the MAS result.

In \textsc{WaLa}, $\mathbf{Y}$ is one of $32$ cross-attention writes
with a further $32$ AdaLN-modulation pathways in parallel; the write
is dispensable as a whole, so the direct analog of the MAS ablation
does not localize anything. Targeting the \emph{drift}
$d\mathbf{Y}\,{=}\,\mathbf{Y}(\rho_{\mathrm{melt}}) - \mathbf{Y}(0)$
instead of $\mathbf{Y}$ itself, and rigorously controlling for
Frobenius magnitude with four complementary controls, shows that a
specific low-rank ($m{\geq}20$) directional correction to the write
rescues all $100$ seeds, while no matched-magnitude control rescues
any seed in $3{,}200$ runs. Spectral entropy $H$, which rises with
$\rho$ alongside the $C$ discontinuity, cannot be the causal variable:
the controls push $H$ both above and below the rescuing value without
rescuing, while the successful surgery leaves $H$ unchanged.

Across both models, the overarching conclusion is the same --- the
cross-attention write at block $4$ of the first denoising step
contains a direction-specific Meltdown signature that can be
surgically neutralized.

%

%
%

\newpage

\section{Exhaustive Within-Block Causal Scan}\label{app:act_patching_extended}

This appendix gives the activation-patching procedure used throughout
the paper in full detail, and reports an exhaustive within-block scan
that tests whether the cross-attention restriction adopted in
Section~\ref{sec:activation_patching} is consistent with the network's
causal map. The same algorithm underlies both analyses: in the main
text it is instantiated for the cross-attention output projection
${\mathbf{Y}}$ alone; in this appendix it is run for
every accessible intermediate activation of every DiT block at every
denoising step.

\paragraph{TL;DR.}
The scan covers $32$ blocks $\times\,8$ denoising steps $\times\,19$
intermediate activations per block, for $4{,}864$ patches per diffusion
seed. Coverage includes three diffusion seeds.
Three findings:
\begin{itemize}\setlength\itemsep{1.5pt}
  \item Of the $32 \cdot 8 = 256$ depth-time cells, $8$ admit a
        within-block (non-residual-stream) rescue. Exactly one of these
        cells is the canonical $(k{=}4,\,t{=}7)$ identified in the main
        text. At that cell, the cross-attention output projection
        (${\mathbf{Y}}$, observed at the three
        equivalent sites \texttt{ca\_preproj}, \texttt{Y\_kt},
        \texttt{ca\_module\_out}) is the unique within-block rescue:
        the other $14$ sites at the cell -- comprising the eight
        self-attention-branch sites, the AdaLN modulation, the four
        CA-pre-write sites, and the two MLP sites -- all fail. The
        remaining $7$ within-block rescues across the grid lie at
        $(k{=}0,\,t\in\{2,\dots,6\})$ and $(k{=}31,\,t\in\{1,2\})$ and
        involve only the post-CA-MLP path, never cross-attention.
  \item Self-attention-branch patches and AdaLN-modulation patches
        produce $0$ rescues at any cell of the grid. Cross-attention
        pre-write patches (\texttt{ca\_module\_in},
        \texttt{ca\_q\_postnorm}, \texttt{ca\_k\_postnorm}) likewise
        produce $0$ rescues at any cell. The cross-attention output
        projection is the only cross-attention-branch site that ever
        rescues, and only at $(k{=}4,\,t{=}7)$.
  \item At $t{=}7$, the depth axis partitions into three contiguous
        zones: a pre-commit zone ($k\in\{0,1,2,3\}$) where no patch of
        any kind rescues, including replacement of the residual stream
        entering block~$3$; a commit point ($k=4$) where the
        cross-attention output projection rescues; and a propagation
        zone ($k\in\{5,\dots,31\}$) where only residual-stream patches
        rescue. The CA-write rescue at the commit point and the
        residual-stream rescues throughout the propagation zone produce
        reconstructions of equivalent sphere quality (chamfer
        $0.0886$ vs $0.0896 \pm 0.0014$, both marginally below the
        $0.0916$ clean baseline; radial standard deviation $0.006$;
        sphericity proxy $1.000$), with no depth-dependent drift across
        the $28$ blocks of the propagation zone. The seven off-canonical
        within-block rescues exhibit degraded geometry (chamfer
        $0.097$--$0.098$, sphericity proxy $1.02$--$1.05$).
\end{itemize}
The findings identify $(k{=}4,\,t{=}7)$ as the unique cell on the grid
at which a single submodule -- the cross-attention output projection
${\mathbf{Y}}$ -- causally controls Meltdown with
full sphere quality, and identify SA, AdaLN, MLP, and CA pre-write
activations as either causally inert (SA, AdaLN, CA pre-write) or
causally peripheral (MLP, with $7$ off-canonical, geometrically
inferior rescues). The cross-attention restriction in
Section~\ref{sec:activation_patching} targets the only cell on the
grid that meets these conditions. The remainder of this appendix
supplies the procedure, the site enumeration, and the per-cell tables
that support these claims.

\subsection{Protocol}\label{app:scan_protocol}

The patching procedure is parameterized by a single \emph{site} $s$,
an intermediate activation of a DiT block.
Algorithm~\ref{alg:activation_patching} specifies the procedure for an
arbitrary site $s$: it caches the activation at $s$ from a forward
pass on the healthy point cloud at every depth-time cell $(k,t)$, runs
an unhealthy forward pass for each cell with the cached value
reinjected at that single cell, and records the resulting
connected-component count $C$. In the main text
(Section~\ref{sec:activation_patching}) the procedure is instantiated
with $s = {\mathbf{Y}}$, the cross-attention output
projection. In the present appendix the same algorithm is run for
each of the $19$ within-block activations listed in
Table~\ref{tab:site_enumeration}, producing the $4{,}864$-patch rescue
map per seed summarized above.

\begin{algorithm}[t]
\caption{Localizing Meltdown via activation patching at site $s$.}
\label{alg:activation_patching}
\begin{algorithmic}[1]
    \Require Encoder $E$; latent diffusion transformer $B$ with $K$ blocks; decoder $D$; healthy point-cloud $\mathcal{P}$; unhealthy point-cloud $\mathcal{Q}$; site $s$ (any one of the within-block activations of Table~\ref{tab:site_enumeration})
    \State $\mathbf{Z}^{0}_{T} \sim \mathcal{N}(0, I)$ \Comment{sample initial noise}
    \Statex \textbf{Record healthy activations:}
    \State $\mathbf{C}_\mathcal{P} \gets E(\mathcal{P})$
    \For{$t = T : 1$}
        \For{$k = 0 : K-1$}
            \State $\mathbf{Z}^{k+1}_t \gets B^k(\mathbf{Z}^{k}_t, \mathbf{C}_\mathcal{P})$ and record activation at site $s$ as $X^{\mathrm{healthy}}_{k,t,s}$
        \EndFor
        \State $\mathbf{Z}_{t-1}^0 \gets \mathrm{DDIM}(\mathbf{Z}_{t}^{K-1})$ \Comment{discrete denoising update}
    \EndFor
    \Statex \textbf{Patch unhealthy activations:}
    \State $\mathbf{C}_\mathcal{Q} \gets E(\mathcal{Q})$
    \For{$t' = T : 1$} \Comment{denoising substitution loop}
        \For{$k' = 0 : K-1$} \Comment{block substitution loop}
            \For{$t = T : 1$}
                \For{$k = 0 : K-1$}
                    \State $\mathbf{Z}^{k+1}_t \gets B^k(\mathbf{Z}^{k}_t, \mathbf{C}_\mathcal{Q})$, overwriting site $s$ with $X^{\mathrm{healthy}}_{k,t,s}$ \textbf{if} $t = t'$ \textbf{and} $k = k'$
                \EndFor
                \State $\mathbf{Z}_{t-1}^0 \gets \mathrm{DDIM}(\mathbf{Z}_{t}^{K-1})$
            \EndFor
            \State $C_{k', t', s} \gets C(D(\mathbf{Z}_0^{K-1}))$ \Comment{decode and count components after patch}
        \EndFor
    \EndFor
    \State \Return repair map $\{C_{k, t, s}\}_{k=0:K-1,\,t=1:T}$
\end{algorithmic}
\end{algorithm}

A patch is judged a rescue when $C = 1$ and the mesh has at least
$200$ faces. This is the same connectivity criterion as
Section~\ref{sec:activation_patching}; geometric quality (chamfer,
mean radius, radial standard deviation, sphericity proxy) is recorded
as auxiliary information and used in
Section~\ref{app:rescue_quality} to stratify rescues. Conditions:
\textsc{WaLa}, $\rho_{\mathrm{crit}} = 0.4$, $N_{\mathrm{points}} = 400$,
DDIM sampling with $T = 8$ denoising steps. The patch acts only on
the conditional half of the classifier-free batch
\citep{heimersheim2024useinterpretactivationpatching}.

\subsection{Site enumeration}\label{app:site_enumeration}

A DiT block in \textsc{WaLa} (Eq.~\ref{eq:block_a}--\ref{eq:block_c})
exposes the $19$ intermediate activations listed in
Table~\ref{tab:site_enumeration}. Sites are grouped by site class and
ordered along the dataflow direction. Three groups of nominally
distinct sites correspond to single dataflow paths: the block boundary
$\texttt{block\_out}_k = \texttt{x\_res}_{k+1}$; the cross-attention
output projection
$(\texttt{ca\_preproj},\texttt{Y\_kt},\texttt{ca\_module\_out})$,
which is the same physical tensor (the projection's input, output, and
the CA module's output coincide once the projection is the module's
last operation, as verified by post-patch chamfer values agreeing to
six decimal places); and the post-CA MLP path
$(\texttt{norm2\_out},\texttt{mlp\_in},\texttt{mlp\_out})$, three
adjacent sites whose patches converge to numerically near-identical
post-patch reconstructions where they rescue.

\begin{table}[h!]
\centering
\small
\setlength{\tabcolsep}{6pt}
\renewcommand{\arraystretch}{1.05}
\begin{tabular}{@{}llll@{}}
\toprule
\textbf{Site} & \textbf{Class} & \textbf{Hook} & \textbf{Dataflow position} \\
\midrule
\texttt{x\_res}          & block input    & pre  & block input residual $\mathbf{Z}^{k}$  \\
\midrule
\texttt{ada}             & AdaLN params   & post & AdaLN modulation parameters \\
\midrule
\texttt{norm1\_out}      & SA branch      & post & first LayerNorm output (pre SA-AdaLN) \\
\texttt{sa\_module\_in}  & SA branch      & pre  & input to SA module ($\mathring{\mathbf{Z}}$) \\
\texttt{sa\_q\_postnorm} & SA branch      & post & SA query post-norm  \\
\texttt{sa\_k\_postnorm} & SA branch      & post & SA key post-norm    \\
\texttt{sa\_preproj}     & SA branch      & pre  & input to SA output projection \\
\texttt{sa\_proj\_out}   & SA branch      & post & output of SA output projection \\
\texttt{sa\_module\_out} & SA branch      & post & output of SA module ($\mathring{\mathbf{Y}}$) \\
\midrule
\texttt{norm2\_out}      & CA pre-write   & post & second LayerNorm output (pre CA-AdaLN) \\
\texttt{ca\_module\_in}  & CA pre-write   & pre  & input to CA module (${\mathbf{Z}}$) \\
\texttt{ca\_q\_postnorm} & CA pre-write   & post & CA query post-norm  \\
\texttt{ca\_k\_postnorm} & CA pre-write   & post & CA key post-norm    \\
\midrule
\texttt{ca\_preproj}     & CA write       & pre  & input to CA output projection \\
\texttt{Y\_kt}           & CA write       & post & output of CA output projection (${\mathbf{Y}}$) \\
\texttt{ca\_module\_out} & CA write       & post & output of CA module (${\mathbf{Y}}$) \\
\midrule
\texttt{mlp\_in}         & MLP branch     & pre  & input to MLP module \\
\texttt{mlp\_out}        & MLP branch     & post & output of MLP module ($\bar{\mathbf{Y}}$) \\
\midrule
\texttt{block\_out}      & block output   & post & block output residual $\mathbf{Z}^{k+1}$ \\
\bottomrule
\end{tabular}
\caption{The $19$ within-block sites covered by the scan and admitted
by Algorithm~\ref{alg:activation_patching} as the parameter $s$.
Hooks marked \emph{pre} replace the input to the named module; hooks
marked \emph{post} replace its output. Dataflow position references
the corresponding tensor in Eq.~(\ref{eq:block_a})--(\ref{eq:block_c})
where applicable. The main-text instantiation
(Section~\ref{sec:activation_patching}) uses $s = \texttt{Y\_kt}$.}
\label{tab:site_enumeration}
\end{table}

\subsection{Spatial structure at $t{=}7$}\label{app:scan_t7}

Holding the denoising step at the canonical $t=7$, the per-block
rescue map partitions the depth axis into three contiguous zones.
Table~\ref{tab:t7_per_block} lists, for every block $k\in\{0,\dots,31\}$,
the subset of the $19$ within-block sites whose patching rescues.

\begin{table}[h!]
\centering
\small
\setlength{\tabcolsep}{6pt}
\renewcommand{\arraystretch}{1.05}
\begin{tabular}{@{}lll@{}}
\toprule
\textbf{Blocks} & \textbf{Zone} & \textbf{Rescuing sites at $t=7$} \\
\midrule
$k\in\{0,1,2\}$           & pre-commit   & none of the $19$ sites \\
$k=3$                     & pre-commit   & \texttt{block\_out} only \\
$k=4$                     & commit       & \texttt{x\_res}, \texttt{ca\_preproj}, \texttt{Y\_kt}, \\
                          &              & \texttt{ca\_module\_out}, \texttt{block\_out} \\
$k\in\{5,\dots,31\}$      & propagation  & \texttt{x\_res}, \texttt{block\_out} \\
\bottomrule
\end{tabular}
\caption{Sites whose patches rescue at $t{=}7$, by block. Identical
across all available seeds. The asymmetry at $k=3$ is mechanical:
$\texttt{block\_out}_3 = \texttt{x\_res}_4$ as a single physical tensor,
so the residual-stream cut between blocks $3$ and $4$ is realized by a
successful patch from either side.}
\label{tab:t7_per_block}
\end{table}

\paragraph{Pre-commit zone ($k\in\{0,1,2,3\}$).}
Patching $\texttt{x\_res}_k$, $\texttt{block\_out}_k$, or any of the
$17$ within-block sites at any of these blocks fails to rescue, with
the single exception of $\texttt{block\_out}_3 = \texttt{x\_res}_4$.
Of $4 \cdot 19 = 76$ unique site-patches in this zone, $1$ rescues, and
that one is the residual stream entering block~$4$. The Meltdown
signal is not localizable to any submodule of blocks $0$--$3$ at
$t=7$ and is not yet committed to the residual stream upstream of
block~$4$.

\paragraph{Commit point ($k=4$).}
Five sites rescue at $(k{=}4,t{=}7)$: the residual-stream sites
$\texttt{x\_res}_4$ and $\texttt{block\_out}_4$, and the
cross-attention output-projection triple
$(\texttt{ca\_preproj},\texttt{Y\_kt},\texttt{ca\_module\_out})$. The
remaining $14$ within-block sites at this cell do not rescue: the
seven SA-branch sites
($\texttt{norm1\_out}$, $\texttt{sa\_module\_in}$, $\texttt{sa\_q\_postnorm}$,
$\texttt{sa\_k\_postnorm}$, $\texttt{sa\_preproj}$, $\texttt{sa\_proj\_out}$,
$\texttt{sa\_module\_out}$);
the AdaLN parameter site $\texttt{ada}$;
the four CA-pre-write sites
($\texttt{norm2\_out}$, $\texttt{ca\_module\_in}$,
$\texttt{ca\_q\_postnorm}$, $\texttt{ca\_k\_postnorm}$);
and the two MLP sites ($\texttt{mlp\_in}$, $\texttt{mlp\_out}$).
The post-patch chamfer is identical to six decimal places across the
three CA-write sites in each available seed
(e.g.\ $0.088541$ in seed~$0$; $0.088708$ in seed~$1$),
confirming that they observe a single physical tensor.

\paragraph{Propagation zone ($k\in\{5,\dots,31\}$).}
At every block in this range, only the residual-stream sites
$\texttt{x\_res}_k$ and $\texttt{block\_out}_k$ rescue. Of the
$27 \cdot 17 = 459$ within-block (non-residual) site-patches in this
zone, $0$ rescue. In particular, $\texttt{Y\_kt}_{k,7}$ for
$k\in\{5,\dots,31\}$ never rescues. The Meltdown signal, once committed
at block~$4$, is carried by the residual stream rather than by any
internal computation of the downstream blocks; cleaning a single
submodule's output downstream of the commit point does not undo it.

\subsection{Temporal structure at $k=4$}\label{app:scan_k4}

Holding the block at the canonical $k=4$, the rescue pattern across
denoising steps is given in Table~\ref{tab:k4_per_step}.

\begin{table}[h!]
\centering
\small
\setlength{\tabcolsep}{6pt}
\renewcommand{\arraystretch}{1.05}
\begin{tabular}{@{}lll@{}}
\toprule
\textbf{Step} & \textbf{Regime} & \textbf{Rescuing sites at $k=4$} \\
\midrule
$t\in\{0,1\}$           & closed       & none of the $19$ sites \\
$t\in\{2,\dots,6\}$     & open         & \texttt{x\_res}, \texttt{block\_out} \\
$t=7$                   & commit       & \texttt{x\_res}, \texttt{ca\_preproj}, \texttt{Y\_kt}, \\
                        &              & \texttt{ca\_module\_out}, \texttt{block\_out} \\
\bottomrule
\end{tabular}
\caption{Sites whose patches rescue at $k=4$, by denoising step.
$\texttt{Y\_kt}_{4,t}$ rescues only at $t=7$. The combination of the
spatial localization to $k=4$ at $t=7$ (Section~\ref{app:scan_t7}) and
the temporal localization to $t=7$ at $k=4$ identifies $(k{=}4,t{=}7)$
as the only cell on the $32{\times}8$ grid at which a cross-attention
write rescues.}
\label{tab:k4_per_step}
\end{table}

The temporal pattern at $k=4$ generalises: across the entire grid, no
within-block site rescues at $t=0$, and only the off-canonical
$(k{=}31,t{=}1)$ MLP-path pocket rescues at $t=1$
(Section~\ref{app:mlp_pockets}). The trajectory thus appears closed at
the last two denoising steps and progressively more localized in
preceding steps.

\subsection{Geometric quality of rescues}\label{app:rescue_quality}

The connectivity criterion ($C=1$, $\geq 200$ faces) is a topological
test. A rescued mesh can have $C=1$ while differing from the clean
output in size, shape, or surface uniformity.
Table~\ref{tab:rescue_quality} reports the four geometric statistics
across the rescue categories identified in
Sections~\ref{app:scan_t7}--\ref{app:scan_k4}.

\begin{table}[h!]
\centering
\small
\setlength{\tabcolsep}{4.5pt}
\renewcommand{\arraystretch}{1.1}
\begin{tabular}{@{}llcccc@{}}
\toprule
\textbf{Category} & \textbf{Description} & $n$
   & \textbf{chamfer} & \textbf{mean radius} & \textbf{radial std} \\
\midrule
A & CA-write trio at $(k{=}4,t{=}7)$               & $6$
   & $0.0886\pm 0.0001$ & $0.9135\pm 0.0001$ & $0.0058$ \\
B & residual stream at $(k{=}4,t{=}7)$             & $4$
   & $0.0887\pm 0.0001$ & $0.9133\pm 0.0000$ & $0.0058$ \\
C & residual stream at $(k{=}3,t{=}7)$             & $2$
   & $0.0887\pm 0.0000$ & $0.9134\pm 0.0000$ & $0.0059$ \\
D & residual stream, propagation zone ($k{\geq}5,t{=}7$) & $72$
   & $0.0896\pm 0.0014$ & $0.9124\pm 0.0015$ & $0.0062$ \\
E & residual stream, open window ($t\in\{2,\dots,6\}$)  & $469$
   & $0.0977\pm 0.0030$ & $0.9037\pm 0.0035$ & $0.0119$ \\
F & MLP-path pocket at $(k{=}0,t\in\{2,\dots,6\})$ & $30$
   & $0.0981\pm 0.0034$ & $0.9031\pm 0.0041$ & $0.0126$ \\
G & MLP-path pocket at $(k{=}31,t\in\{1,2\})$      & $6$
   & $0.0967\pm 0.0101$ & $0.9037\pm 0.0116$ & $0.0157$ \\
\midrule
  & clean baseline ($\rho=0$)                      & --
   & $0.0916$           & $0.9104$           & $0.0058$ \\
  & corrupt baseline ($\rho=0.4$)                  & --
   & $0.1118$           & --                 & --       \\
\bottomrule
\end{tabular}
\caption{Geometric statistics of rescued meshes by category. Reported
as mean $\pm$ standard deviation, aggregated over all rescue rows in
each category and over the available seeds. Sphericity proxy (surface
area divided by $4\pi r^2$) is $1.000\pm 0.000$ for categories A--D and
$1.015$--$1.05$ for categories E--G; omitted from the table for
compactness. Categories A--D have chamfer slightly below the clean
baseline; this reflects the chamfer being measured against an ideal
$10{,}000$-point unit sphere, with respect to which the patched runs
at $t{=}7$ produce reconstructions marginally closer to ideal than the
unperturbed clean run.}
\label{tab:rescue_quality}
\end{table}

The categories partition into two regimes. Categories A--D, which
share the canonical denoising step $t{=}7$, have chamfer
$0.089 \pm 0.002$, mean radius $0.913 \pm 0.002$, radial standard
deviation $0.006 \pm 0.001$, and sphericity proxy $1.000 \pm 0.000$ --
matching or marginally exceeding the clean baseline on every statistic.
Categories E--G, which share the property of patching at non-canonical
denoising steps, have chamfer $0.097 \pm 0.003$, mean radius
$0.903 \pm 0.004$, radial standard deviation $0.013 \pm 0.005$, and
sphericity proxy $1.02 \pm 0.02$ -- recovering connectivity but
producing a slightly smaller, less uniform, less spherical
reconstruction than the clean baseline.

Two consequences for the cross-attention restriction. First, within
the canonical step, replacing only the cross-attention output
projection at the commit point (category A, $n{=}6$, mean chamfer
$0.0886$) is geometrically equivalent to replacing the entire block
input or output residual at the same cell (category B, $n{=}4$, mean
chamfer $0.0887$): the chamfer means differ by $10^{-4}$. The
cross-attention write therefore carries the full structural content
of the residual stream at the commit point; the SA, AdaLN, and MLP
contributions to block~$4$'s residual update at $t{=}7$ are not
geometrically informative once ${\mathbf{Y}}_{4,7}$
is correct. Second, the propagation-zone rescues (category D)
reproduce the canonical rescue's chamfer to within $10^{-3}$ across
all $28$ blocks, with no detectable depth-dependent drift; the
residual stream downstream of the commit point transports the
committed signal without modification. Both observations sharpen the
interpretation of the cross-attention output projection at
$(k{=}4,t{=}7)$ as the unique commit lever, with downstream blocks
acting as transport rather than as additional sources of structural
information.

\subsection{Off-canonical MLP-path pockets}\label{app:mlp_pockets}

Two pockets of within-block rescue lie outside the canonical Meltdown
circuit. Both involve the post-CA MLP path
$(\texttt{norm2\_out},\texttt{mlp\_in},\texttt{mlp\_out})$, with all
three sites rescuing together:
\begin{itemize}\setlength\itemsep{1.5pt}
  \item $(k{=}0,\,t\in\{2,3,4,5,6\})$: $5$ cells at the first DiT
        block, across the open temporal window.
  \item $(k{=}31,\,t\in\{1,2\})$: $2$ cells at the last DiT block, at
        the boundary of the closed and open windows.
\end{itemize}
The pockets are spatially disjoint from the canonical cell
($k\in\{0,31\}$ versus $k=4$), temporally disjoint
($t\in\{1,\dots,6\}$ versus $t=7$), modulewise disjoint (MLP path
versus cross-attention output projection), and geometrically inferior
(categories F and G in Table~\ref{tab:rescue_quality}). They are not
part of the Meltdown circuit localized in
Section~\ref{sec:activation_patching}.

\subsection{Converse direction: noising scan}\label{app:noising_scan}

Sections~\ref{app:scan_t7}--\ref{app:mlp_pockets} report the
\emph{rescue} scan: at every cell on the depth-time grid, replace the
unhealthy activation with the cached healthy one, and ask whether
connectivity is restored. We now report the converse: at every cell,
replace the healthy activation with the cached unhealthy one, and ask
whether the run fragments. The protocol is otherwise identical to
Algorithm~\ref{alg:activation_patching}, with the roles of $\mathcal{P}$
and $\mathcal{Q}$ swapped. The verdict has two tiers, mirroring the
rescue verdict: \emph{lax noise} requires the output to be fragmented
($C>1$, $\geq 200$ faces). Coverage:
the full $32 \cdot 8 \cdot 19 = 4{,}864$ patches at three diffusion seeds,
$\rho_{\mathrm{crit}} = 0.4$, $N_{\mathrm{points}} = 400$, DDIM with
$T = 8$.

\paragraph{TL;DR.} Three findings, all of which mirror or sharpen
claims made in the rescue direction.
\begin{itemize}\setlength\itemsep{1.5pt}
  \item \emph{The commit lever is asymmetric.} At the canonical cell
        $(k{=}4,\,t{=}7)$, the cross-attention output projection
        $(\texttt{ca\_preproj},\texttt{Y\_kt},\texttt{ca\_module\_out})$
        does not noise: replacing the healthy $\mathbf{Y}_{4,7}$ with the
        cached unhealthy value produces a clean sphere
        ($C=1$, chamfer $0.0893$, sphericity $1.000$), indistinguishable
        from the unintervened clean run. The same holds across the
        entire grid: $0/256$ lax-noise events for $\mathbf{Y}_{k,t}$ at
        any $(k,t)$, and $0/768$ across all three CA-write sites. Y is a
        sufficient lever in the rescue direction (Section~\ref{app:scan_t7})
        but not in the noise direction.
  \item \emph{SA, AdaLN, and CA pre-write are inert in both directions.}
        Self-attention-branch patches yield $0/1{,}792$ noise events;
        AdaLN-modulation patches yield $0/256$; CA pre-write patches
        yield $0/1{,}024$ at the canonical cell and $8/1{,}024$ overall,
        all confined to $(k{=}31)$ on \texttt{norm2\_out} (see fourth
        bullet). These site classes produce no rescues anywhere on the
        grid (Section~\ref{app:scan_t7}) and produce no noise events at
        any cell that lies on the canonical Meltdown circuit.
  \item \emph{Residual-stream noise propagation mirrors the rescue
        propagation zone.} At $t{=}7$, $\texttt{x\_res}_k$ and
        $\texttt{block\_out}_k$ chamfer-noise for $k \geq 4$ on
        $\texttt{block\_out}$ and $k \geq 5$ on $\texttt{x\_res}$ (the
        offset is mechanical: $\texttt{block\_out}_k=\texttt{x\_res}_{k+1}$),
        and produce a clean sphere for $k \leq 3$. The rescue scan's
        propagation zone ($k\geq 5$, residual stream rescues; $k=4$,
        $\texttt{x\_res}_4$ and $\texttt{block\_out}_4$ rescue) and the
        rescue scan's pre-commit zone ($k\leq 3$, no within-block patch
        rescues except $\texttt{block\_out}_3=\texttt{x\_res}_4$) appear
        in the noising scan as the noise-propagation zone and the
        noise-inert zone respectively. Both directions therefore agree
        that block~$4$ at $t{=}7$ is where the corrupt signal first
        appears in the residual stream.
  \item \emph{Off-canonical MLP-path pockets mirror the rescue
        pockets.} Non-residual chamfer-noise events occur only on the
        post-CA MLP path (\texttt{norm2\_out}, \texttt{mlp\_in},
        \texttt{mlp\_out}) and only at the boundary blocks
        $k\in\{0, 31\}$: at $(k{=}0,\,t\in\{2,3,4,6\})$ and at
        $(k{=}31,\,t\in\{0,1,7\})$. The rescue scan's MLP-path pockets
        at $(k{=}0,\,t\in\{2,\dots,6\})$ and $(k{=}31,\,t\in\{1,2\})$
        cover almost the same set of cells. The boundary-block MLP path
        is therefore bidirectionally pluripotent --- it can both rescue
        and noise --- but, like its rescue counterpart, it is spatially,
        temporally, and modulewise disjoint from the canonical commit
        cell.
\end{itemize}

\paragraph{The asymmetry at the commit cell.}
At $(k{=}4,\,t{=}7)$, the rescue and noising verdicts disagree on which
within-block sites are causally active.
Table~\ref{tab:noising_canonical} lists every site at the cell
alongside its rescue-direction verdict from
Section~\ref{app:scan_t7}. The two directions agree on three sites: the
SA branch, AdaLN, and CA pre-write are inert in both. They disagree on
two: the CA-write triple is rescue-active but noise-inert, and
$\texttt{block\_out}_4$ is the only site that both rescues and noises
at this cell. (The block-input residual $\texttt{x\_res}_4$ rescues but
does not noise: it carries the clean signal forward into block~$4$ in
the rescue direction, but injecting the corrupt signal at the entry to
block~$4$ is too early --- the subsequent steps re-anchor the trajectory
toward sphere.)

\begin{table}[h!]
\centering
\small
\setlength{\tabcolsep}{6pt}
\renewcommand{\arraystretch}{1.05}
\begin{tabular}{@{}llcc@{}}
\toprule
\textbf{Site} & \textbf{Class} & \textbf{Rescue} & \textbf{Noise (chamfer)} \\
\midrule
\texttt{x\_res}            & block input    & \checkmark & --- \\
\texttt{block\_out}        & block output   & \checkmark & \checkmark \\
\texttt{ada}               & AdaLN params   & ---        & --- \\
SA-branch sites ($\times 7$) & SA branch    & ---        & --- \\
CA pre-write sites ($\times 4$) & CA pre-write & ---     & --- \\
\texttt{ca\_preproj}       & CA write       & \checkmark & --- \\
\texttt{Y\_kt}             & CA write       & \checkmark & --- \\
\texttt{ca\_module\_out}   & CA write       & \checkmark & --- \\
MLP-branch sites ($\times 2$) & MLP branch  & ---        & --- \\
\bottomrule
\end{tabular}
\caption{Every within-block site at the canonical commit cell
$(k{=}4,\,t{=}7)$, with its rescue-direction verdict
(Section~\ref{app:scan_t7}) and its noise-direction verdict (this
section). The CA-write triple rescues but does not noise; only the
block-output residual stream is causally active in both directions.
The block-output residual at $k{=}4$ coincides with the block-input
residual at $k{=}5$, so this entry is also the start of the
noise-propagation zone described in the third TL;DR bullet.}
\label{tab:noising_canonical}
\end{table}

\paragraph{Interpretation.}
A symmetric one-cell-isolates-the-commit picture would predict that
patching $\mathbf{Y}_{4,7}$ in either direction transports the run
between attractors. The rescue scan confirms the forward half of this
prediction; the noising scan refutes the backward half. The asymmetry
is consistent with the diffusion-dynamics view of
\S\ref{sec:diffusion-dynamics}: the trajectory must be in the basin of
the speckle attractor to commit to fragmentation, and the basin
boundary is crossed by a low-rank \emph{drift} in $\mathbf{Y}$ (the
direction-specific surgery of \S\ref{sec:spectral_causality}) rather
than by any single value of $\mathbf{Y}$ in isolation. Replacing
$\mathbf{Y}_{4,7}$ with the cached unhealthy value transplants the
endpoint of this drift but not the cumulative state of the rest of the
denoising trajectory; with the remaining seven steps conditioned on
the healthy $\mathbf{C}$ and starting from a healthy latent, the
trajectory re-anchors to the sphere basin. Conversely, the cumulative
state \emph{is} carried by the residual stream, which is why
$\texttt{block\_out}_4$ is the one within-block site at the commit
cell that is causally active in both directions and why all
$k\geq 5$ residual-stream cells at $t{=}7$ noise (and rescue) the run.
$\mathbf{Y}_{4,7}$ is thus the \emph{site} of the commit but not, in
isolation, a \emph{sufficient cause} of it; the surgery in
\S\ref{sec:spectral_causality} succeeds because it modifies the
trajectory's drift, not because it sets a single activation to a
single rescuing value.

\section{Statistical Properties of the Input Cloud}
\label{app:input_sweep}

\paragraph{Cloud ensemble.}
We construct an ensemble of perturbed sphere clouds by displacing the
Fibonacci reference $\mathcal{P}_0$ along random tangent fields built
from real spherical harmonics. For each wavenumber
$\ell \in \{2, 3, 4, 6, 8, 12, 16, 24, 32\}$ we draw five independent
Gaussian coefficient vectors over the $2\ell+1$ basis functions
$Y_\ell^m$, take the surface gradient of the resulting eigenfunction,
and apply the geodesic exponential map at amplitude $\varepsilon$ from
a $10$-point grid in $[0.001, 0.5]$ radians, normalized so that
$\varepsilon$ is the maximum geodesic displacement of any point.
Together with the slerp path of Section~\ref{sec:Meltdown} (14 values
of $\rho$, two random target clouds), and across three sample sizes
$N \in \{400, 500, 600\}$, this yields $1{,}437$ unique input clouds;
each is decoded under five independent diffusion seeds, for $7{,}185$
forward passes total.

\paragraph{Riesz $s{=}2$ energy.}
For a finite point set $\mathcal{P} = \{p_i\}_{i=1}^N \subset
\mathbb{S}^2$, the Riesz $s{=}2$ energy is
\begin{equation}
\label{eq:riesz}
E_2(\mathcal{P}) \;=\; \sum_{i \neq j}\, \|p_i - p_j\|^{-2},
\end{equation}
with $\|\cdot\|$ the Euclidean distance in $\mathbb{R}^3$. The Riesz
$s$-energy is a classical sphere-uniformity functional whose
minimizers approach the uniform measure on $\mathbb{S}^2$ as
$N \to \infty$ \citep{hardin2004discretizing, brauchart2015qmc}; the
Fibonacci sample $\mathcal{P}_0$ is a quasi-optimal minimizer at
finite $N$. We summarize departures from this reference by the
dimensionless \emph{Riesz excess}
\begin{equation}
\Delta E(\mathcal{P}) \;=\; E_2(\mathcal{P}) / E_2(\mathcal{P}_0) - 1,
\end{equation}
which is approximately zero on $\mathcal{P}_0$ and grows as
$\mathcal{P}$ becomes more crowded than uniform.

\begin{figure}[h]
    \centering
    \includegraphics[width=0.62\linewidth]{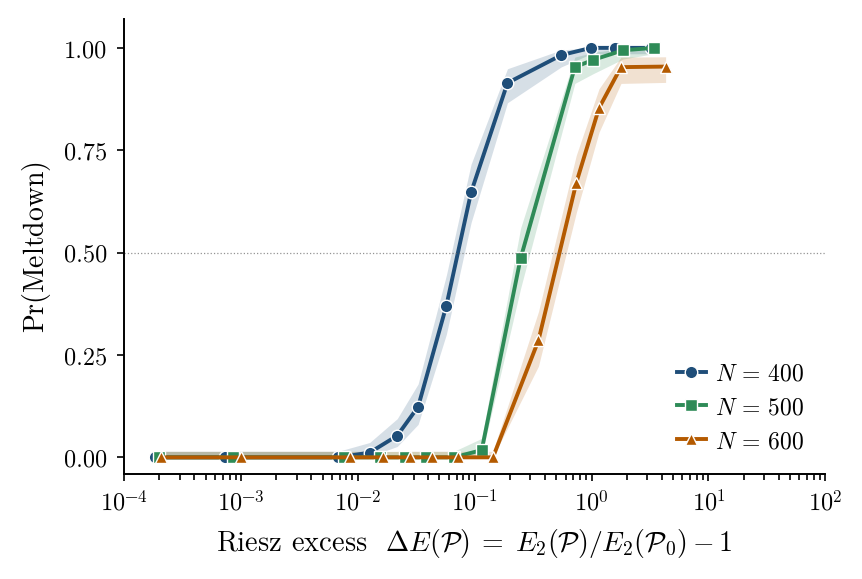}
    \caption{Probability of Meltdown as a function of the Riesz excess
    $\Delta E$ across the $7{,}185$-trial ensemble, broken down by
    sample size $N \in \{400, 500, 600\}$. Markers are per-bin
    binomial means; shaded bands are $95\%$ Jeffreys intervals over
    five diffusion seeds per cloud. The dotted line marks the $50\%$
    rate.}
    \label{fig:riesz_meltdown}
\end{figure}

\paragraph{Result.}
For each $N$ we bin the $2{,}395$ trials into $14$ quantile bins on
$\log_{10} \Delta E$ and report the binomial probability of Meltdown
with $95\%$ Jeffreys credible intervals over the diffusion-seed
dimension. Figure~\ref{fig:riesz_meltdown} shows that across all three
sample sizes, $\Delta E$ separates a Meltdown-free regime from a
saturation regime through a single-decade transition. For $N{=}400$,
no cloud melts in the $675$ trials with $\Delta E < 10^{-2}$, $99.6\%$
($762/765$) of the trials with $\Delta E > 0.20$ melt, and the $50\%$
probability crossing sits at $\Delta E^\star \approx 0.07$. The slerp
path of Section~\ref{sec:Meltdown} crosses $\Delta E^\star$ in the
same $\rho$-interval over which $C(\rho)$ jumps: at $\rho{=}0$ the
slerp cloud coincides with $\mathcal{P}_0$ ($\Delta E = 0$, $C{=}1$),
and at the smallest non-trivial grid sample $\rho{=}0.05$ we already
have $\Delta E \approx 0.18 \gg \Delta E^\star$ and $C \approx 73$.

\section{Encoder Propagation of Input Non-uniformity}
\label{app:encoder_propagation}

Appendix~\ref{app:input_sweep} established that Meltdown is governed by a
sharp threshold on the Riesz $s{=}2$ energy of the input cloud. We now ask
how this scalar reaches the diffusion backbone: which encoder stage carries
it, and whether the encoder adds, removes, or distorts the melt-relevant
signal en route to the conditioning $\mathbf{C}$.

\paragraph{Protocol.}
We reuse the cloud ensemble of Appendix~\ref{app:input_sweep}, restricted
to a single diffusion seed per cloud so that melt outcome is a per-cloud
binary. This yields $1{,}434$ perturbed clouds across the three sample
sizes ($479$ at each $N$), of which $484$ melt and $950$ do not.
For each cloud we run the WaLa PointNet encoder once and capture
post-LayerNorm activations at the seven sequential stages
(\texttt{ln1}--\texttt{ln4}, the MAB pool, \texttt{ln\_f1},
\texttt{ln\_f2}) plus the encoder output $\mathbf{C}$. Letting $X^L$
denote stage $L$ and $X^L_0$ the same stage on the Fibonacci reference
$\mathcal{P}_0$, we report the relative Frobenius drift
$\delta^L \;=\; \|X^L - X^L_0\|_F / \|X^L_0\|_F$,
averaged separately over melted and healthy clouds at each $N$.

\paragraph{Layer-wise drift profile.}
Table~\ref{tab:encoder_drift} reports $\delta^L$ for melted and healthy
clouds and their ratio. Three observations hold uniformly across $N$.
First, the four per-point linear stages amplify the melted-vs-healthy
drift ratio by $5.6\text{--}6.8\times$. Second, the MAB pool compresses
this to $2.1\text{--}2.4\times$, a factor that is then preserved through
\texttt{ln\_f1}, \texttt{ln\_f2}, and into $\mathbf{C}$. Third, while
the absolute relative drifts shift with $N$ (lower $N$ has smaller
healthy drift, so ratios are larger), the qualitative profile
--- per-point amplification, MAB compression, downstream preservation
--- is identical. The MAB is therefore the encoder stage at which
per-point displacement is converted into a token-level conditioning
signal, and the gap between $\mathbf{Y}_{\mathrm{melt}}$ and
$\mathbf{Y}_{\mathrm{heal}}$ that the backbone subsequently reads is
already established at the encoder's output.

\begin{table}[h]
\centering
\small
\setlength{\tabcolsep}{4.5pt}
\renewcommand{\arraystretch}{1.05}
\begin{tabular}{l ccc ccc ccc}
\toprule
& \multicolumn{3}{c}{$N{=}400$} & \multicolumn{3}{c}{$N{=}500$} & \multicolumn{3}{c}{$N{=}600$} \\
\cmidrule(lr){2-4}\cmidrule(lr){5-7}\cmidrule(lr){8-10}
stage & heal & melt & ratio & heal & melt & ratio & heal & melt & ratio \\
\midrule
\texttt{ln1}      & 0.020 & 0.135 & $6.8\times$ & 0.027 & 0.163 & $6.2\times$ & 0.030 & 0.178 & $5.9\times$ \\
\texttt{ln2}      & 0.024 & 0.162 & $6.7\times$ & 0.032 & 0.195 & $6.1\times$ & 0.037 & 0.213 & $5.8\times$ \\
\texttt{ln3}      & 0.027 & 0.182 & $6.6\times$ & 0.037 & 0.219 & $6.0\times$ & 0.042 & 0.238 & $5.7\times$ \\
\texttt{ln4}      & 0.026 & 0.171 & $6.6\times$ & 0.035 & 0.206 & $5.9\times$ & 0.040 & 0.223 & $5.6\times$ \\
\midrule
MAB pool          & 0.091 & 0.219 & $2.4\times$ & 0.096 & 0.208 & $2.2\times$ & 0.091 & 0.193 & $2.1\times$ \\
\texttt{ln\_f1}   & 0.065 & 0.156 & $2.4\times$ & 0.068 & 0.148 & $2.2\times$ & 0.065 & 0.138 & $2.1\times$ \\
\texttt{ln\_f2}   & 0.013 & 0.030 & $2.4\times$ & 0.013 & 0.029 & $2.2\times$ & 0.013 & 0.027 & $2.1\times$ \\
$\mathbf{C}$      & 0.131 & 0.310 & $2.4\times$ & 0.138 & 0.294 & $2.1\times$ & 0.131 & 0.274 & $2.1\times$ \\
\bottomrule
\end{tabular}
\caption{Relative Frobenius drift $\delta^L$ at each encoder stage,
averaged over melted and healthy clouds at each sample size. The
per-point stages (\texttt{ln1}--\texttt{ln4}) carry a $\sim$$6\times$
melted-vs-healthy gap; the MAB pool compresses this to $\sim$$2\times$
which is then preserved to the encoder output $\mathbf{C}$.}
\label{tab:encoder_drift}
\end{table}

\paragraph{Faithful transduction.}
A simple consistency check is whether melt-prediction accuracy is lost
or gained as the signal traverses the encoder. We score melt outcome
with each scalar feature individually, computing the area under the
ROC curve at each $N$. The Riesz excess $\Delta E$ achieves
AUC $\in [0.981, 0.995]$ across the three $N$; the relative drift of
$\mathbf{C}$ achieves AUC $\in [0.974, 0.982]$; no intermediate stage
is more or less melt-predictive than its neighbours by more than
$0.01$. The encoder neither manufactures nor discards melt-relevant
information: it transduces the input statistic from a property of the
cloud into a property of the conditioning $\mathbf{C}$ that is read
by the diffusion backbone.

\paragraph{Structural, not scalar.}
The encoder is melt-faithful but not a linear $\Delta E$ transducer.
Reducing $\mathbf{C}$ to its top-$64$ PCA components (explaining
$\geq 99\%$ of the variance across the ensemble) and fitting a $5$-fold
cross-validated ridge regression $\mathrm{PCA}(\mathbf{C}) \to \Delta E$
yields $R^2 \in [-0.18, 0.09]$ across the three $N$: the input scalar
is not recoverable as a linear direction in $\mathbf{C}$. Yet a
logistic probe on the same features predicts melt at AUC
$\in [0.974, 0.981]$, and after residualizing each PCA dimension
against $\Delta E$ before the probe the melt AUC remains
$\in [0.864, 0.937]$. Melt-relevant content is therefore written into
$\mathbf{C}$ structurally rather than along any single scalar axis.
This is consistent with the activation-patching result of
\S\ref{sec:activation_patching}: the cross-attention write at
$(k{=}4, t{=}7)$ commits the trajectory by reading directional structure
in the encoded condition.

\section{PowerRemap Site Sweep}
\label{app:powerremap_sweep}

The patching scan of Appendix~\ref{app:act_patching_extended} localizes
the Meltdown commit to the cross-attention output projection
$\mathbf{Y}_{4,7}$. \texttt{PowerRemap} (\S\ref{sec:power_remap_main})
is derived from this localization, but its operation differs from
patching: it does not import a healthy activation, but reshapes the
singular spectrum of whatever activation is present at the targeted
site. We therefore verify directly that \texttt{PowerRemap} inherits
the patching localization, by sweeping the intervention across every
cross-attention and MLP cell of \textsc{WaLa}'s U-ViT and asking which
sites, if any, rescue Meltdown. Self-attention is excluded: a prior
sweep over all $256$ self-attention cells produced $0$ rescues, and the
patching scan likewise finds $0$ self-attention rescues across the
grid (Appendix~\ref{app:act_patching_extended}).

\paragraph{Protocol.} For each $(\text{component}, k, t)$ cell with
component $\in\{\texttt{cross\_attn},\,\texttt{mlp}\}$, $k\in\{0,\dots,31\}$,
and $t\in\{7,\dots,0\}$, we register a forward hook on the corresponding
sub-module that applies \texttt{PowerRemap} ($\gamma=100$) to the
sub-module's output at step $t$ and lets the rest of the reverse process
run unmodified. This gives $2\cdot 32\cdot 8 = 512$ tested sites
($768$ counting the excluded self-attention grid). Conditions match the
patching scan: $\rho_{\mathrm{melt}}=0.4$, $N_{\mathrm{points}}=400$,
DDIM with $T=8$, baseline $C=140$. A site is recorded as a
\emph{rescue} when the output mesh has a single connected component
($C=1$). Because $C=1$ alone is necessary but not sufficient, we additionally validate
each rescue by sphere-fit residual ($\sigma_{\|\cdot\|}/\bar{R}$, where
$R$ is the mean vertex-to-centroid distance) and the isoperimetric
sphericity proxy $\pi^{1/3}(6V)^{2/3}/A$. A rescue is declared
\emph{valid} when sphere-fit residual $<0.25$ and sphericity $>0.4$
--- thresholds chosen generously; clean baseline outputs have residual
$\approx 0.001$ and sphericity $\approx 1.000$.

\paragraph{Results.} Of the $512$ sites tested, four produce a valid
rescue. Table~\ref{tab:powerremap_sweep} lists them. No other site, in
either component, at any block, at any step, recovers connectivity:
the remaining $508$ sites leave $C$ statistically indistinguishable
from the baseline $C=140$. Counting against the full $768$-cell grid
including the excluded self-attention sweep, valid-rescue specificity
is $4/768 = 0.5\%$.

\begin{table}[h!]
\centering
\small
\setlength{\tabcolsep}{6pt}
\renewcommand{\arraystretch}{1.05}
\begin{tabular}{@{}llccccc@{}}
\toprule
& \textbf{site} & \textbf{block} & \textbf{step}
& \textbf{sphericity} & \textbf{sphere-fit residual} & \textbf{role} \\
\midrule
A & \texttt{cross\_attn} & $4$ & $7$ & $0.9998$ & $0.0014$ & canonical commit ($\mathbf{Y}_{4,7}$) \\
B & \texttt{cross\_attn} & $3$ & $7$ & $0.9998$ & $0.0014$ & upstream feeder at $t{=}7$ \\
C & \texttt{mlp}         & $0$ & $7$ & $0.9998$ & $0.0014$ & upstream feeder at $t{=}7$ \\
D & \texttt{mlp}         & $0$ & $0$ & $0.9296$ & $0.0880$ & off-canonical MLP-path pocket \\
\bottomrule
\end{tabular}
\caption{The four sites at which \texttt{PowerRemap} ($\gamma=100$)
rescues Meltdown out of $512$ swept (or $0.5\%$ of the full $768$-cell
grid including excluded self-attention). Sites A--C are at the
canonical denoising step $t{=}7$, lie on the residual-stream pathway
upstream of $\mathbf{Y}_{4,7}$, and produce sphere reconstructions
indistinguishable in quality from the canonical rescue (sphericity
agreement to four decimals). Site D is at the boundary block $k{=}0$
at $t{=}0$ and produces a geometrically inferior reconstruction
(sphericity $0.93$, residual $\sim 60\times$ larger), matching the
off-canonical MLP-path pocket pattern documented in
Appendix~\ref{app:act_patching_extended}.}
\label{tab:powerremap_sweep}
\end{table}

\paragraph{Interpretation.} The three $t{=}7$ rescues (A--C) form a
contiguous residual-stream pathway: $\mathrm{MLP}_{0,7}$ writes into
the residual stream at block $0$, that signal flows forward through
blocks $1$--$3$ where the cross-attention output projection
$\mathbf{Y}_{3,7}$ contributes additional structure, and the canonical
$\mathbf{Y}_{4,7}$ is the final commit. Compressing the spectrum at
any of these three points heads off the corruption before it is
written into the residual stream that feeds block $4$'s commit.
Patches \emph{at} a site downstream of $\mathbf{Y}_{4,7}$ at $t{=}7$
do not rescue, consistent with the patching scan: once the residual
stream carries the committed signal, single-submodule interventions
at later blocks cannot undo it.

The off-canonical site D ($\mathrm{MLP}_{0,0}$) reproduces a pattern
already isolated in the patching scan: spurious connectivity rescues
on the post-CA MLP path at the boundary blocks $k\in\{0,31\}$ at
non-canonical denoising steps (Appendix~\ref{app:act_patching_extended},
\S\ref{app:mlp_pockets}). These pockets are spatially, temporally, and
modulewise disjoint from the canonical Meltdown circuit, and produce
geometrically inferior reconstructions: site D's sphericity ($0.93$)
and sphere-fit residual ($0.088$) are inferior to the canonical
rescue's by orders of magnitude on the residual axis, and match the
quality of the off-canonical patching rescues reported in
Table~\ref{tab:rescue_quality} (categories F and G). The two scans,
run on disjoint experimental designs, recover the same circuit and
the same boundary-block side-channel.

Across both methods of intervention --- single-cell patching with a
healthy activation, and single-cell spectral compression of the
existing activation --- the locus of causal control over Meltdown is
the same: the cross-attention pathway feeding $\mathbf{Y}_{4,7}$ at
the first denoising step. \texttt{PowerRemap} inherits this
localization.

\section{Simpler Baselines}
\label{app:simpler_baselines}

This appendix reports the protocol and full results behind the
``simpler interventions'' paragraph in
\S\ref{sec:eval_at_scale}. We test three deployment-fair alternatives
to \texttt{PowerRemap}, one per stage of the causal chain we
identified in \S\ref{sec:MI}: input-cloud uniformization,
diffusion-trajectory noise injection, and initial-noise resampling.
Each baseline is the canonical SOTA representative of its category and
operates without surface knowledge or quality oracles, matching the
deployment regime considered in \S\ref{sec:introduction}. None
rescues at meaningful rates.

\paragraph{TL;DR.}
On the canonical sphere setup (\S\ref{app:experiments_sphere_wala_ddim}) at
$\rho_{\mathrm{crit}}{=}0.4$ with $6$ diffusion seeds:
\begin{itemize}\setlength\itemsep{1.5pt}
  \item \textbf{Stage 1: input uniformization (WLOP).} $0/48$ sphere
        rescues across iteration counts $\{1,2,5,15\}$. At small counts
        (it.\ $1,2$) the operator partially uniformizes the cloud
        (Voronoi-area COV: $0.517 \to 0.422$, an $18\%$ reduction)
        without crossing the basin separatrix; at larger counts the
        cloud destabilizes (collapse at it.\ $15$, minimum pairwise arc
        $\to 10^{-4}$).
  \item \textbf{Stage 1 oracle ablation (WLOP with $\mathbb{S}^2$
        projection).} $0/18$ sphere rescues across iteration counts
        $\{2,5,15\}$. Surface knowledge does not change the verdict for
        local-repulsion uniformizers: the on-manifold variant exhibits
        the same partial-uniformization plateau and the same
        long-iteration destabilization.
  \item \textbf{Stage 2: noise injection at every denoising step.}
        $0/96$ sphere rescues across $8$ steps $\times$ $2$ strengths
        ($s\!\in\!\{0.3,0.6\}$) $\times$ $6$ seeds. Aggressive
        injection at the earliest steps meaningfully disturbs the
        trajectory ($\bar{C}{:}\,145.3 \to 71.2$ at $t{=}T,\,s{=}0.6$)
        but never produces a sphere; injection at the bifurcation
        step $\tau^\star{\approx}\,5$ identified in
        \S\ref{sec:diffusion-dynamics} leaves the trajectory
        essentially unmoved.
  \item \textbf{Stage 3: best-of-K initial noise.} $0/24$ sphere
        rescues across $K{=}1{,}\dots{,}4$. Minimum $C$ over $24$
        independent $\mathbf{x}_T$ draws is $134$.
  \item \texttt{PowerRemap} ($\gamma{=}100$, anchor on the same seeds):
        $6/6$ sphere rescues, mean chamfer $0.0887$, mean sphericity
        proxy $1.0001$.
\end{itemize}

\subsection{Deployment regime and fair-baseline criteria}
\label{app:baseline_criteria}

The paper considers reconstruction from sparse point clouds of
geometry that the surface-recovery model is itself responsible for
inferring (\S\ref{sec:introduction}, \S\ref{sec:Meltdown}). A baseline
is \emph{deployment-fair} if it can be run by a practitioner who has
only the input cloud $\mathcal{P}$ and the diffusion model
$G{=}D \circ B \circ E$ at hand. Concretely, a fair baseline must
satisfy three conditions: (i) it uses only $\mathcal{P}$ and $G$ ---
no ground-truth surface, no clean reference cloud, no surface-quality
metric; (ii) it preserves the input size $N$, since the user has the
points the sensor produced; (iii) selection criteria, where applicable,
depend only on output-side observables computable from the mesh
(e.g.\ topological connectivity), not on properties of an underlying
surface. These conditions match the criterion that
\texttt{PowerRemap}'s $\gamma$-grid for \textsc{Make-a-Shape} satisfies
in \S\ref{sec:eval_at_scale}: connectivity $C{=}1$ as the selection
target requires no ground-truth surface and is therefore deployable at
test time. They also exclude several classical alternatives.
Tangent-plane-based redistribution and Voronoi-cell methods on a
parametric surface presuppose the surface; learned point-cloud
denoisers (e.g.\ \cite{rakotosaona2020pointcleannet}) introduce a
separately trained model, with its own training distribution, on top
of the diffusion pipeline.

\subsection{Choice of representative baselines}
\label{app:baseline_choices}

We choose one representative per stage of the causal chain identified
in \S\ref{sec:MI}, taking the canonical SOTA reference at each.

\paragraph{Stage 1 --- input cloud.}
The corruption $\mathcal{P}_\rho$ in \S\ref{sec:Meltdown} is on-surface
non-uniformity rather than off-surface noise: each $p_i(\rho)$ lies
exactly on $\mathbb{S}^2$ by construction of the per-point SLERP. The
appropriate point-cloud operator is therefore a \emph{redistribution}
operator at fixed $N$, not a denoiser. Bilateral filters and their
variants are
designed for off-surface noise and project points along estimated
normals toward local tangent planes; they would be approximately a
no-op on our perturbation, since every input is already on its local
tangent plane to numerical precision.

We therefore use Locally Optimal Projection (LOP)
\citep{lipman2007parameterization}, a parameterization-free,
surface-agnostic redistribution operator that iteratively moves each
particle toward an L1-median data target while a repulsion term spreads
particles apart. Specifically we run its density-weighted variant
WLOP \citep{huang2009consolidation}, which is the canonical
extension to non-uniform inputs --- our setting. WLOP is parameter-light
(repulsion weight $\mu$ and support radius $h$) and requires no normal
estimation, making it the strictest deployment-fair representative of
the surface-agnostic uniformization literature.

\paragraph{Stage 2 --- diffusion sampling.}
Stage 2 baselines must operate on the reverse-time trajectory using
only the trained denoiser. The canonical SOTA representative of this
category is variance-preserving noise injection at intermediate
denoising steps, formalized as ``Langevin churn'' in \citet{karras2022elucidating}.
This operator is also the natural test of the bifurcation hypothesis
in \S\ref{sec:diffusion-dynamics}: if the reverse trajectory is poised
on a basin separatrix at $\tau^\star$, noise added near $\tau^\star$
should sometimes flip basin allocation. We sweep injection at every
denoising step at two strengths to make the test exhaustive.

\paragraph{Stage 3 --- alternative outside the activation pathway.}
Best-of-K initial-noise selection \citep{ma2025inferencescaling} is the
simplest non-activation-based intervention: vary the seed
$\mathbf{x}_T$ and let the model reconcile. The connectivity criterion
$C{=}1$ used for selection is computable from the output mesh alone
without surface knowledge, so this baseline is deployment-fair. It is
the SOTA representative of inference-time scaling that does not modify
the model.

\subsection{Shared protocol}
\label{app:baseline_protocol}

All experiments use the WaLa sphere setup of
Appendix~\ref{app:experiments_sphere_wala_ddim}: $N{=}400$ Fibonacci
points, target cloud generated by Gaussian jitter ($\sigma{=}0.1$) and
re-projection to $\mathbb{S}^2$, per-point SLERP with control parameter
$\rho$, and $\rho_{\mathrm{crit}}{=}0.4$ throughout. We use $6$
diffusion seeds and report rescue under three connectivity-based
verdicts of increasing stringency: \emph{lax}
($C{=}1$ and $\geq{200}$ faces), \emph{chamfer-rescue}
(lax plus chamfer within $0.020$ of the clean baseline), and
\emph{strict} (chamfer-rescue plus mean radius, radial std, and
sphericity proxy within tolerances of clean). All headline rates below
report chamfer-rescue. The clean baseline ($6$ seeds) has $C{=}1$,
mean chamfer $0.0917$, mean sphericity proxy $1.0006$, and
$\sim{513}{,}000$ faces; the corrupt baseline ($6$ seeds) has
mean $C{=}145.3$, mean chamfer $0.1122$, mean sphericity proxy
$0.2575$, and $\sim{129}{,}500$ faces. \texttt{PowerRemap} ($\gamma{=}100$) on the same
seeds rescues $6/6$ with mean chamfer $0.0887$ and mean sphericity
proxy $1.0001$. We use these reference values throughout.

We additionally report two input-cloud uniformity statistics for
Stage 1: the Riesz $s{=}2$ energy
$E_2(\mathcal{P})\,{=}\,\sum_{i\neq j}\|p_i-p_j\|^{-2}$
(Appendix~\ref{app:input_sweep}), with reference values $E_2{=}113{,}461$
on clean and $E_2{=}237{,}309$ on corrupt; and the spherical-Voronoi
area coefficient of variation $\mathrm{COV}_V$, with reference values
$\mathrm{COV}_V{=}0.010$ on clean and $\mathrm{COV}_V{=}0.517$ on
corrupt. These statistics are linked to Meltdown by the input-cloud
analysis of Appendix~\ref{app:input_sweep}: $E_2$ is the scalar that
determines whether $\mathcal{P}$ falls into the Meltdown regime.

\subsection{Stage 1: input-cloud uniformization (LOP / WLOP)}
\label{app:baseline_stage1}

The WLOP iteration moves each particle $q_i$ to
$q_i^{\mathrm{new}}\,{=}\,d_i + \mu\,r_i$, where $d_i$ is a
density-weighted L1-median data target computed from $\mathcal{P}$
and $r_i$ is a repulsion gradient with respect to all other particles
$\{q_j\}_{j\neq i}$. We use the parameter values recommended in
\citet{huang2009consolidation}: repulsion weight $\mu{=}0.45$, support
radius $h$ set automatically to four times the median nearest-neighbor
distance, and a step-size cap of $0.3 h$ for numerical stability.
Particles are initialized to the corrupt cloud $\mathcal{P}_\rho$ for
in-place consolidation at fixed $N{=}400$. We run iteration counts
$\{1,2,5,15\}$ and apply the resulting consolidated cloud as the
diffusion input.

\subsubsection{LOP without surface knowledge (deployment-fair)}
\label{app:baseline_stage1_noproj}

Table~\ref{tab:baseline_lop_noproj} reports rescue rates and
input-statistics changes after WLOP. At small iteration counts
(it.\ $1$, $2$), the operator measurably uniformizes the cloud:
$\mathrm{COV}_V$ drops from $0.517$ on the corrupt input to $0.436$
($16\%$ reduction) at it.\ $1$ and to $0.422$ ($18\%$ reduction) at
it.\ $2$, and the Riesz energy drops from $237{,}309$ to
$\sim{158}{,}000$. Neither produces a sphere rescue. At
it.\ $5$, partial uniformization continues
($\mathrm{COV}_V{=}0.406$) but the cloud begins to drift off
$\mathbb{S}^2$ (Riesz energy increases to $179{,}673$ and the minimum
pairwise arc proxy collapses to $0.008$). At it.\ $15$, the
operator destabilizes catastrophically: the minimum pairwise arc proxy
collapses to $10^{-4}$, indicating particle coincidence, and the
Riesz energy rises by three orders of magnitude. This is the
documented off-manifold drift pathology of fixed-$N$
WLOP applied without explicit surface projection
\citep{lee2021electric, stotko2022incompletegamma}: the L1-median data
target of an on-manifold cloud lies inside the manifold, so particles
drift inward iteration by iteration, and the inverse-square repulsion
gradient diverges as inter-particle distances shrink.

\begin{table}[h!]
\centering
\small
\setlength{\tabcolsep}{5pt}
\renewcommand{\arraystretch}{1.05}
\begin{tabular}{@{}lcccccc@{}}
\toprule
\textbf{Variant} & \textbf{Iters} & \textbf{rescue}
& $\bar{C}$ & $\overline{\mathrm{COV}_V}$
& $\overline{E_2}$ & $\overline{\mathrm{arc}_{\min}}$ \\
\midrule
WLOP            & $1$  & $0/6$ & $163.0$ & $0.436$ & $1.58{\times}10^{5}$ & $0.017$ \\
WLOP            & $2$  & $0/6$ & $147.8$ & $0.422$ & $1.58{\times}10^{5}$ & $0.020$ \\
WLOP            & $5$  & $0/6$ & $164.7$ & $0.406$ & $1.80{\times}10^{5}$ & $0.008$ \\
WLOP            & $15$ & $0/6$ & $172.5$ & $0.457$ & $1.46{\times}10^{8}$ & $1{\times}10^{-4}$ \\
\midrule
\multicolumn{2}{l}{corrupt baseline}     & $0/6$ & $145.3$ & $0.517$ & $2.37{\times}10^{5}$ & --- \\
\multicolumn{2}{l}{clean baseline}       & $6/6$ & $1.0$   & $0.010$ & $1.13{\times}10^{5}$ & --- \\
\multicolumn{2}{l}{\texttt{PowerRemap}}  & $6/6$ & $1.0$   & $0.517$\textsuperscript{$\dagger$} & $2.37{\times}10^{5}$\textsuperscript{$\dagger$} & --- \\
\bottomrule
\end{tabular}
\caption{Stage 1, deployment-fair: WLOP \citep{huang2009consolidation}
without surface projection. Iteration counts $\{1,2\}$ partially
uniformize the cloud without rescuing; counts $\{5,15\}$ destabilize.
$^\dagger$\,\texttt{PowerRemap} acts on the activation $\mathbf{Y}_{4,7}$
and leaves the input cloud unchanged, so its input statistics equal
those of the corrupt cloud.}
\label{tab:baseline_lop_noproj}
\end{table}

\subsubsection{LOP with $\mathbb{S}^2$ projection (oracle ablation)}
\label{app:baseline_stage1_s2proj}

To separate the failure into ``surface-agnostic does not work'' versus
``local repulsion at fixed $N$ does not work,'' we run an oracle
ablation: identical WLOP iteration with a final radial re-projection
to $\mathbb{S}^2$ after each step, $q_i \leftarrow q_i / \lVert q_i \rVert$.
This injects the surface as a hard constraint and is therefore
\emph{not} deployment-fair; we use it only to calibrate the failure.
Table~\ref{tab:baseline_lop_s2proj} reports the result.

\begin{table}[h!]
\centering
\small
\setlength{\tabcolsep}{5pt}
\renewcommand{\arraystretch}{1.05}
\begin{tabular}{@{}lcccccc@{}}
\toprule
\textbf{Variant} & \textbf{Iters} & \textbf{rescue}
& $\bar{C}$ & $\overline{\mathrm{COV}_V}$
& $\overline{E_2}$ & $\overline{\mathrm{cham}}$ \\
\midrule
WLOP $+\,\Pi_{\mathbb{S}^2}$ & $2$  & $0/6$ & $144.5$ & $0.422$ & $1.58{\times}10^{5}$ & $0.107$ \\
WLOP $+\,\Pi_{\mathbb{S}^2}$ & $5$  & $0/6$ & $166.5$ & $0.406$ & $1.83{\times}10^{5}$ & $0.107$ \\
WLOP $+\,\Pi_{\mathbb{S}^2}$ & $15$ & $0/6$ & $169.7$ & $0.455$ & $1.50{\times}10^{8}$ & $0.110$ \\
\bottomrule
\end{tabular}
\caption{Stage 1, oracle ablation: WLOP with $\mathbb{S}^2$ projection
after each iteration. Despite explicit surface knowledge, no
iteration count produces a rescue. Surface projection is not
sufficient to push WLOP past the basin transition at fixed $N$,
because radial re-projection does not prevent particle coincidence:
two particles whose updates point in similar directions land at
the same location on $\mathbb{S}^2$, reproducing the
long-iteration collapse.}
\label{tab:baseline_lop_s2proj}
\end{table}

The oracle variant exhibits the same qualitative pattern as the
deployment-fair variant: at it.\ $2$ both achieve
$\mathrm{COV}_V{=}0.422$; at it.\ $15$ both collapse with
$E_2$ exceeding $10^{8}$. Surface projection therefore does not change
the verdict at any iteration count: WLOP-style local-repulsion
uniformization at fixed $N$, with or without $\mathbb{S}^2$
projection, fails to cross the basin separatrix.

\subsubsection{Joint interpretation of Stage 1}
\label{app:baseline_stage1_joint}

The deployment-fair and oracle variants together support a stronger
conclusion than either alone. The deployment-fair variant fails for two
distinct reasons depending on iteration count: at low counts the cloud
remains insufficiently uniformized to cross the basin transition
($\mathrm{COV}_V \in [0.42,\,0.44]$ versus the clean reference
$\mathrm{COV}_V{=}0.010$); at high counts the cloud destabilizes off
the manifold. A natural concern is that the high-iteration failure is
specifically the off-manifold drift, and that surface knowledge would
fix it. The oracle ablation rules this out: even with explicit
$\mathbb{S}^2$ projection, the WLOP iteration does not reach a uniformity
level that crosses the rescue threshold, and at high iteration counts
particles still coincide. The conclusion is that local-repulsion
uniformizers, regardless of surface knowledge, do not produce
sufficiently uniform fixed-$N$ point sets on $\mathbb{S}^2$ to flip the
diffusion trajectory back to the sphere basin. Together, Stage 1
delivers $0/48$ rescues across iteration counts and surface
conditions.

\subsection{Stage 2: noise injection at every denoising step}
\label{app:baseline_stage2}

We test EDM-style variance-preserving noise injection at every
denoising step. At step $t_{\mathrm{inj}}$, we replace the latent
$\mathbf{x}_t$ with
$\mathbf{x}_t' = \sqrt{1-s^2}\,\mathbf{x}_t + s\,\boldsymbol\epsilon$,
$\boldsymbol\epsilon \sim \mathcal{N}(0, I)$, on the conditional slot
of the CFG batch (the unconditional slot is left untouched, consistent
with the rest of the analysis being restricted to the conditional
stream; \S\ref{app:experiments_general}). The remainder of reverse
diffusion runs unmodified. We sweep $t_{\mathrm{inj}} \in \{0,1,\dots,7\}$
across the entire DDIM schedule and two strengths
$s \in \{0.3,\,0.6\}$, giving $8 \times 2 \times 6 = 96$ trials.
Strength $s{=}0.6$ is aggressive: the latent retains $\sqrt{1-0.6^2}{=}0.8$ of its
norm and is mixed with a fresh Gaussian of magnitude $0.6$ relative to
unit variance. Sweeping the entire schedule preempts the question of
whether a different choice of $t_{\mathrm{inj}}$ would have rescued.

Table~\ref{tab:baseline_noise_inject} reports the per-cell mean
connected-component count $\bar{C}$. No setting produces a sphere:
$0/96$ rescues, with $C$ ranging from $55$ to $153$. The
finest-grained pattern is monotone with $t_{\mathrm{inj}}$ at fixed
$s$: injection at the earliest steps disturbs the trajectory most
($\bar{C}{=}71.2$ at $(t_{\mathrm{inj}}{=}0,\,s{=}0.6)$, down from
$143.0$ at $(t_{\mathrm{inj}}{=}7,\,s{=}0.6)$), reflecting that an
early perturbation has more reverse-diffusion steps over which to
propagate. None of these disturbances flips basin allocation: even
where $\bar{C}$ drops to $71$, the resulting mesh is fragmented rather
than spherical (mean chamfer at $(t_{\mathrm{inj}}{=}0,\,s{=}0.6)$ is
$0.112$, identical to the corrupt baseline).

\begin{table}[h!]
\centering
\small
\setlength{\tabcolsep}{6pt}
\renewcommand{\arraystretch}{1.05}
\begin{tabular}{@{}cccccccccc@{}}
\toprule
& \multicolumn{8}{c}{\textbf{injection step $t_{\mathrm{inj}}$}} & \\
\cmidrule(lr){2-9}
\textbf{strength $s$} & $0$ & $1$ & $2$ & $3$ & $4$ & $5$ ($\tau^\star$) & $6$ & $7$ & \textbf{rescue} \\
\midrule
$0.3$ & $115.8$ & $130.8$ & $140.3$ & $141.7$ & $145.2$ & $145.7$ & $146.3$ & $147.0$ & $0/48$ \\
$0.6$ & $\phantom{0}71.2$  & $102.3$ & $129.7$ & $137.8$ & $139.3$ & $143.2$ & $142.0$ & $143.0$ & $0/48$ \\
\midrule
\multicolumn{9}{l}{corrupt baseline (no intervention): $\bar{C}{=}145.3$} & --- \\
\bottomrule
\end{tabular}
\caption{Stage 2: per-cell mean connected-component count $\bar{C}$
for variance-preserving noise injection at every denoising step,
$6$ seeds per cell. Step $5$ corresponds to the bifurcation step
$\tau^\star{\approx}\,5$ identified by the dip-test analysis in
\S\ref{sec:diffusion-dynamics}. No cell rescues at either strength.
The strict-rescue ($C{=}1$ and chamfer below clean
$+0.020$) verdict is $0/96$.}
\label{tab:baseline_noise_inject}
\end{table}

The result at the bifurcation step deserves explicit comment.
\S\ref{sec:diffusion-dynamics} identifies $\tau^\star{\approx}\,5$ as the
step at which the reverse-time potential bifurcates and the trajectory
commits to a basin (Figs.~\ref{fig:trajectories}, \ref{fig:potential};
Hartigan dip test rejects unimodality from $t{=}5$ onward). A
naive reading would predict that noise added at $\tau^\star$ should
be the most effective intervention, since trajectories there are
poised on the basin separatrix. The data refute that prediction
sharply: at $t_{\mathrm{inj}}{=}5$, even the aggressive $s{=}0.6$
injection moves $\bar{C}$ only from $145.3$ (corrupt baseline) to
$143.2$, and produces $0/6$ rescues. The bifurcation-theoretic
interpretation is consistent: the basin separatrix is sharp, and a
trajectory that has accumulated bias from earlier denoising steps
(the directional drift in $\mathbf{Y}_{4,7}$ documented in
\S\ref{app:wala_drift_surgery}) cannot be transported back across
the separatrix by symmetric Gaussian noise. The Stage 2 result
therefore both fails as a Meltdown rescue and supports the
diffusion-dynamics account.

\subsection{Stage 3: best-of-K initial-noise selection}
\label{app:baseline_stage3}

For each of the $6$ seeds, we run $K{=}4$ independent reverse-diffusion
trajectories from different initial noises $\mathbf{x}_T$ and the
identical input cloud $\mathcal{P}_\rho$, recording the per-attempt
mesh and its component count.
Table~\ref{tab:baseline_bestofk} reports the resulting best-of-$K$
chamfer-rescue rate as a function of $K$. The minimum $C$ across all
$24$ independent draws is $134$. At
$\rho_{\mathrm{crit}}{=}0.4$, the speckle attractor's basin of
attraction is dominant enough that all $24$ initial noises tested land
in it, and best-of-K selection at $K{\leq}4$ does not rescue.

\begin{table}[h!]
\centering
\small
\setlength{\tabcolsep}{8pt}
\renewcommand{\arraystretch}{1.05}
\begin{tabular}{@{}lcccc@{}}
\toprule
\textbf{Best-of-K rate} & $K{=}1$ & $K{=}2$ & $K{=}3$ & $K{=}4$ \\
\midrule
sphere rescue (chamfer)  & $0/6$ & $0/6$ & $0/6$ & $0/6$ \\
\bottomrule
\end{tabular}
\caption{Stage 3: best-of-$K$ initial-noise selection
\citep{ma2025inferencescaling}. $C{=}1$ is the selection criterion
and is computable from the output mesh alone. Across $K \in \{1,2,3,4\}$
on $6$ seeds, the rescue rate is $0$. Total $24$ independent
$\mathbf{x}_T$ draws, minimum $C{=}134$.}
\label{tab:baseline_bestofk}
\end{table}

\subsection{Joint summary}
\label{app:baseline_joint_summary}

Across the three stages of the causal chain, the canonical SOTA
deployment-fair baselines together yield $0/168$ sphere rescues at
$\rho_{\mathrm{crit}}$, while \texttt{PowerRemap}, applied at the
activation site identified by the localization analysis of
\S\ref{sec:activation_patching}, rescues $6/6$ on the same seeds with
output quality matching the clean baseline.
Table~\ref{tab:baseline_summary} consolidates the headline numbers.

\begin{table}[h!]
\centering
\small
\setlength{\tabcolsep}{5pt}
\renewcommand{\arraystretch}{1.05}
\begin{tabular}{@{}llcc@{}}
\toprule
\textbf{Stage} & \textbf{Baseline (representative)} & \textbf{Trials} & \textbf{Sphere rescue} \\
\midrule
1 (input)      & WLOP \citep{huang2009consolidation}, it.\ $\in\{1,2,5,15\}$        & $48$ & $0/48$ \\
1 (oracle)     & WLOP $+\,\Pi_{\mathbb{S}^2}$, it.\ $\in\{2,5,15\}$                 & $18$ & $0/18$ \\
2 (diffusion)  & EDM-style noise inj.\ \citep{karras2022elucidating}, all $8$ steps $\times$ $2$ strengths & $96$ & $0/96$ \\
3 (alternative)& Best-of-K \citep{ma2025inferencescaling}, $K\!\in\!\{1,2,3,4\}$    & $24$ & $0/24$ \\
\midrule
---            & \texttt{PowerRemap} ($\gamma{=}100$, $\mathbf{Y}_{4,7}$)            & $6$  & $\mathbf{6/6}$ \\
\bottomrule
\end{tabular}
\caption{Headline rescue rates. The Stage 1 oracle row is included for
calibration only and is not deployment-fair. \texttt{PowerRemap} is
evaluated on the same $6$ seeds as the baselines.}
\label{tab:baseline_summary}
\end{table}

The three deployment-fair baselines fail for distinct, mechanistically
informative reasons. Stage 1 (WLOP) fails because local-repulsion
uniformization at fixed $N$ does not reach a uniformity level that
crosses the basin separatrix --- a conclusion confirmed by the oracle
ablation. Stage 2 (noise injection) fails because the basin separatrix
is sharp and the trajectory's accumulated directional bias in
$\mathbf{Y}_{4,7}$ cannot be transported across by symmetric Gaussian
noise --- a conclusion consistent with the diffusion-dynamics analysis
of \S\ref{sec:diffusion-dynamics} and the directional-drift result of
\S\ref{sec:spectral_causality}. Stage 3 (best-of-K) fails because the
speckle attractor's basin is dominant at $\rho_{\mathrm{crit}}$, so
seed retry does not transport the trajectory.

The unified picture is that no single intervention upstream of the
identified commit site --- in input space, in mid-trajectory noise, or
in initial-noise resampling --- rescues at the operating point we
study. \texttt{PowerRemap} succeeds because it acts at the commit site
itself, on the directional content of the activation that the
mechanistic analysis identifies as the lever
(\S\ref{sec:activation_patching}, \S\ref{sec:spectral_causality}).

\end{document}